\documentclass[10pt]{article}

\usepackage{microtype}
\usepackage{graphicx}
\usepackage{subcaption}
\usepackage{booktabs} 

\usepackage[accepted]{tmlr-style-file-main/tmlr}

\def\month{09}   
\def\year{2026}  
\def\openreview{\url{https://openreview.net/forum?id=ckWRSrY8i4}}

\usepackage{algorithm}
\usepackage{algorithmic}

\usepackage{hyperref}
\usepackage{url}

\usepackage{amsmath}
\usepackage{amssymb}
\usepackage{mathtools}
\usepackage{amsthm}

\usepackage{amsmath,amsfonts,bm}
\usepackage{hyperref}
\usepackage{url}
\usepackage{algorithm}
\usepackage{algorithmic}
\usepackage{xspace}
\usepackage{enumitem}
\usepackage{amsmath}
\usepackage{bbm}
\usepackage{graphicx}
\usepackage{subcaption}
\usepackage{booktabs}       
\usepackage{lipsum}

\def\eqref#1{equation~\ref{#1}}

\def\1{\bm{1}}

\DeclareMathAlphabet{\mathsfit}{\encodingdefault}{\sfdefault}{m}{sl}
\SetMathAlphabet{\mathsfit}{bold}{\encodingdefault}{\sfdefault}{bx}{n}

\newcommand{\E}{\mathbb{E}}

\makeatletter
\newcommand{\ALC@uniqueautorefname}{Line} 
\makeatother

\newcommand{\algcomment}[1]{\textcolor{orange}{\small \textbf{\texttt{\# #1}}}}
\renewcommand{\algorithmiccomment}[1]{\hfill $\rhd$ #1} 

\newcommand{\autorefp}[1]{(\autoref{#1})}

\newcommand{\op}[1]{\operatorname{#1}}

\newcommand{\gaussian}{\mathcal{N}}
\newcommand{\mean}{\boldsymbol{\mu}}
\newcommand{\cov}{\boldsymbol{\Sigma}}

\newcommand{\dataset}{\mathcal{D}}

\newcommand{\pool}{\mathcal{P}}
\newcommand{\bel}[1][]{\boldsymbol{b}^{#1}}
\newcommand{\projMatrix}{\mathbf{A}}
\newcommand{\paramRMOffset}{\boldsymbol{\theta}_*}

\newcommand{\dynamicsNoise}{\mathbf{U}}
\newcommand{\measurementNoise}{\mathbf{V}}
\newcommand{\priorNoise}{\mathbf{W}}

\newcommand{\paramRM}{\boldsymbol{\theta}}
\newcommand{\paramSpaceRM}{\boldsymbol{\Theta}}

\newcommand{\paramSubspace}{\boldsymbol{z}}

\newcommand{\rewardHuman}{r}
\newcommand{\rewardLearned}{r_{\paramRM}}

\newcommand{\trajWin}{\tau_a}
\newcommand{\trajLose}{\tau_b}
\newcommand{\datasetTraj}{\mathcal{D}^{traj}}
\newcommand{\datasetQuery}{\mathcal{D}}

\newcommand{\subEKF}{PreferenceEKF\xspace}
\newcommand{\ensemble}{DeepEnsemble\xspace}
\newcommand{\dropout}{Dropout\xspace}
\newcommand{\laplaceapprox}{Laplace\xspace}
\newcommand{\llmcmc}{LLMCMC\xspace}

\usepackage[capitalize,noabbrev]{cleveref}

\theoremstyle{plain}

\theoremstyle{definition}

\theoremstyle{remark}

\usepackage[textsize=tiny]{todonotes}

\title{Subspace Inference Enables Efficient\\Active Reward Learning from Preferences}

\author{\name Yutai Zhou \email yutaizho@usc.edu \\
      \addr Thomas Lord Department of Computer Science\\
      University of Southern California, Los Angeles, USA
      \AND
      \name Erdem Bıyık \email biyik@usc.edu \\
      \addr Thomas Lord Department of Computer Science\\
      University of Southern California, Los Angeles, USA}

\begin{document}

\maketitle

%
%
%
%
%
%
%

\begin{abstract}
Reinforcement learning from human feedback (RLHF) has emerged as a powerful yet sample-inefficient approach for learning reward models from human preferences, making active learning a critical component in synthesizing informative preference queries. However, effective uncertainty quantification required for active learning remains a key challenge for large neural network reward models. 
In this paper, we introduce PreferenceEKF, a sample-efficient approach that tracks reward model uncertainty by framing active preference learning as a sequential Bayesian filtering problem. 
Instead of relying on computationally prohibitive posterior inference over the full neural network parameter space, our method performs sequential inference via an extended Kalman filter within a low-dimensional parameter subspace, continuously updating the reward model posterior as new preference queries arrive. Our approach enables scalable sampling of neural network parameters to efficiently compute acquisition functions for active reward learning.
Experiments on the D4RL and V-D4RL benchmarks demonstrate that our approach achieves better sample efficiency, runtime, scalability, and calibration compared to other Bayesian deep learning approaches, and the learned reward models lead to competitive offline reinforcement learning policy performance. 
This highlights the potential of scalable Bayesian methods for preference-based reward modeling in RLHF. 
\footnote{Our code is available at \url{https://github.com/yutaizhou/bnn_pref}.}
\end{abstract}

\section{Introduction}  \label{sec:intro}
In recent years, reinforcement learning from human feedback (RLHF) has become the dominant technique for aligning decision-making agents with human intentions \citep{christiano2017deep, ouyang2022training}.
The ease of providing preference feedback has been a crucial factor in its popularity as a feedback type for reward modeling, but since each feedback provides at most one bit of information, they are also known for their poor sample efficiency; asking a human thousands of comparison questions to learn a reward model (RM) is often not scalable.

A core problem of RLHF is active learning, where we want an agent to be judicious about the queries it asks to learn about a human's preferences as efficiently as possible \citep{sadigh2017active, casper2023open, baraka2025humaninteractive}. Many active learning approaches require probabilistic modeling of uncertainty for computing data acquisition functions, making proper uncertainty representation an active area of research \citep{ovadia2019can, tran2020methods, papamarkou2024position}. While Bayesian methods are well-principled, they are hard to scale to large neural networks (NN) \citep{izmailov2021what}. On the other hand, the simplicity of ensemble methods \citep{dietterich2000ensemble, lakshminarayanan2016simple} and dropout \citep{srivastava2014dropout,gal2016dropout} has made them popular choices for active learning. However, training multiple models can be computationally intensive, especially for large NN reward models.
Although dropout gets around this issue, its effectiveness has been a subject of debate \citep{fort2020deep,osband2022neural,hron2018variational}.

Due to recent advancements in approximate inference, Bayesian deep learning has become increasingly scalable \citep{daxberger2024laplace, shen2024variational}. In this work, we develop a method called \subEKF that enables efficient training of Bayesian neural networks for representing reward models in active preference-based reward learning.
Specifically, by performing Bayesian filtering in a constructed neural network subspace, we maintain model uncertainty in a compute- and memory-efficient manner. The reduced dimensionality of the subspace enables application of the extended Kalman filter (EKF), a classic inference method, for training neural networks. This allows sampling of an arbitrary number of reward models from the model posterior, and usage of the samples for computing common uncertainty-based acquisition functions such as expected information gain and disagreement \citep{hennig2012entropy,hernandez-lobato2014predictive, biyik2022learning}. 

To the best of our knowledge, we are the first to leverage subspace filtering to train neural network reward models from preference feedback. We compare our method, \subEKF, to four widely used Bayesian deep learning methods for active preference-based reward learning in continuous control tasks. We further test whether the learned reward models are useful for policy optimization in offline RL tasks \citep{levine2020offline}. We conduct our experiments in the D4RL \citep{fu2020d4rl} and the V-D4RL \citep{lu2023challenges} benchmarks. Our findings are as follows:

\begin{itemize}[nosep]
    \item \subEKF performs on par with or better than all Bayesian deep learning baselines in terms of sample efficiency and calibration in preference modeling tasks.
    \item \subEKF's runtime is not only much faster, but also scales better with both model size and number of posterior samples compared to all other methods. 
    \item When used for policy optimization, reward models learned using \subEKF resulted in policies with performance competitive with those optimized with reward models learned using other methods.
\end{itemize}

\section{Related Work} \label{sec:related}

\textbf{Reinforcement learning from human preferences.}
While early works in reward learning focused on learning from expert demonstrations \citep{abbeel2004apprenticeship, finn2016guided, ho2016generative}, much of the recent interest has focused on reward learning from pairwise comparisons where human annotators are asked to compare two potential outcomes, e.g., labels, responses, or trajectories \citep{wirth2017survey, christiano2017deep, brown2019extrapolating}. Although preference feedback is much easier for annotators to provide than demonstrations, the minimal amount of information contained within a binary preference query necessitates collection of large amounts of feedback data.

Active learning is a widely used approach for minimizing the time-consuming process of collecting human feedback. It is a sequential problem in nature, as it iteratively collects the most useful data sample based on the model's current state, such as parameter posterior uncertainty. \citep{sadigh2017active, settles2009active}. 
While Bayesian methods have been successfully applied to obtain posteriors for active reward learning using lower-dimensional linear and Gaussian process reward models \citep{biyik2022learning, biyik2024active}, they have not been widely adopted for neural reward models, since acquisition functions typically require sampling from the high-dimensional distribution of model parameters. Instead, ensembles and dropout have been the key enabler of neural network-based active reward learning \citep{lee2021pebble, christiano2017deep}. Our work focuses on efficient yet performant posterior inference for active reward learning, without expensive training of multiple independent models.

\textbf{Uncertainty quantification for neural networks.} 
Classic Bayesian methods that have been successfully used for neural network uncertainty quantification include Laplace approximation \citep{daxberger2024laplace}, Hamiltonian Monte Carlo \citep{neal2011mcmc}, and variational inference \citep{blei2017variational}. While not strictly motivated by Bayesian principles, the simplicity of ensembling \citep{dietterich2000ensemble, lakshminarayanan2016simple} and dropout \citep{srivastava2014dropout, gal2016dropout} has made them popular for uncertainty quantification (UQ). Although the dropout method gets around ensemble method's cost of training multiple independent models, it has been shown to lead to poor posterior approximation quality \citep{hron2018variational, osband2022neural}.

Bayesian filtering methods, which focus on inferring hidden states from noisy observations, provide a principled approach to sequential learning, and have been widely used in robotics and signal processing \citep{thrun2005probabilistic, sarkka2023bayesian}.
Application of Bayesian filtering for training neural networks \citep{singhal1988training, defreitas2000hierarchical} has only recently been applied to deep networks via subspace methods by \citet{duran-martin2022efficient}.

Instead of deriving epistemic uncertainty from posterior inference, a separate line of work has focused on leveraging nonparametric statistics techniques such as the bootstrap to perform UQ \citep{efron1992bootstrap}, and has successfully applied this technique for exploration in deep reinforcement learning \citep{osband2018randomized, osband2016deep}. The same group of authors has also leveraged joint predictions for UQ, and has applied the idea to finetuning large language models \citep{osband2023epistemic, osband2023finetuning}. Our work leverages Bayesian filtering to train neural network reward models in active reward learning settings, where we focus primarily on parameter uncertainty instead of joint predictive uncertainty.

\textbf{Subspace methods for neural networks.} \label{sec:related_subspace}
While there exists a vast literature on decreasing neural network size for efficient training and serving via architecture search \citep{elsken2019neural}, quantization \citep{gholami2021survey}, and pruning \citep{frankle2022lottery}, we focus only on works that enable tractable inference in the reduced model. Specifically, there is growing evidence that the number of parameters required for a neural network to solve a task is often much smaller than the network's total parameter count; optimization and inference in the subspace spanned by these sub-parameters offer not only computational efficiency, but also tractability of applying Bayesian methods for neural network training \citep{fort2020deep, larsen2022how}. These parameters are found either as a subset of neural network parameters, or within a lower-dimensional subspace thereof.

Methods focusing on parameter subsets typically apply Bayesian methods such as Bayesian linear regression or variational inference to the last layer of the neural network, and point estimation methods like stochastic gradient descent (SGD) to the intermediate layers \citep{snoek2015scalable, harrison2023variational, brunzema2024bayesian}. On the other hand, subspace methods typically construct the low-dimensional subspaces via either random projection or singular value decomposition of SGD iterates of the full network; any inference or optimization technique such as sliced sampling \citep{izmailov2020subspace} or SGD \citet{li2018measuring} can then be applied in the subspace in a tractable manner.

\section{Preliminaries}  \label{sec:prelim}

\textbf{Preference-based reward modeling.}
We consider a Markov decision process (MDP) $\langle\mathcal{S},\mathcal{A},\mathcal{T},r,\gamma\rangle$ with state space $\mathcal{S}$, action space $\mathcal{A}$, transition function $\mathcal{T}$, reward function $r:\mathcal{S}\rightarrow\mathbb{R}$, and discount factor $\gamma \in [0,1)$. 
We assume access to a dataset of trajectories $\datasetTraj=\{\tau_1,\ldots,\tau_N\}$, where each trajectory $\tau_i$ is a sequence of $T$ steps $\tau_i=\{(s_{i,t},a_{i,t},s_{i,t+1})\}_{t=0}^{T-1}$, with each step consisting of state $s_t \in \mathcal{S}$, action $a_t \in \mathcal{A}$, and next-state $s_{t+1} \in \mathcal{S}$.
In preference-based reward modeling, we do not assume access to a reward function. Instead, our task supervision comes from annotators who provide binary preference labels over pairwise trajectory comparisons, and the goal is to learn the annotator's reward function that informed their preference.

Formally, an annotator takes a query $Q_i=(\trajWin^i,\trajLose^i)$, and returns a preference label over the two trajectories $y_i=\mathbbm{1} (\trajWin^i \succ \trajLose^i) \in \{0,1\}$ according to their internal reward function $\rewardHuman$. Given a dataset of queries and responses $\datasetQuery=\{Q_i,y_i\}_i$, a widely-used approach for preference learning is to approximate $\rewardHuman$ with a parameterized reward model $\rewardLearned$ via maximum likelihood estimation, where the likelihood $p_\theta(y\mid \trajWin, \trajLose)$ is typically defined using the Bradley-Terry (BT) model \citep{bradley1952rank},
\begin{align}
p_\theta(y\mid \trajWin, \trajLose)
&=p_\theta(\trajWin \succ \trajLose) =\frac{\exp(\beta \cdot \mathcal{R}_{\paramRM}(\trajWin))}{\exp(\beta \cdot\mathcal{R}_{\paramRM}(\trajWin)) + \exp(\beta \cdot\mathcal{R}_{\paramRM}(\trajLose))}\:. \label{eq:bradleyTerry}
\end{align}
In particular, $\beta$ is a temperature parameter that models noisily optimal behavior of an annotator, and $\mathcal{R}_{\paramRM}(\tau_i)$ is the return of trajectory $\tau_i$ where the per-timestep reward is computed using a neural network-based RM $\rewardLearned$, i.e., $\mathcal{R}_{\paramRM}(\tau_i)=\sum_{t=0}^{T-1} \rewardLearned(s_{i,t})$ \citep{lee2021bpref}. 
\footnote{This formalism extends to state or state-action RMs, and whole trajectories or partial trajectory segments. Our experiments use state-based RM and partial trajectories.}

\textbf{Information-theoretic active learning.}
We adopt the acquisition function from \citet{biyik2020asking}, InfoGain, for active preference-based reward learning, which assumes a distribution over RM parameters $p(\paramRM)$ such that, given a query-response pair $(Q_i,y_i)$, the predictive distribution is given by $p(y \mid Q)=\mathbb{E}_{p(\paramRM)}[p(y \mid Q, \paramRM)]$.
Using InfoGain leads to selection of the query $Q_i$ that maximizes the mutual information between the query's preference label $y_i$ and model parameter $\paramRM$:
\begin{subequations}
\begin{align}
Q_i^* 
& =\underset{Q_i}{\arg \max }~ I\left(\paramRM ; y_i \mid Q_i, \bel[i-1]\right) \label{eq:infogain}\\
& =\underset{Q_i}{\arg \max }~ H\left(y_i \mid Q_i, \bel[i-1]\right)-\mathbb{E}_{\paramRM} \left[H(y_i \mid \paramRM, Q_i)\right] \label{eq:infogain_pes}
\end{align}
\end{subequations}
where $I$ is the mutual information, $H$ is the Shannon entropy \citep{cover2006elements}, and $\bel[i-1]=p(\paramRM \mid \datasetQuery_{1:i-1})$ is the posterior distribution over RM parameters after learning from $(i-1)$ queries. This acquisition function is developed from the Bayesian active learning literature, which we detail in \autoref{appendix:acq}. We approximate this acquisition function via sampling as follows:
\begin{equation}
Q_i^* \doteq \ \underset{Q_i}{\arg \max }~ \frac{1}{M} \sum_{y_i \in \{0,1\}} \sum_{\paramRM \in \paramSpaceRM} 
P(y_i \mid Q_i, \paramRM) \log _2\left(\frac{M \cdot P\left(y_i \mid Q_i, \paramRM\right)}{\sum_{\paramRM^{\prime} \in \paramSpaceRM} P\left(y_i \mid Q_i, \paramRM^{\prime}\right)}\right)  \label{eq:infogain_sample}
\end{equation}

where $\paramSpaceRM$ is the set of models sampled from the posterior $\bel[i-1]$, and $M$ is the number of drawn samples. This approximation is asymptotically equivalent to \autoref{eq:infogain_pes} as $M \rightarrow \infty$. We refer to Section 9.1 of \citet{biyik2020asking} for the proof. Due to the necessity of sampling models from the model posterior $\bel[i-1]$, \citet{biyik2020asking} restricted themselves to low-dimensional RMs, such as linear models. We now present our method, \subEKF, which enables sampling of high-dimensional RMs like neural networks, which in turn allows us to scalably compute sampling-based acquisition functions like InfoGain to perform active learning.
\section{Method}  \label{sec:method}

Sampling neural network models to approximate acquisition functions as in \autoref{eq:infogain_sample} can be expensive due to the high-dimensional parameter space of neural networks \citep{izmailov2021what}. 
We leverage the insight that neural networks are overparameterized and that solutions actually live in a much smaller subspace \citep{li2018measuring}, and perform posterior inference within this subspace. This allows us to sample an arbitrary number of models from a lower-dimensional posterior to approximate \autoref{eq:infogain_sample}, without, e.g., the overhead of training ensembles. 
We first show how to use extended Kalman filter (EKF), a widely used filtering algorithm, to train neural network reward models from preference data, then we show how to scale EKF to deep neural networks using subspace methods \citep{duran-martin2022efficient}. The full algorithm is shown in \autoref{alg:subspaceEKF}.

\textbf{EKF for training neural networks.} 
Using the formulation of sequential Bayesian inference, we perform posterior inference of neural network parameters from streaming data $\mathcal{D}_{1: i-1}=\{(Q_1, y_1), \ldots,(Q_{i-1}, y_{i-1})\}$. Starting from some prior belief $\bel[0]=p(\paramRM)$ on the parameters, our posterior after observing $i$ samples can be expressed using Bayes' rule as follows:
\begin{equation}
\begin{aligned}
p(\paramRM_i \mid \dataset_{1:i})
&\propto \underbrace{p(\dataset_{i} \mid \paramRM_{i})}_{\text{Measurement}} \ \color{blue} p(\paramRM_{i} \mid \dataset_{1:i-1}) \\
\color{blue} p(\paramRM_{i} \mid \dataset_{1:i-1})
&= \int \underbrace{p(\paramRM_{i} \mid \paramRM_{i-1})}_{\text{Dynamics}} \ \underbrace{p(\paramRM_{i-1} \mid \dataset_{1:i-1})}_{\text{Previous posterior}} d\paramRM_{i-1}
\label{eq:sequential_bayesian}
\end{aligned}
\end{equation}
where $p(\paramRM_{i-1} \mid \dataset_{1:i-1})$ is the posterior belief over parameters after observing $i-1$ samples, which is combined with a parameter dynamics model and measurement model to form the posterior after observing the $i^{\textrm{th}}$ example. This formulation naturally allows for recursive estimation of model parameters by observing samples one at a time. 

To tractably compute \autoref{eq:sequential_bayesian}, we assume additive Gaussian noise for the dynamics model $p(\paramRM_i \mid \paramRM_{i-1}) = \mathcal{N}(\paramRM_i \mid g(\paramRM_{i-1}), \dynamicsNoise)$ and the measurement model $p(\dataset_i \mid \paramRM_{i}) = \mathcal{N}(y_i \mid h(\paramRM_{i}, Q_i), \measurementNoise)$, where $\dynamicsNoise \in \mathbb{R}^{|\paramRM| \times |\paramRM|}$ and $\measurementNoise \in \mathbb{R}^{|y| \times |y|}$ are prespecified Gaussian noise covariance matrices.
We treat neural network parameters as hidden states, and model the state dynamics $g(\paramRM_{i-1}): \mathbb{R}^{|\paramRM|} \rightarrow \mathbb{R}^{|\paramRM|}$ using an identity function. For preference learning, we model measurements $h(\paramRM_{i}, Q_i): \mathbb{R}^{|\paramRM|} \times \mathbb{R}^{|Q|} \rightarrow \mathbb{R}^{|y|}$ using BT model $p_\theta(\tau_a \succ \tau_b)$ computed using the learned RM $\rewardLearned$ (\autoref{eq:bradleyTerry}).
These assumptions make the model parameter inference objective in \autoref{eq:sequential_bayesian} solvable in closed-form with the EKF algorithm, where the posterior takes a Gaussian form $\bel[i]=\mathcal{N}(\mean_i,\cov_i)$ with $\mean_i \in \mathbb{R}^{|\paramRM|}$ and $\cov_i \in \mathbb{R}^{|\paramRM| \times |\paramRM|}$. In \autoref{appendix:ekf_formula}, we show the exact form of the EKF update procedure and provide further discussion on its linearized Gaussian form of the BT likelihood.

\textbf{Subspace inference.}
Inference in the full parameter space of a neural network is difficult, as the size of the covariance matrix $\cov_i$ of the Gaussian posterior scales in $O(|\paramRM| ^2)$. We instead perform EKF in a learned subspace of the NN: we denote the full space parameter as $\paramRM$ and subspace parameter as $\paramSubspace$, where $|\paramSubspace| \ll |\paramRM|$, 
resulting in posterior $\bel[i]=\mathcal{N}(\mean_i',\cov_i')$ where $\mean_i' \in \mathbb{R}^{|\paramSubspace|}$ and $\cov_i' \in \mathbb{R}^{|\paramSubspace| \times |\paramSubspace|}$.
We further assume a learned affine mapping $\paramRM(\paramSubspace)=\projMatrix \paramSubspace+\paramRMOffset$ that allows us to transform the subspace parameters to the full space. Here $\paramRMOffset$ is initialized via SGD on a small warm-up dataset in the full space. $\mathbf{A} \in \mathbb{R}^{|\paramRM| \times |\paramSubspace|}$ is a fixed projection matrix obtained from applying SVD to the SGD iterates ran in the full space, as shown on Line \ref{algl:sgd_init} through Line \ref{algl:svd}. 
Alternatively, we can construct $\mathbf{A}$ via random projections without relying on SGD iterates. See \autoref{appendix:ekf_implementation} for subspace construction details and \autoref{sec:experiment_subdim} for an ablation study. We further note that although \autoref{alg:subspaceEKF} indicates reliance on an initial dataset, we show in \autoref{appendix:no_warmup} that \subEKF is still effective without it.

We perform EKF inference in the subspace to obtain an estimate $\bel[i]=p(\paramSubspace \mid \datasetQuery_{1:i})$ after observing each query-response pair $\dataset_i =(Q_i, y_i)$, then project each model sampled from $\bel[i]$ back to the full space via affine projection $\paramRM(\paramSubspace)$ to perform the forward pass of the neural network to predict $\mathbbm{1} (\trajWin^i \succ\trajLose^i)$. 
The predictive distribution is similarly computed via sampling followed by projection as $p(y \mid Q) = \mathbb{E}_{p(\paramSubspace)}[p(y \mid Q, \projMatrix \paramSubspace+\paramRMOffset)]$.
While computing the acquisition function and the predictive distribution require sampling from $\bel[i]$, the posterior update procedure itself is completely deterministic. 

\textbf{Active learning using subspace inference.}
We refer to our approach as \subEKF, and draw comparison to the commonly used ensemble method, which we refer to as \ensemble. We assume pool-based active learning \citep{settles2009active}, where we denote the pool of possible binary preference queries as $\pool$.\footnote{Given a dataset of $N$ trajectories, there would be $|\pool| = \binom{N}{2}$ possible pairwise comparison queries.}
For belief initialization (Line \autoref{algl:belief_init}), whereas \subEKF uses a zero-mean isotropic Gaussian of subspace dimension $|\paramSubspace|$, \ensemble initializes $M$ independent models each of dimension $|\paramRM|$.

After belief initialization, the sequential phase of active learning begins. For random querying, Line \autoref{algl:computeQuery} amounts to simply retrieving a random query from the query pool $\mathcal{P}$, whereas active learning algorithms compute an acquisition function for the optimal query to retrieve from the pool. The algorithm then receives the corresponding label for the retrieved query from an annotator in Line \autoref{algl:get_label}, where the annotator can either be a human-in-the-loop or a simulated oracle that expresses preference based on ground truth environment rewards. 
For belief update (Line \autoref{algl:belief_update}), whereas \subEKF performs Bayesian update in the constructed subspace on only the most recent query-response pair $\datasetQuery_i$, \ensemble trains each of the $M$ models using gradient descent on all data seen so far. 

The most common uncertainty-based acquisition function is ensemble disagreement, i.e., pick the query $Q_i$ for which the predicted preference label $\mathbbm{1} (\trajWin^i \succ\trajLose^i)$ has the highest variance across the ensemble. Disagreement has been popular for neural network-based active learning where it is expensive to scale Bayesian methods to high-dimensional settings \citep{christiano2017deep, lee2021pebble}, while InfoGain is the current state-of-the-art acquisition function for lower-dimensional reward learning settings \citep{biyik2020asking, biyik2024active, ellis2024generalized}. Although our method can be used to compute any sampling-based acquisition function, we specifically leverage \subEKF's ability to sample from high-dimensional distributions to scale InfoGain (\autoref{eq:infogain_sample}) to neural network models.
Due to the difficulty of sampling from high-dimensional parameter distributions, \ensemble approximates InfoGain by training multiple independent models, while \dropout does so by sampling parameter dropout masks during inference.


\begin{algorithm}[hbtp] 
    \caption{\subEKF for active preference-based reward learning}
    \label{alg:subspaceEKF}
    \begin{algorithmic}[1]
        \STATE \textbf{Input:}
        \STATE $\pool$: Pool of all binary preference queries without labels
        \STATE $\dataset^{\text{init}}= \{(Q_i, y_i)\}_{i=1}^\tau$: Initial preference dataset with $\tau$ (query, label) pairs
        \STATE $B$: query budget limit
        \STATE $w$: number of SGD iterations for subspace construction
        \STATE \textbf{Procedure:} 
            \STATE \algcomment{Subspace Construction}
            \STATE  \label{algl:sgd_init} $\paramRM_{1:w} = \op{SGD}({\dataset^{\text{init}}})$ \COMMENT{$\paramRM_{1:w}=[\paramRM_1, \ldots, \paramRM_w]$}
            \STATE $\paramRM_*=\paramRM_w$ 
                \COMMENT{Parameter offset: $\paramRMOffset \in \mathbb{R}^{|\paramRM|}$}
            \STATE $\projMatrix=\op{SVD}\left(\paramRM_{1:w}\right)$  \label{algl:svd}
                \COMMENT{Projection matrix: $\projMatrix \in \mathbb{R}^{|\paramRM| \times |\paramSubspace|}$}
            
            \STATE \algcomment{Subspace Inference}
            \STATE $\bel[0](\paramSubspace)=\gaussian(\mean_0',\cov_0')$ \label{algl:belief_init}
            \FOR{$t=1:B$} \label{algl:for_loop}
                
                \STATE $Q_t = \op{ComputeQuery}(\bel[t-1], \projMatrix,\paramRMOffset, \mathcal{P})$
                    \label{algl:computeQuery}
                \STATE $y_t = \op{GetLabel}(Q_t)$
                    \label{algl:get_label}
                \STATE $\bel[t]=\op{EKF}(\bel[t-1],(Q_t, y_t))$ 
                    \label{algl:belief_update}
            \ENDFOR
            
    \end{algorithmic}
\end{algorithm}
\section{Experiments} \label{sec:experiments}

\textbf{Baselines and Evaluation.}
We compare \subEKF to four Bayesian deep learning baselines commonly used for reward modeling: \ensemble, \dropout, \laplaceapprox, and last-layer Markov chain Monte Carlo (\llmcmc), which we detail in \autoref{appendix:baselines}.
We address the following questions:
(1) Does preference-based reward learning with \subEKF lead to better sample-efficiency, likelihood-based evaluation, and model calibration compared to the baselines?
(2) Does subspace representation of parameter uncertainty $p(\paramRM \mid \datasetQuery)$ lead to computational advantages over other representations such as ensembles and dropout masks? 
(3) How does the choice of subspace construction method impact our method's performance?
(4) Can reward models sampled from \subEKF's posterior be used for policy optimization via offline RL?
(5) Can we extend our method to image-based trajectories with sparse preference feedback?

In the reward learning experiments, given a limited query budget $B$, we would like to learn RMs from preference queries as sample-efficiently as possible. Evaluation is done by comparing the BT log-likelihood (\autoref{eq:bradleyTerry}) achieved by an RM on a held-out set of test queries throughout training. 
We focus our main experiments on synthetically generated preference labels. To create the preference query pool $\pool$, we randomly sample pairwise partial trajectories from a trajectory dataset $\datasetTraj$, then generate noisily optimal synthetic labels as follows: for a given pair of trajectories, we compute their returns and sample a preference label according to the BT model, where a trajectory is preferred with probability exponentially proportional to its return. 

In the offline RL experiments, the learned RMs are then used for training parameterized policies $\pi_\phi(a\mid s)$ via offline RL. This is done by first labeling the trajectory dataset $\datasetTraj$ with the learned RM: we take the average predicted reward over $M$ models $\rewardLearned^M(s_{i,t})=\frac{1}{M} \sum_{m=1}^M \rewardLearned^m (s_{i,t})$ for each state, where $\rewardLearned^m$ is the $m^{\textrm{th}}$ sampled reward model or ensemble member. A reward-labeled trajectory takes the form, $\tau_i=\{(s_{i,t},a_{i,t},s_{i,t+1}, \rewardLearned^M(s_{i,t}))\}_{t=0}^{T-1}$. We train policies on the reward-labeled $\datasetTraj$ using Implicit Q-Learning (IQL) \citep{kostrikov2021offline}, an empirically successful offline RL algorithm. We evaluate policies by comparing their empirical rollout returns throughout RL training.

\textbf{Tasks.}
We evaluate our approach in D4RL \citep{fu2020d4rl}, a popular offline RL benchmark, and choose a mixture of environments spanning MuJoCo locomotion \citep{todorov2012mujoco}, Adroit Shadow Hand \citep{rajeswaran2018learning}, and Maze2D navigation.
Within each environment, we choose trajectory datasets of varying characteristics: MuJoCo trajectories (HalfCheetah, Hopper, Walker2d) span a range of performance quality, Adroit trajectories (pen twirling) are generated by a human operator and a fine-tuned expert-level RL policy, and maze navigation trajectories are collected from policies executed in mazes of varying difficulty. We consider each dataset as a separate task, for a total of 12 tasks.

\textbf{Implementation Details.}
Unless otherwise stated, all experiments are done on a single node with 8 NVIDIA RTX A6000 GPUs via sharding, query budget $B=60$, and trajectory segments of length 50. On the belief update step (Line \autoref{algl:belief_update}), \subEKF learns from only the most recent query-label pair, while all baselines learn from all data seen so far. With the exception of the scaling experiments in \autoref{sec:experiment_scale} and the ablation experiments in \autoref{sec:experiment_subdim}, all reward models are represented as multi-layer perceptrons (MLP) with two hidden layers of 64 units, using subspace dimensionality $|\paramSubspace|=200$.

\subsection{Does \subEKF lead to sample-efficient active reward learning?} \label{section:experiments_activelearning}
Given a fixed query budget per task, we evaluate each method over 12 seeds. We use state-based partial trajectories, and compute return of each trajectory as $\mathcal{R}_{\paramRM}(\tau_i)=\sum_{t=1}^T \rewardLearned(s_{i,t})$. 
We show in \autoref{fig:logpdf_agg} that aggregated over all tasks (see \autoref{appendix:main} for details on task-aggregation), 
both random and active variants of \subEKF perform on par with or outperform all other baselines in terms of both sample efficiency and the final evaluated log-likelihood.
We show per-task results in \autoref{fig:logpdf_all}, where we found that in most tasks, active \subEKF outperforms 
all other methods in terms of sample efficiency and final log-likelihood. 
We refer to \autoref{appendix:stats} for statistical significance tests backing up these empirical observations. 

\textbf{On acquisition functions:} Although our main result was obtained with all methods using the InfoGain acquisition function, we show additional results using disagreement and entropy acquisition functions in \autoref{appendix:acq}. We found InfoGain to be the only acquisition function that enabled the active variant of all five methods to perform better than or on par with their random counterpart. This reflects findings from previous works that demonstrated InfoGain's higher sample-efficiency compared to other acquisition functions \citep{biyik2020asking}.

\textbf{On posterior sampling:} \ensemble is the only method that needs to train multiple models to represent the posterior, so we set $M=5$ as is commonly done for ensemble-based uncertainty quantification \citep{ovadia2019can}; all other methods can sample an arbitrary number $M$ of models from the learned posterior, so we set $M=100$ for them. This raises the question of whether our method’s higher sample efficiency in preference learning is solely due to the larger number of posterior samples, or whether the learned posterior indeed captures the annotator’s preference. For fair comparison, we show in \autoref{appendix:same_nmodels} results where all methods use the same number of models $M=5$, and found \subEKF to still be the most successful method in terms of sample-efficiency and final log-likelihood.

\textbf{On the unimodality of EKF:} Due to the unimodal Gaussian assumption that EKF places on the measurement function, dynamics functions, and the resulting posterior, we note that our approach is designed primarily for learning the preferences of only a single annotator. We show in \autoref{appendix:human} results where we learn from labels from crowd-sourced human annotators. Due to the multimodality of the resulting preference distribution, none of the methods achieved great log-likelihood evaluation. While we acknowledge pluralistic alignment as a critical open problem of RL from human feedback \citep{casper2023open, sorensen2024position}, we view our work as addressing the complementary and largely orthogonal challenge of improving sample efficiency of preference-based reward learning. 
Extending our framework to accommodate multimodal preference distributions \citep{myers2021learning} may be done by performing inference using non-parametric methods such as particle filters \citep{thrun2005probabilistic}, but this may come as a trade-off for inference efficiency in sequential learning settings. We leave such explorations to future work.

\begin{figure*}[htbp]
    \centering
    \begin{subfigure}{0.49\textwidth}
        \centering
        \includegraphics[width=\textwidth]{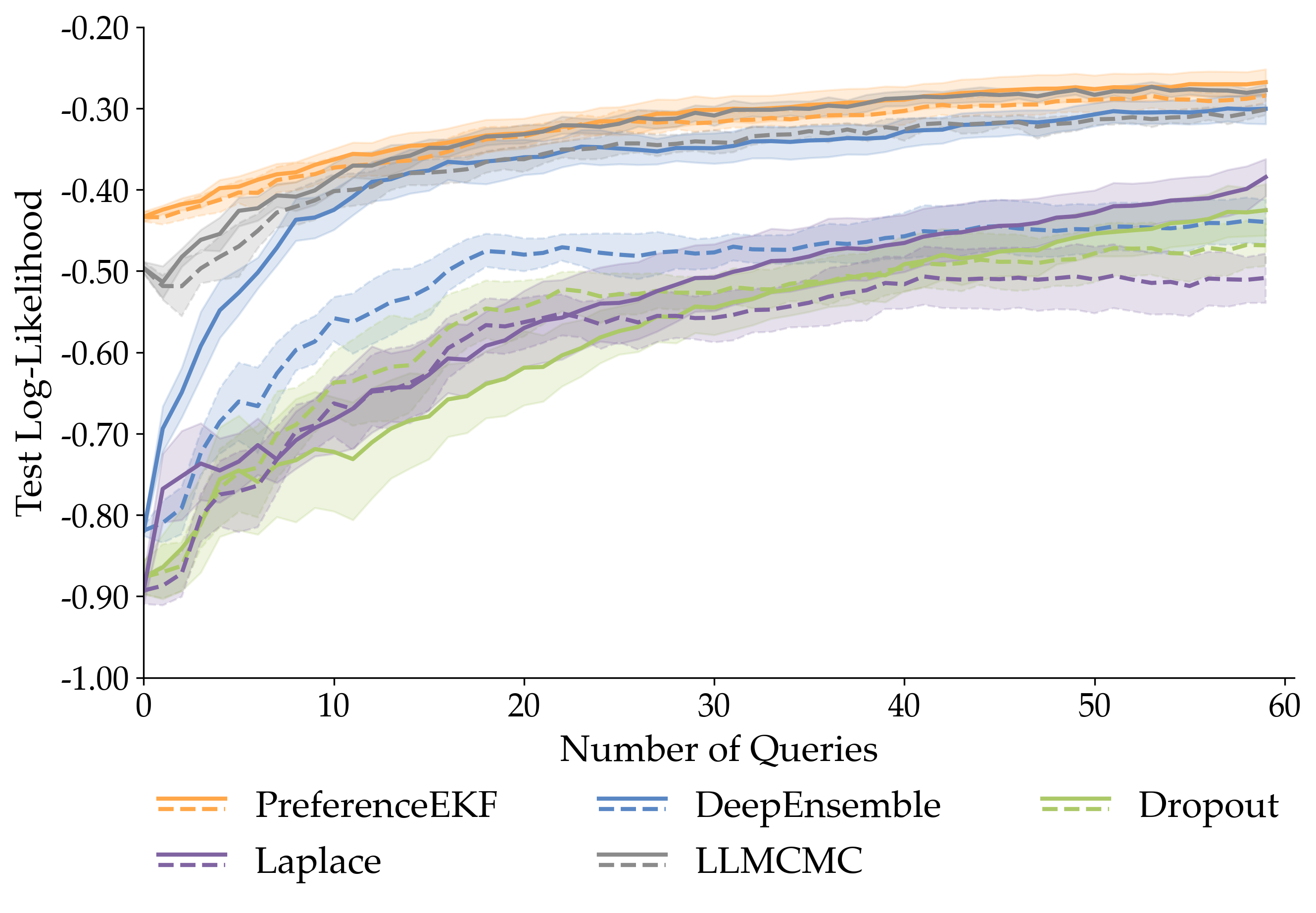}
        \caption{}
        \label{fig:logpdf_agg}
    \end{subfigure}
    \begin{subfigure}{0.49\textwidth}
        \centering
        \raisebox{20pt}
        {\includegraphics[width=\textwidth]{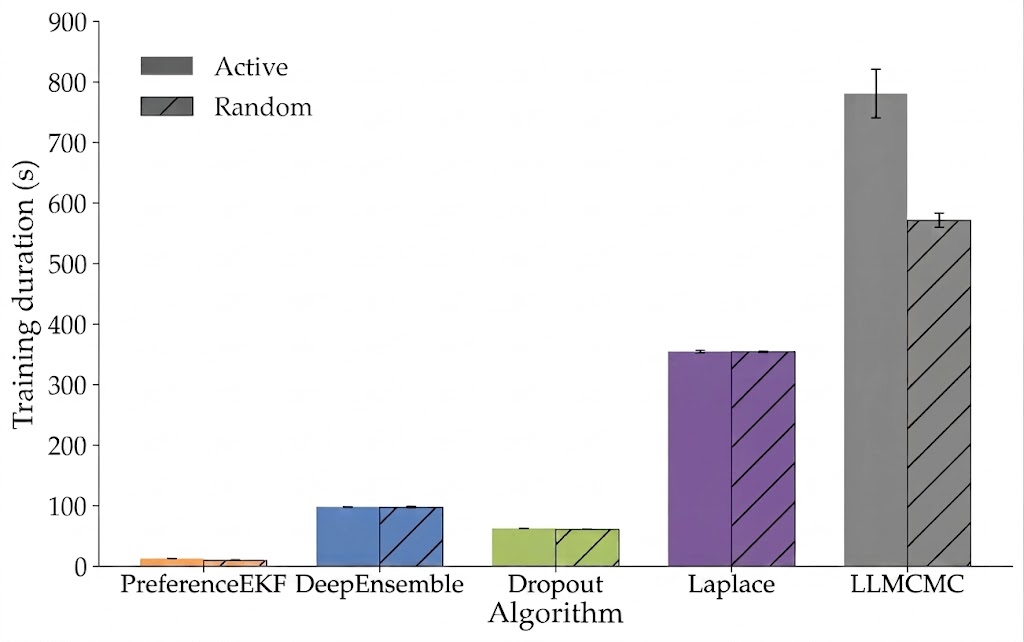}}
        \caption{}
        \label{fig:train_duration}
    \end{subfigure}
    \caption{
    \autoref{fig:logpdf_agg} shows log-likelihood comparison of the random (dashed line) and active (solid line) variants of each method using the InfoGain acquisition function (higher means better fitting of annotator preference distribution).
    \autoref{fig:train_duration} shows training runtime duration of both active and random variants of each method (lower means faster training). For a table version of this plot, please see \autoref{appendix:runtime_table}.
    Each line plot and bar plot is aggregated over 12 D4RL tasks (mean $\pm$ 95\% bootstrap confidence interval over 12 seeds)
    }
\end{figure*}

\subsection{How does training runtime with \subEKF scale?} \label{sec:experiment_scale}
We first note in \autoref{fig:train_duration} that \subEKF training is vastly faster than the baselines, achieving roughly $5\times$ speedup compared to \ensemble and over $40\times$ speedup compared to \llmcmc. This is primarily due to the sequential nature of EKF, making it unnecessary for \subEKF to repeatedly train / perform posterior updates on queries it has already seen. All baseline methods require this so as to prevent catastrophic forgetting, thus slowing down their runtime. We show results that relax this assumption in \autoref{appendix:runtime}.
Lastly, note that the long runtime of \llmcmc is due to the necessity for the MCMC chain to converge, which takes upwards of a thousand steps for every posterior update upon receiving a new query. 

Next, we investigate whether subspace filtering can serve as a scalable alternative to gradient descent for preference learning, with respect to both larger reward models and more model samples. As such, we only compare our method to \ensemble and \dropout, which are primarily based on SGD.
We run all scaling experiments on CPUs as the larger models and ensemble sizes led to out-of-memory errors on GPUs. 
We show in \autoref{fig:scaling_M_duration} that given a fixed architecture of a two-layer MLP with 64 units per layer, the runtime of \subEKF for learning a reward model from $B=60$ queries scales much more gracefully with increasing $M$ compared to other methods. While \dropout does not need to maintain multiple models, it is still slower than \subEKF as it performs model update in full parameter space instead of a lower-dimensional subspace.
\autoref{fig:scaling_M_logpdf} demonstrates that final test log-likelihood favors \subEKF over the other methods, showcasing that our approach maintains consistent performance on top of computational efficiency given increasing $M$. 
\autoref{fig:scaling_param_duration} and \autoref{fig:scaling_param_logpdf} show similar favorable scaling properties of \subEKF except that we fix the number of model samples ($M=5$) and increase the neural network architecture instead. This showcases the scalability of subspace training to not only settings where we need a large number of model samples $M$, but also to settings where we need larger neural networks $| \paramRM |$.

\newcommand{\subfigwidth}{0.23\textwidth}
\begin{figure}[!ht]
    \centering
    \begin{subfigure}[b]{\subfigwidth}
        \centering
        \includegraphics[width=\textwidth]{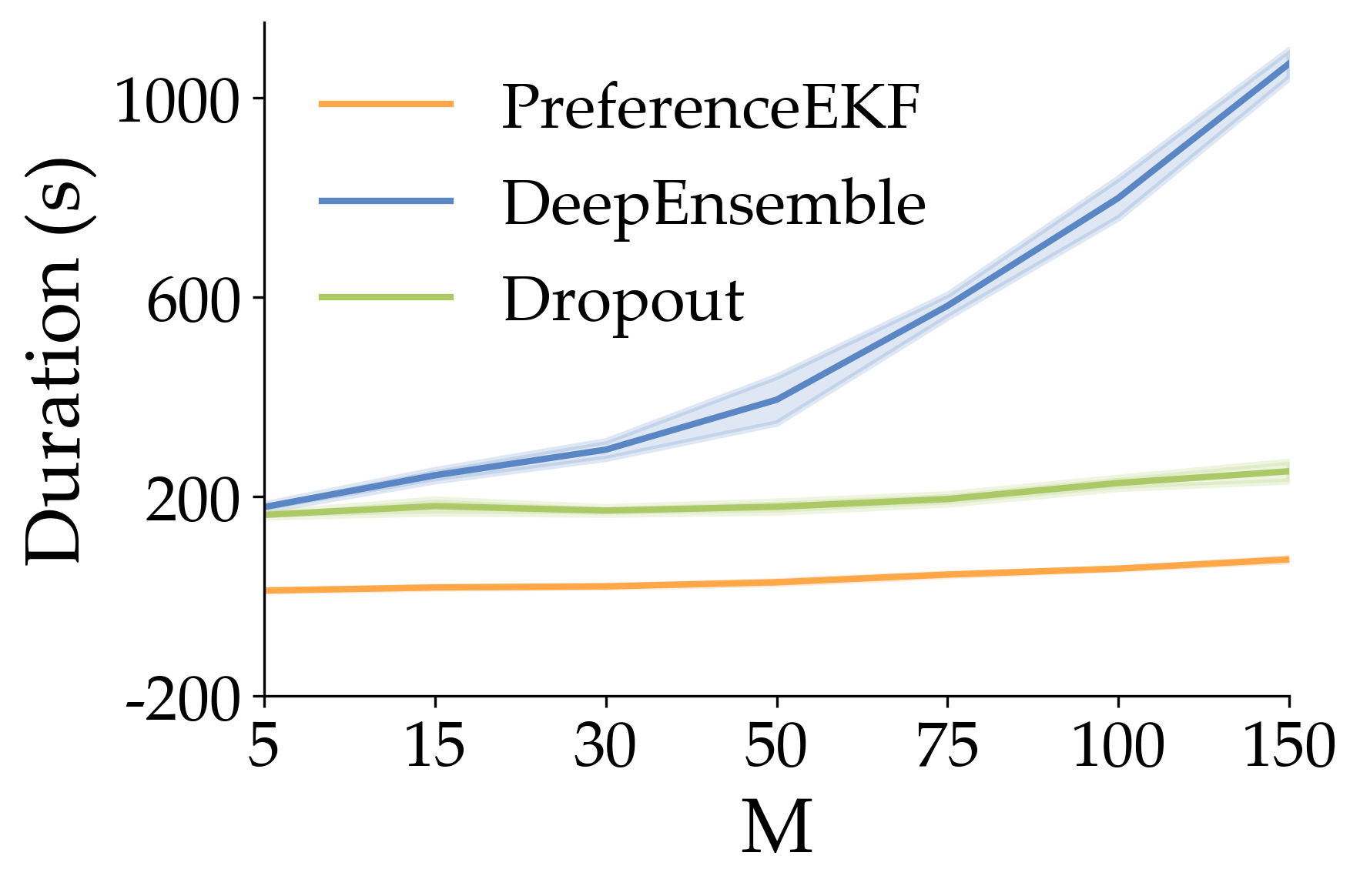}
        \caption{}
        \label{fig:scaling_M_duration}
    \end{subfigure}
    \hspace{0.01\textwidth}
    \begin{subfigure}[b]{\subfigwidth}
        \centering
        \includegraphics[width=\textwidth]{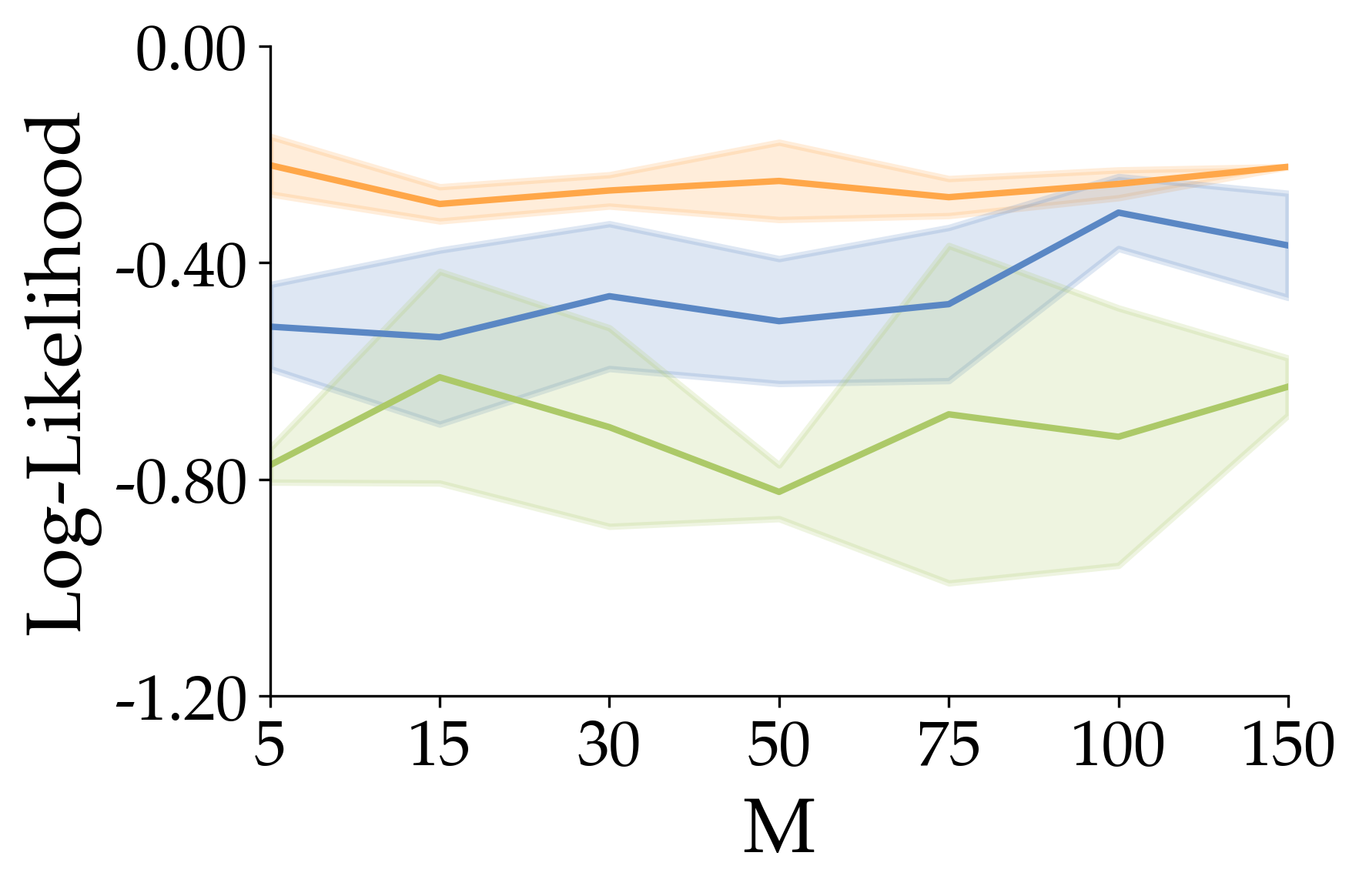}
        \caption{}
        \label{fig:scaling_M_logpdf}
    \end{subfigure}
    \hspace{0.01\textwidth}
    \begin{subfigure}[b]{\subfigwidth}
        \centering
        \includegraphics[width=\textwidth]{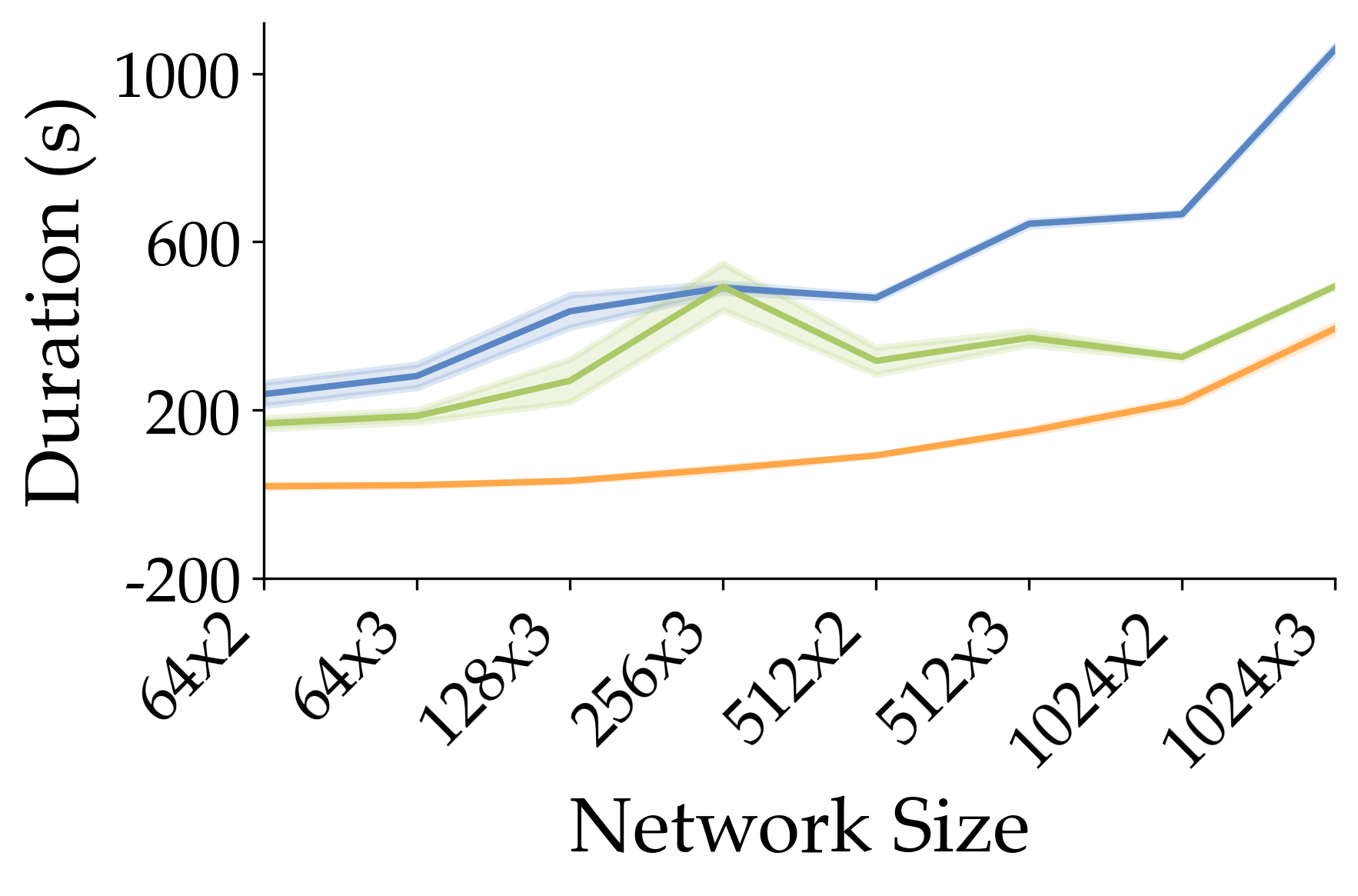}
        \caption{}
        \label{fig:scaling_param_duration}
    \end{subfigure}
    \hspace{0.01\textwidth}
    \begin{subfigure}[b]{\subfigwidth}
        \centering
        \includegraphics[width=\textwidth]{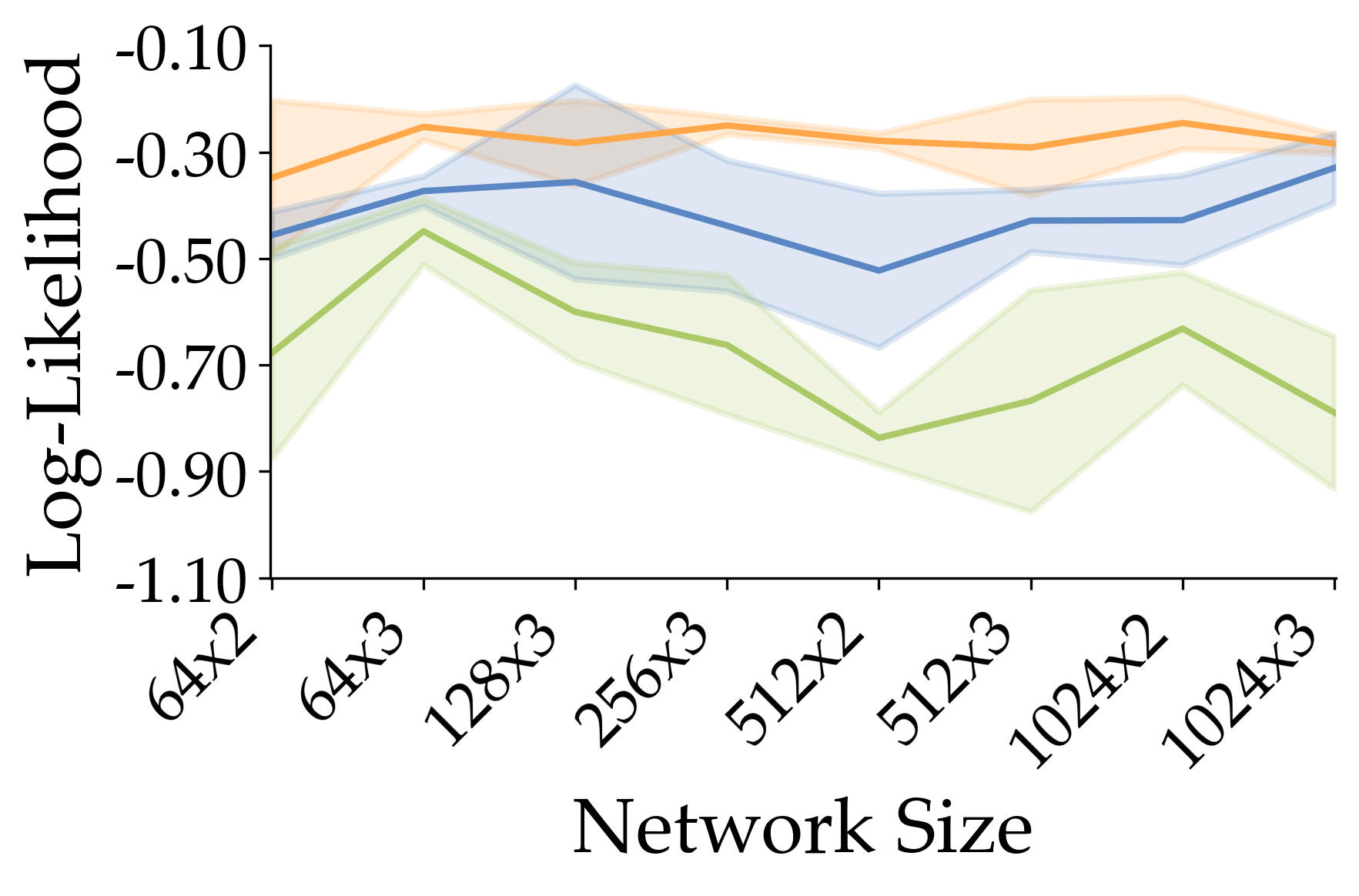}
        \caption{}
        \label{fig:scaling_param_logpdf}
    \end{subfigure}
    \caption{
    \autoref{fig:scaling_M_duration} and \autoref{fig:scaling_M_logpdf} show how runtime scales with the number of model samples $M$ in the active learning setting (mean $\pm$ std over 3 seeds). \autoref{fig:scaling_param_duration} and \autoref{fig:scaling_param_logpdf} show runtime scaling with neural network architecture size.
    Overall, \subEKF has the fastest runtime and the best scaling trend, while retaining high log-likelihood evaluation.
    }
    \label{fig:scaling}
\end{figure}

\subsection{Does \subEKF lead to better model calibration?} \label{sec:experiment_calibration}
While effective representation of parameter uncertainty is crucial for efficient active learning, it is also important for calibration of model predictions \citep{guo2017calibration, ovadia2019can}. We study whether uncertainty quantification (UQ) using subspace inference methods leads to better calibrated model predictions compared to UQ using the baselines, as quantified by two commonly used UQ metrics: expected calibration error (ECE) \citep{naeini2015obtaining, pavlovic2025understanding} and Brier score \citep{brier1950verification, degroot1983comparison}. 

We show in \autoref{fig:calibration} that \subEKF has the lowest ECE among all methods, and the second lowest Brier score behind active \ensemble.
This highlights the quality of posterior approximation achieved by subspace inference methods compared to the other Bayesian deep learning baselines. We provide further calibration experiment details and reliability diagrams in \autoref{appendix:calibration}.

\subsection{Ablation study on subspace construction} \label{sec:experiment_subdim}
\begin{figure*}[htbp]
    \centering
    \begin{subfigure}{0.495\textwidth}
        \centering
        \includegraphics[width=\linewidth]{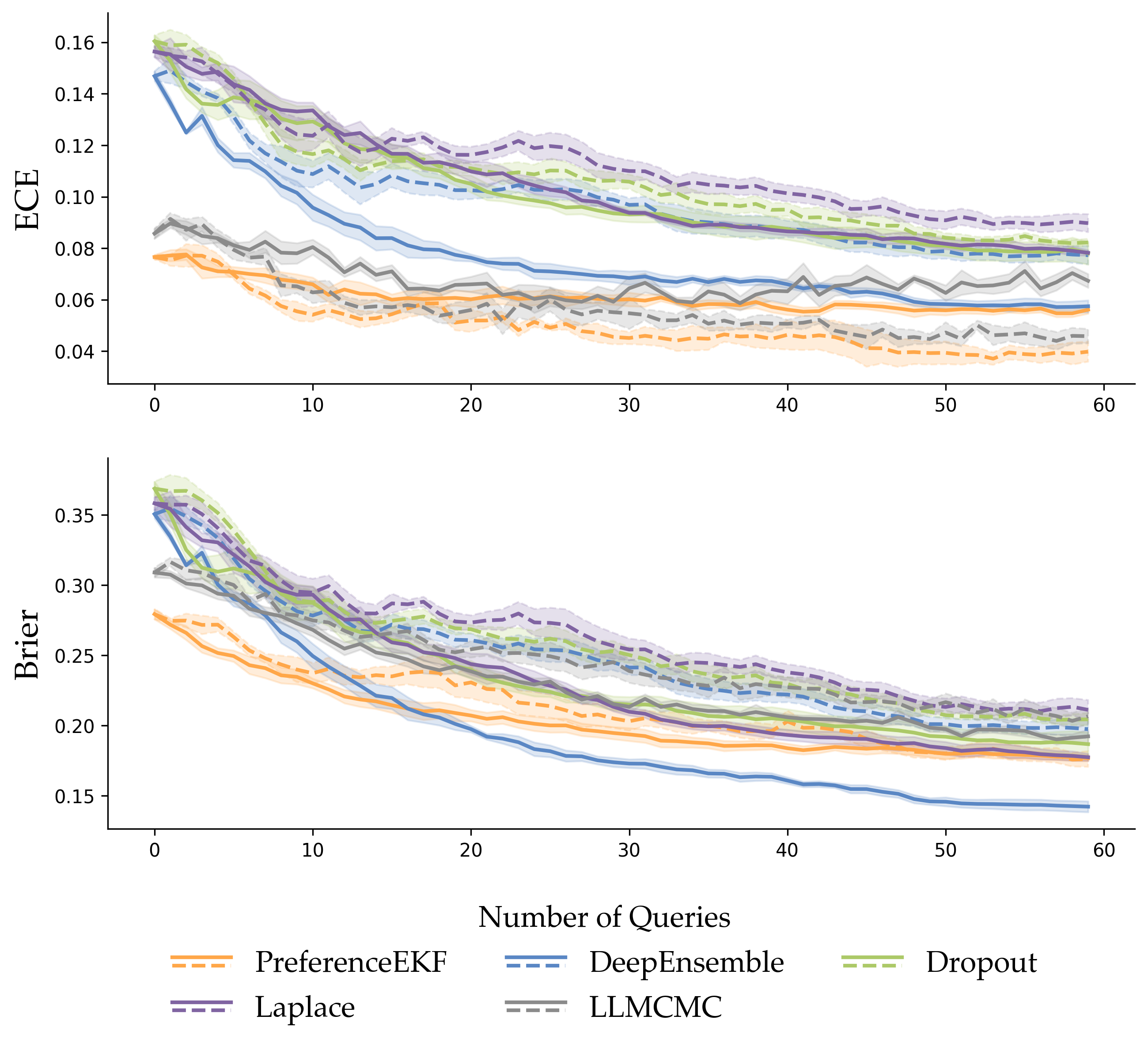}
        \caption{}
        \label{fig:calibration}
    \end{subfigure}
    \hfill
    \begin{subfigure}{0.495\textwidth}
        \centering
        \raisebox{20pt}
        {\includegraphics[width=\linewidth]{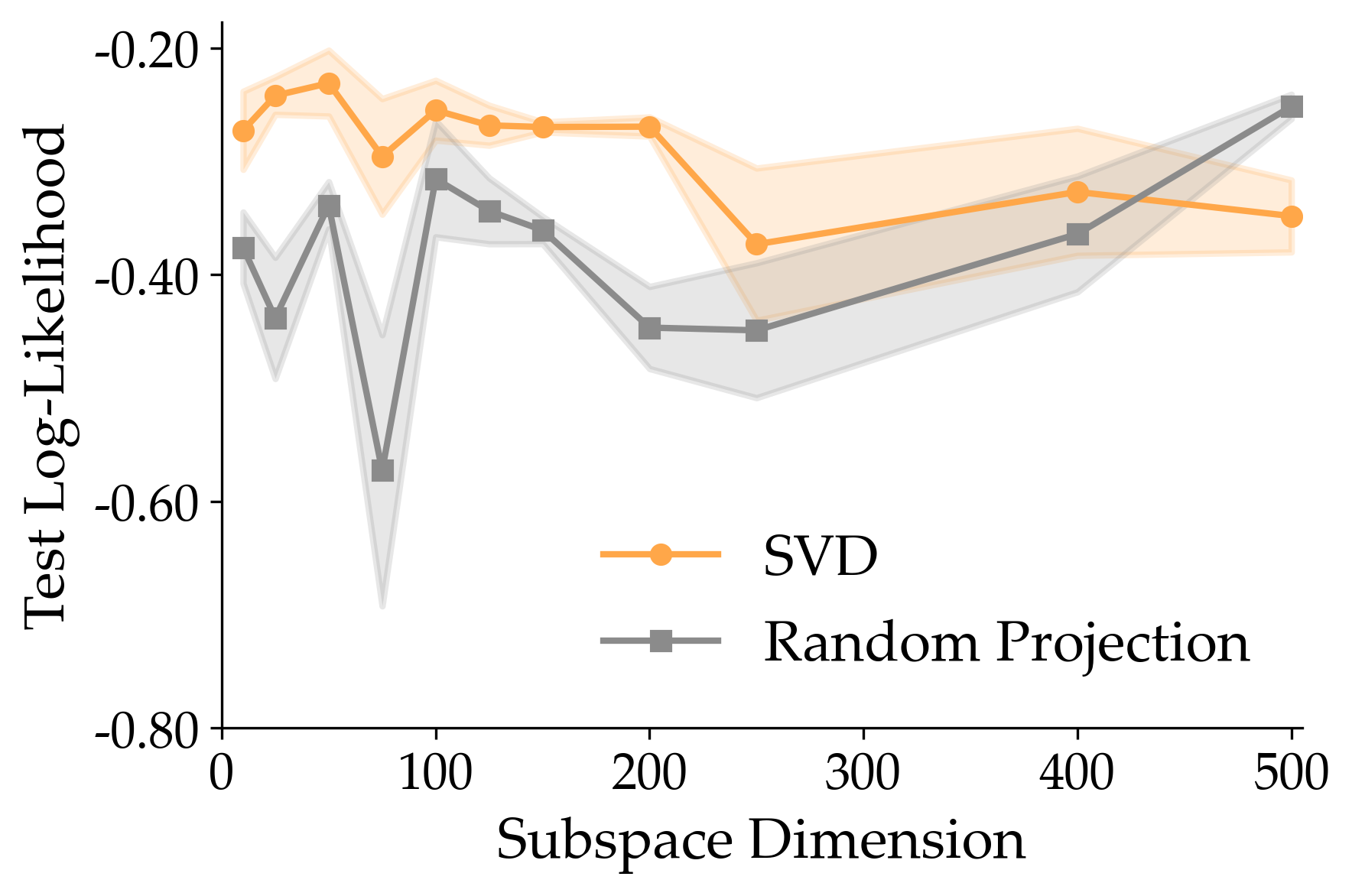}}
        \caption{}
        \label{fig:subdim_ablation}
    \end{subfigure}
    \caption{
    \autoref{fig:calibration} shows calibration results of the random (dashed line) and active (solid line) variants of the methods, as evaluated by expected calibration error (using 10 bins) and Brier score on a test dataset (lower is better for both metrics).
    \autoref{fig:subdim_ablation} shows an ablation over the subspace construction technique for \subEKF, as evaluated by log-likelihood on a test dataset (higher is better). Both the UQ experiment and ablation analysis here are performed over 3 seeds (mean $\pm$ std) on the Walker Medium Expert task.
    }
\end{figure*}

The method for subspace construction for \subEKF can be modified to 1) use varying dimensionality of the subspace, and to 2) use random projection to generate the subspace basis instead of running SVD on SGD iterates \citep{li2018measuring, izmailov2020subspace}. While all of our experiments so far use a fixed dimensionality of $|\paramSubspace|=200$ with SVD-based construction, we perform an ablation analysis over these choices, as shown in \autoref{fig:subdim_ablation}. We observe that while the SVD-based approach works well for smaller subspace dimensions, the random projection approach can eventually reach performance on par with or even outperform the SVD approach as the subspace dimension increases.

We further show in \autoref{appendix:no_warmup} that in the case where no initial dataset is available, belief initialization using the random projection approach is often sufficient for good reward learning performance. This result decouples \subEKF's reliance on SGD altogether. 
For consistency, unless otherwise stated, our main \subEKF experiments are performed with the SVD-based approach that relies on SGD.

\subsection{Can RMs learned using \subEKF be used for policy optimization?}
The goal of the offline RL experiments is to test whether a reward model learned from a limited number of preference queries can be used to optimize a policy that reaches or exceeds the performance of a policy trained with ground-truth environment rewards (GT policy).
All policies are trained using IQL \citep{kostrikov2021offline} over 5 seeds on the reward-labeled dataset for 1M steps, and evaluation is done via 5 rollouts every 50K steps. 
We show in \autoref{fig:rlScore_agg} that aggregated across all tasks, policies induced by reward models learned from all active preference learning methods converge to similar policy performance, with all policies performing on par with or slightly worse than the GT policy. This showcases that our method is capable of producing reward models suitable for policy optimization. 
As the primary goal of our work is to improve the sample efficiency of preference-based reward learning, we leave studies on the interplay between reward learning and policy learning to future work. We provide further discussion of this result in \autoref{appendix:offlineRL_results}.

\subsection{Can \subEKF learn from image data and sparse preference feedback?}
While our main experiments showcase the effectiveness of \subEKF in state-based control tasks, where preference labels are synthetically generated by comparing sum of dense rewards between two trajectories, we apply our method to two additional challenging yet common settings: (1) sparse comparative feedback, and (2) pixel-based control.

Our experiments thus far have relied on using dense trajectory rewards to generate synthetic preference labels, which has allowed us to perform preference learning on partial trajectory segments, thus easing the reward credit assignment problem \citep{wirth2017survey}. However, real robot datasets often only have sparse binary success / failure labels for each trajectory, making it impossible to rely on dense comparative feedback signal for preference learning.
We apply our method to this challenging setting, where we use full real robot trajectories from the SOAR dataset \citep{zhou2024autonomous} across multiple manipulation tasks, and observed favorable results for our approach, which we detail in \autoref{appendix:real_robotics}.

Despite subspace filtering's effectiveness in handling large parameter counts, the difficulty of scaling \subEKF to pixel-based reward models is that the update step of EKF scales cubically with dimensionality of the observation space, which poses scalability challenges to high-dimensional inputs such as images. We resolve this issue by relying on pretrained image embeddings rather than raw pixel inputs, and observed promising results of active preference-based learning of pixel reward models. We refer to \autoref{appendix:pixel} for results and further details on pixel-based tasks. Overall, we believe that the two favorable sets of results here highlight the applicability of our method to the high-dimensional and sparse feedback nature of common real robot data. 

\section{Conclusion} \label{sec:conclusion}
In this work, we successfully adopted extended Kalman filters to train neural networks in an active preference-based reward modeling setting. We showed several advantages of maintaining a subspace distribution over neural network parameters $p(\paramRM \mid \dataset)$, in comparison to four other widely used Bayesian deep learning methods for active reward learning. 
Our approach led to more sample-efficient active reward learning, similarly performant RL policy optimization, better runtime scaling with respect to model size and model sample count, and better calibration through higher-quality uncertainty representation. 

\textbf{Limitations and future work.}
While we found subspace methods to be an effective tool for scaling Bayesian filtering methods for neural network training, it is unclear whether this approach will be effective for applying Bayesian methods to foundation model-scale reward models \citep{mahan2024generative, zhang2024generative}. Due to the unimodality of the Gaussian distribution that the extended Kalman filter maintains, alternative methods may need to be investigated for approximating multimodal posteriors, e.g., learning reward functions from annotators with differing preferences \citep{poddar2024personalizing,siththaranjan2023distributional}. 
We would further like to evaluate uncertainty quantification using the recent works on epistemic neural networks \citep{osband2023epistemic}, which focuses on joint predictive uncertainty instead of the marginal predictive distribution.

Our work primarily focuses on improving sample-efficiency of reward modeling in RLHF, but we would like to further investigate how learned posterior distribution of reward models can aid in an RL policy's exploration and serve as a mechanism for mitigating reward hacking \citep{yang2024bayesian, gao2022scaling, hadfield-menell2017inverse}. Finally, due to its sample-efficiency and adaptivity to non-stationary distributions, we believe the subspace filtering method to be a viable candidate for uncertainty quantification and large model finetuning in robot learning domains \citep{bellemare2017distributional, fridovich-keil2020confidenceaware,bobu2020quantifying}.



\section*{Broader Impacts}
Our work presents an algorithm for active learning in preference-based reward modeling, enhancing the efficiency and accuracy of neural network training in applications requiring subjective human evaluations, such as natural language processing, personalized recommendations, and human-robot interaction. By optimizing data collection around uncertain or high-impact preferences, our approach can reduce labeling costs and improve model alignment with human intentions. However, it is possible that working in the reduced subspace and performing inference with the extended Kalman filter may introduce suboptimalities in preference modeling such as bias amplification or neglect of minority preferences. To mitigate these risks, future research should investigate the robustness of \subEKF and potential information loss caused by subspace reduction.

\section*{Acknowledgments}
We thank Aleyna Kara for initial discussions about the method. Yutai Zhou was partially supported by a fellowship from USC - Capital One Center for Responsible AI and Decision Making in Finance (CREDIF). Erdem Bıyık acknowledges funding by the Airbus Institute for Engineering Research (AIER).

\bibliographystyle{tmlr-style-file-main/tmlr}
\bibliography{_paper/reference_tmlr2026}

@inproceedings{abbeel2004apprenticeship,
  title = {Apprenticeship Learning via Inverse Reinforcement Learning},
  booktitle = {Twenty-First International Conference on {{Machine}} Learning  - {{ICML}} '04},
  author = {Abbeel, Pieter and Ng, Andrew Y.},
  year = 2004,
  pages = {1},
  publisher = {ACM Press},
  address = {Banff, Alberta, Canada},
  doi = {10.1145/1015330.1015430},
  urldate = {2023-09-14},
  langid = {english}
}

@misc{baraka2025humaninteractive,
  title = {Human-{{Interactive Robot Learning}}: {{Definition}}, {{Challenges}}, and {{Recommendations}}},
  author = {Baraka, Kim and Idrees, Ifrah and Faulkner, Taylor Kessler and Biyik, Erdem and Booth, Serena and Chetouani, Mohamed and Grollman, Daniel H and Saran, Akanksha and Senft, Emmanuel and Tulli, Silvia and Vollmer, Anna-Lisa and Andriella, Antonio and Beierling, Helen and Horter, Tiffany and Kober, Jens and Sheidlower, Isaac and Taylor, Matthew E and Xiao, Xuesu},
  year = 2025,
  langid = {english}
}

@article{bell1993iterated,
  title = {The Iterated {{Kalman}} Filter Update as a {{Gauss-Newton}} Method},
  shorttitle = {Iterated {{EKF}}},
  author = {Bell, B.M. and Cathey, F.W.},
  year = 1993,
  month = feb,
  journal = {IEEE Transactions on Automatic Control},
  volume = {38},
  number = {2},
  pages = {294--297},
  issn = {1558-2523},
  doi = {10.1109/9.250476},
  urldate = {2025-05-20}
}

@inproceedings{bellemare2017distributional,
  title = {A {{Distributional Perspective}} on {{Reinforcement Learning}}},
  shorttitle = {C51},
  booktitle = {Proceedings of the 34th {{International Conference}} on {{Machine Learning}}},
  author = {Bellemare, Marc G. and Dabney, Will and Munos, R{\'e}mi},
  year = 2017,
  month = jul,
  pages = {449--458},
  publisher = {PMLR},
  issn = {2640-3498},
  urldate = {2025-04-24},
  langid = {english}
}

@inproceedings{biyik2020active,
  title = {Active {{Preference-Based Gaussian Process Regression}} for {{Reward Learning}}},
  booktitle = {Robotics: {{Science}} and {{Systems XVI}}},
  author = {Biyik, Erdem and Huynh, Nicolas and Kochenderfer, Mykel and Sadigh, Dorsa},
  year = 2020,
  month = jul,
  volume = {16},
  urldate = {2025-11-27},
  isbn = {978-0-9923747-6-1}
}

@inproceedings{biyik2020asking,
  title = {Asking {{Easy Questions}}: {{A User-Friendly Approach}} to {{Active Reward Learning}}},
  shorttitle = {Asking {{Easy Questions}}},
  booktitle = {Proceedings of the {{Conference}} on {{Robot Learning}}},
  author = {B{\i}y{\i}k, Erdem and Palan, Malayandi and Landolfi, Nicholas C. and Losey, Dylan P. and Sadigh, Dorsa},
  year = 2020,
  month = may,
  pages = {1177--1190},
  publisher = {PMLR},
  issn = {2640-3498},
  urldate = {2025-05-13},
  langid = {english}
}

@article{biyik2022learning,
  title = {Learning {{Reward Functions}} from {{Diverse Sources}} of {{Human Feedback}}: {{Optimally Integrating Demonstrations}} and {{Preferences}}},
  shorttitle = {Asking {{Easy}} / {{DemPref}}},
  author = {B{\i}y{\i}k, Erdem and Losey, Dylan P. and Palan, Malayandi and Landolfi, Nicholas C. and Shevchuk, Gleb and Sadigh, Dorsa},
  year = 2022,
  month = jan,
  journal = {The International Journal of Robotics Research},
  volume = {41},
  number = {1},
  pages = {45--67},
  publisher = {SAGE Publications Ltd STM},
  issn = {0278-3649},
  doi = {10.1177/02783649211041652},
  urldate = {2023-10-07},
  langid = {english}
}

@article{biyik2024active,
  title = {Active {{Preference-Based Gaussian Process Regression}} for {{Reward Learning}} and {{Optimization}}},
  author = {B{\i}y{\i}k, Erdem and Huynh, Nicolas and Kochenderfer, Mykel J. and Sadigh, Dorsa},
  year = 2024,
  month = apr,
  journal = {The International Journal of Robotics Research},
  volume = {43},
  number = {5},
  pages = {665--684},
  publisher = {SAGE Publications Ltd STM},
  issn = {0278-3649},
  doi = {10.1177/02783649231208729},
  urldate = {2024-06-12},
  langid = {english}
}

@article{blei2017variational,
  title = {Variational {{Inference}}: {{A Review}} for {{Statisticians}}},
  shorttitle = {{{VI Review}}},
  author = {Blei, David M. and Kucukelbir, Alp and McAuliffe, Jon D.},
  year = 2017,
  month = apr,
  journal = {Journal of the American Statistical Association},
  volume = {112},
  number = {518},
  eprint = {1601.00670},
  pages = {859--877},
  issn = {0162-1459, 1537-274X},
  doi = {10.1080/01621459.2017.1285773},
  urldate = {2021-01-11},
  archiveprefix = {arXiv},
  langid = {english}
}

@article{bobu2020quantifying,
  title = {Quantifying {{Hypothesis Space Misspecification}} in {{Learning From Human}}--{{Robot Demonstrations}} and {{Physical Corrections}}},
  author = {Bobu, Andreea and Bajcsy, Andrea and Fisac, Jaime F. and Deglurkar, Sampada and Dragan, Anca D.},
  year = 2020,
  month = jun,
  journal = {IEEE Transactions on Robotics},
  volume = {36},
  number = {3},
  pages = {835--854},
  issn = {1941-0468},
  doi = {10.1109/TRO.2020.2971415},
  urldate = {2024-06-11}
}

@article{bradley1952rank,
  title = {Rank {{Analysis}} of {{Incomplete Block Designs}}: {{I}}. {{The Method}} of {{Paired Comparisons}}},
  shorttitle = {Rank {{Analysis}} of {{Incomplete Block Designs}}},
  author = {Bradley, Ralph Allan and Terry, Milton E.},
  year = 1952,
  journal = {Biometrika},
  volume = {39},
  number = {3/4},
  eprint = {2334029},
  eprinttype = {jstor},
  pages = {324--345},
  publisher = {[Oxford University Press, Biometrika Trust]},
  issn = {0006-3444},
  doi = {10.2307/2334029},
  urldate = {2025-03-29}
}

@article{brier1950verification,
  title = {Verification of {{Forecasts Expressed}} in {{Terms}} of {{Probability}}},
  author = {Brier, Glenn W.},
  year = 1950,
  month = jan,
  journal = {Monthly Weather Review},
  volume = {78},
  pages = {1},
  publisher = {AMS},
  issn = {0027-0644},
  doi = {10.1175/1520-0493(1950)078<0001:VOFEIT>2.0.CO;2},
  urldate = {2025-09-24}
}

@inproceedings{brown2019extrapolating,
  title = {Extrapolating {{Beyond Suboptimal Demonstrations}} via {{Inverse Reinforcement Learning}} from {{Observations}}},
  shorttitle = {T-{{REX}}},
  booktitle = {Proceedings of the 36th {{International Conference}} on {{Machine Learning}}},
  author = {Brown, Daniel and Goo, Wonjoon and Nagarajan, Prabhat and Niekum, Scott},
  year = 2019,
  month = may,
  pages = {783--792},
  publisher = {PMLR},
  issn = {2640-3498},
  urldate = {2025-05-15},
  langid = {english}
}

@article{brown2020safe,
  title = {Safe {{Imitation Learning}} via {{Fast Bayesian Reward Inference}} from {{Preferences}}},
  shorttitle = {B-{{REX}}},
  author = {Brown, Daniel S. and Coleman, Russell and Srinivasan, R. and Niekum, S.},
  year = 2020,
  month = feb,
  journal = {ArXiv},
  urldate = {2024-06-21}
}

@inproceedings{brunzema2024bayesian,
  title = {Bayesian {{Optimization}} via {{Continual Variational Last Layer Training}}},
  shorttitle = {{{VBLL}} for {{BayesOpt}}},
  booktitle = {The {{Thirteenth International Conference}} on {{Learning Representations}}},
  author = {Brunzema, Paul and Jordahn, Mikkel and Willes, John and Trimpe, Sebastian and Snoek, Jasper and Harrison, James},
  year = 2024,
  month = oct,
  urldate = {2025-02-07},
  langid = {english}
}

@misc{cabezas2024blackjax,
  title = {{{BlackJAX}}: {{Composable Bayesian}} Inference in {{JAX}}},
  shorttitle = {{{BlackJAX}}},
  author = {Cabezas, Alberto and Corenflos, Adrien and Lao, Junpeng and Louf, R{\'e}mi and Carnec, Antoine and Chaudhari, Kaustubh and {Cohn-Gordon}, Reuben and Coullon, Jeremie and Deng, Wei and Duffield, Sam and {Dur{\'a}n-Mart{\'i}n}, Gerardo and Elantkowski, Marcin and {Foreman-Mackey}, Dan and Gregori, Michele and Iguaran, Carlos and Kumar, Ravin and Lysy, Martin and Murphy, Kevin and Orduz, Juan Camilo and Patel, Karm and Wang, Xi and Zinkov, Rob},
  year = 2024,
  month = feb,
  number = {arXiv:2402.10797},
  eprint = {2402.10797},
  primaryclass = {cs},
  publisher = {arXiv},
  doi = {10.48550/arXiv.2402.10797},
  urldate = {2025-02-13},
  archiveprefix = {arXiv}
}

@article{casper2023open,
  title = {Open {{Problems}} and {{Fundamental Limitations}} of {{Reinforcement Learning}} from {{Human Feedback}}},
  author = {Casper, Stephen and Davies, Xander and Shi, Claudia and Gilbert, Thomas Krendl and Scheurer, J{\'e}r{\'e}my and Rando, Javier and Freedman, Rachel and Korbak, Tomek and Lindner, David and Freire, Pedro and Wang, Tony Tong and Marks, Samuel and Segerie, Charbel-Raphael and Carroll, Micah and Peng, Andi and Christoffersen, Phillip J. K. and Damani, Mehul and Slocum, Stewart and Anwar, Usman and Siththaranjan, Anand and Nadeau, Max and Michaud, Eric J. and Pfau, Jacob and Krasheninnikov, Dmitrii and Chen, Xin and Langosco, Lauro and Hase, Peter and Biyik, Erdem and Dragan, Anca and Krueger, David and Sadigh, Dorsa and {Hadfield-Menell}, Dylan},
  year = 2023,
  month = sep,
  journal = {Transactions on Machine Learning Research},
  issn = {2835-8856},
  urldate = {2026-04-05},
  langid = {english}
}

@inproceedings{chen2020randomized,
  title = {Randomized {{Ensembled Double Q-Learning}}: {{Learning Fast Without}} a {{Model}}},
  shorttitle = {{{REDQ}}},
  booktitle = {International {{Conference}} on {{Learning Representations}}},
  author = {Chen, Xinyue and Wang, Che and Zhou, Zijian and Ross, Keith W.},
  year = 2020,
  month = oct,
  urldate = {2025-04-24},
  langid = {english}
}

@inproceedings{christiano2017deep,
  title = {Deep {{Reinforcement Learning}} from {{Human Preferences}}},
  shorttitle = {{{RLHF}}},
  booktitle = {Advances in {{Neural Information Processing Systems}}},
  author = {Christiano, Paul F and Leike, Jan and Brown, Tom and Martic, Miljan and Legg, Shane and Amodei, Dario},
  year = 2017,
  volume = {30},
  publisher = {Curran Associates, Inc.},
  urldate = {2023-10-07}
}

@book{cover2006elements,
  title = {Elements of Information Theory},
  author = {Cover, T. M. and Thomas, Joy A.},
  year = 2006,
  edition = {2nd ed},
  publisher = {Wiley-Interscience},
  address = {Hoboken, N.J},
  isbn = {978-0-471-24195-9},
  lccn = {Q360 .C68 2006}
}

@misc{dangel2025position,
  title = {Position: {{Curvature Matrices Should Be Democratized}} via {{Linear Operators}}},
  shorttitle = {Position},
  author = {Dangel, Felix and Eschenhagen, Runa and Ormaniec, Weronika and Fernandez, Andres and Tatzel, Lukas and Kristiadi, Agustinus},
  year = 2025,
  month = jan,
  number = {arXiv:2501.19183},
  eprint = {2501.19183},
  primaryclass = {cs},
  publisher = {arXiv},
  doi = {10.48550/arXiv.2501.19183},
  urldate = {2025-11-27},
  archiveprefix = {arXiv}
}

@inproceedings{daxberger2021bayesian,
  title = {Bayesian {{Deep Learning}} via {{Subnetwork Inference}}},
  booktitle = {Proceedings of the 38th {{International Conference}} on {{Machine Learning}}},
  author = {Daxberger, Erik and Nalisnick, Eric and Allingham, James U. and Antoran, Javier and {Hernandez-Lobato}, Jose Miguel},
  year = 2021,
  month = jul,
  pages = {2510--2521},
  publisher = {PMLR},
  issn = {2640-3498},
  urldate = {2025-04-07},
  langid = {english}
}

@inproceedings{daxberger2024laplace,
  title = {Laplace Redux -- Effortless {{Bayesian}} Deep Learning},
  booktitle = {Proceedings of the 35th {{International Conference}} on {{Neural Information Processing Systems}}},
  author = {Daxberger, Erik and Kristiadi, Agustinus and Immer, Alexander and Eschenhagen, Runa and Bauer, Matthias and Hennig, Philipp},
  year = 2024,
  month = jun,
  series = {{{NIPS}} '21},
  pages = {20089--20103},
  publisher = {Curran Associates Inc.},
  address = {Red Hook, NY, USA},
  urldate = {2024-08-20},
  isbn = {978-1-7138-4539-3}
}

@misc{deepmind2020jax,
  title = {The {{DeepMind JAX Ecosystem}}},
  author = {{DeepMind} and Babuschkin, Igor and Baumli, Kate and Bell, Alison and Bhupatiraju, Surya and Bruce, Jake and Buchlovsky, Peter and Budden, David and Cai, Trevor and Clark, Aidan and Danihelka, Ivo and Dedieu, Antoine and Fantacci, Claudio and Godwin, Jonathan and Jones, Chris and Hemsley, Ross and Hennigan, Tom and Hessel, Matteo and Hou, Shaobo and Kapturowski, Steven and Keck, Thomas and Kemaev, Iurii and King, Michael and Kunesch, Markus and Martens, Lena and Merzic, Hamza and Mikulik, Vladimir and Norman, Tamara and Papamakarios, George and Quan, John and Ring, Roman and Ruiz, Francisco and Sanchez, Alvaro and Sartran, Laurent and Schneider, Rosalia and Sezener, Eren and Spencer, Stephen and Srinivasan, Srivatsan and Stanojevi{\'c}, Milo{\v s} and Stokowiec, Wojciech and Wang, Luyu and Zhou, Guangyao and Viola, Fabio},
  year = 2020
}

@article{defreitas2000hierarchical,
  title = {Hierarchical Bayesian Models for Regularisation in Sequential Learning},
  author = {{de Freitas}, J.F.G. and Niranjan, M. and Gee, A. H.},
  year = 2000,
  month = apr,
  journal = {Neural Computation},
  volume = {12},
  number = {4},
  pages = {933--953},
  issn = {1530-888X},
  doi = {10.1162/089976600300015655},
  langid = {english},
  pmid = {10770838}
}

@article{degroot1983comparison,
  title = {The {{Comparison}} and {{Evaluation}} of {{Forecasters}}},
  author = {DeGroot, Morris H. and Fienberg, Stephen E.},
  year = 1983,
  journal = {Journal of the Royal Statistical Society. Series D (The Statistician)},
  volume = {32},
  number = {1},
  eprint = {2987588},
  eprinttype = {jstor},
  pages = {12--22},
  publisher = {[Royal Statistical Society, Wiley]},
  issn = {0039-0526},
  doi = {10.2307/2987588},
  urldate = {2025-09-24}
}

@inproceedings{deng2009imagenet,
  title = {{{ImageNet}}: {{A}} Large-Scale Hierarchical Image Database},
  shorttitle = {{{ImageNet}}},
  booktitle = {2009 {{IEEE Conference}} on {{Computer Vision}} and {{Pattern Recognition}}},
  author = {Deng, Jia and Dong, Wei and Socher, Richard and Li, Li-Jia and Li, Kai and {Fei-Fei}, Li},
  year = 2009,
  month = jun,
  pages = {248--255},
  issn = {1063-6919},
  doi = {10.1109/CVPR.2009.5206848},
  urldate = {2025-09-16}
}

@inproceedings{dietterich2000ensemble,
  title = {Ensemble {{Methods}} in {{Machine Learning}}},
  booktitle = {Multiple {{Classifier Systems}}},
  author = {Dietterich, Thomas G.},
  year = 2000,
  pages = {1--15},
  publisher = {Springer},
  address = {Berlin, Heidelberg},
  doi = {10.1007/3-540-45014-9_1},
  isbn = {978-3-540-45014-6},
  langid = {english}
}

@inproceedings{duran-martin2022efficient,
  title = {Efficient {{Online Bayesian Inference}} for {{Neural Bandits}}},
  shorttitle = {{{SubspaceEKF}}},
  booktitle = {Proceedings of {{The}} 25th {{International Conference}} on {{Artificial Intelligence}} and {{Statistics}}},
  author = {{Duran-Martin}, Gerardo and Kara, Aleyna and Murphy, Kevin},
  year = 2022,
  month = may,
  pages = {6002--6021},
  publisher = {PMLR},
  issn = {2640-3498},
  urldate = {2024-05-30},
  langid = {english}
}

@incollection{efron1992bootstrap,
  title = {Bootstrap {{Methods}}: {{Another Look}} at the {{Jackknife}}},
  shorttitle = {Bootstrap {{Methods}}},
  booktitle = {Breakthroughs in {{Statistics}}: {{Methodology}} and {{Distribution}}},
  author = {Efron, Bradley},
  editor = {Kotz, Samuel and Johnson, Norman L.},
  year = 1992,
  pages = {569--593},
  publisher = {Springer},
  address = {New York, NY},
  doi = {10.1007/978-1-4612-4380-9_41},
  urldate = {2025-08-01},
  isbn = {978-1-4612-4380-9},
  langid = {english}
}

@inproceedings{ellis2024generalized,
  title = {A {{Generalized Acquisition Function}} for {{Preference-based Reward Learning}}},
  booktitle = {2024 {{IEEE International Conference}} on {{Robotics}} and {{Automation}} ({{ICRA}})},
  author = {Ellis, Evan and Ghosal, Gaurav R. and Russell, Stuart J. and Dragan, Anca and B{\i}y{\i}k, Erdem},
  year = 2024,
  month = may,
  pages = {2814--2821},
  doi = {10.1109/ICRA57147.2024.10611472},
  urldate = {2025-05-09}
}

@article{elsken2019neural,
  title = {Neural {{Architecture Search}}: {{A Survey}}},
  shorttitle = {Neural {{Architecture Search}}},
  author = {Elsken, Thomas and Metzen, Jan Hendrik and Hutter, Frank},
  year = 2019,
  journal = {Journal of Machine Learning Research},
  volume = {20},
  number = {55},
  pages = {1--21},
  issn = {1533-7928},
  urldate = {2025-03-29}
}

@inproceedings{finn2016guided,
  title = {Guided {{Cost Learning}}: {{Deep Inverse Optimal Control Via Policy Optimization}}},
  shorttitle = {{{GCL}}},
  booktitle = {Proceedings of the 33rd {{International Conference}} on {{International Conference}} on {{Machine Learning}} - {{Volume}} 48},
  author = {Finn, Chelsea and Levine, Sergey and Abbeel, Pieter},
  year = 2016,
  month = jun,
  series = {{{ICML}}'16},
  pages = {49--58},
  publisher = {JMLR.org},
  address = {New York, NY, USA},
  urldate = {2024-04-22}
}

@misc{fort2020deep,
  title = {Deep {{Ensembles}}: {{A Loss Landscape Perspective}}},
  shorttitle = {Deep {{Ensembles}}},
  author = {Fort, Stanislav and Hu, Huiyi and Lakshminarayanan, Balaji},
  year = 2020,
  month = jun,
  number = {arXiv:1912.02757},
  eprint = {1912.02757},
  primaryclass = {stat},
  publisher = {arXiv},
  doi = {10.48550/arXiv.1912.02757},
  urldate = {2025-01-24},
  archiveprefix = {arXiv}
}

@inproceedings{frankle2022lottery,
  title = {The {{Lottery Ticket Hypothesis}}: {{Finding Sparse}}, {{Trainable Neural Networks}}},
  shorttitle = {The {{Lottery Ticket Hypothesis}}},
  booktitle = {International {{Conference}} on {{Learning Representations}}},
  author = {Frankle, Jonathan and Carbin, Michael},
  year = 2022,
  month = feb,
  urldate = {2022-06-24},
  langid = {english}
}

@article{fridovich-keil2020confidenceaware,
  title = {Confidence-{{Aware Motion Prediction}} for {{Real-Time Collision Avoidance}}},
  author = {{Fridovich-Keil}, David and Bajcsy, Andrea and Fisac, Jaime F and Herbert, Sylvia L and Wang, Steven and Dragan, Anca D and Tomlin, Claire J},
  year = 2020,
  month = mar,
  journal = {The International Journal of Robotics Research},
  volume = {39},
  number = {2-3},
  pages = {250--265},
  publisher = {SAGE Publications Ltd STM},
  issn = {0278-3649},
  doi = {10.1177/0278364919859436},
  urldate = {2024-07-26},
  langid = {english}
}

@misc{fu2020d4rl,
  title = {{{D4RL}}: {{Datasets}} for {{Deep Data-Driven Reinforcement Learning}}},
  shorttitle = {{{D4RL}}},
  author = {Fu, Justin and Kumar, Aviral and Nachum, Ofir and Tucker, George and Levine, Sergey},
  year = 2020,
  month = apr,
  urldate = {2021-10-23},
  langid = {english}
}

@inproceedings{gal2016dropout,
  title = {Dropout as a {{Bayesian Approximation}}: {{Representing Model Uncertainty}} in {{Deep Learning}}},
  shorttitle = {Dropout as a {{Bayesian Approximation}}},
  booktitle = {Proceedings of {{The}} 33rd {{International Conference}} on {{Machine Learning}}},
  author = {Gal, Yarin and Ghahramani, Zoubin},
  year = 2016,
  month = jun,
  pages = {1050--1059},
  publisher = {PMLR},
  issn = {1938-7228},
  urldate = {2025-01-31},
  langid = {english}
}

@inproceedings{gao2022scaling,
  title = {Scaling {{Laws}} for {{Reward Model Overoptimization}}},
  booktitle = {International {{Conference}} on {{Machine Learning}}},
  author = {Gao, Leo and Schulman, John and Hilton, Jacob},
  year = 2022,
  month = oct,
  urldate = {2024-10-24}
}

@misc{gholami2021survey,
  title = {A {{Survey}} of {{Quantization Methods}} for {{Efficient Neural Network Inference}}},
  author = {Gholami, Amir and Kim, Sehoon and Dong, Zhen and Yao, Zhewei and Mahoney, Michael W. and Keutzer, Kurt},
  year = 2021,
  month = jun,
  number = {arXiv:2103.13630},
  eprint = {2103.13630},
  primaryclass = {cs},
  publisher = {arXiv},
  doi = {10.48550/arXiv.2103.13630},
  urldate = {2022-09-23},
  archiveprefix = {arXiv}
}

@misc{gleave2022uncertainty,
  title = {Uncertainty {{Estimation}} for {{Language Reward Models}}},
  author = {Gleave, Adam and Irving, Geoffrey},
  year = 2022,
  month = mar,
  number = {arXiv:2203.07472},
  eprint = {2203.07472},
  primaryclass = {cs},
  publisher = {arXiv},
  doi = {10.48550/arXiv.2203.07472},
  urldate = {2025-05-21},
  archiveprefix = {arXiv}
}

@inproceedings{guo2017calibration,
  title = {On {{Calibration}} of {{Modern Neural Networks}}},
  booktitle = {Proceedings of the 34th {{International Conference}} on {{Machine Learning}}},
  author = {Guo, Chuan and Pleiss, Geoff and Sun, Yu and Weinberger, Kilian Q.},
  year = 2017,
  month = jul,
  pages = {1321--1330},
  publisher = {PMLR},
  issn = {2640-3498},
  urldate = {2025-02-03},
  langid = {english}
}

@inproceedings{hadfield-menell2017inverse,
  title = {Inverse {{Reward Design}}},
  booktitle = {Proceedings of the 31st {{International Conference}} on {{Neural Information Processing Systems}}},
  author = {{Hadfield-Menell}, Dylan and Milli, Smitha and Abbeel, Pieter and Russell, Stuart and Dragan, Anca D.},
  year = 2017,
  month = dec,
  series = {{{NIPS}}'17},
  pages = {6768--6777},
  publisher = {Curran Associates Inc.},
  address = {Red Hook, NY, USA},
  urldate = {2025-05-16},
  isbn = {978-1-5108-6096-4}
}

@inproceedings{harrison2023variational,
  title = {Variational {{Bayesian Last Layers}}},
  shorttitle = {{{VBLL}}},
  booktitle = {The {{Twelfth International Conference}} on {{Learning Representations}}},
  author = {Harrison, James and Willes, John and Snoek, Jasper},
  year = 2023,
  month = oct,
  urldate = {2025-02-07},
  langid = {english}
}

@inproceedings{he2016deep,
  title = {Deep {{Residual Learning}} for {{Image Recognition}}},
  shorttitle = {{{ResNet}}},
  booktitle = {Proceedings of the {{IEEE Conference}} on {{Computer Vision}} and {{Pattern Recognition}}},
  author = {He, Kaiming and Zhang, Xiangyu and Ren, Shaoqing and Sun, Jian},
  year = 2016,
  pages = {770--778},
  urldate = {2025-09-16}
}

@article{hennig2012entropy,
  title = {Entropy {{Search}} for {{Information-Efficient Global Optimization}}},
  shorttitle = {Entropy {{Search}}},
  author = {Hennig, Philipp and Schuler, Christian J.},
  year = 2012,
  journal = {Journal of Machine Learning Research},
  volume = {13},
  number = {57},
  pages = {1809--1837},
  issn = {1533-7928},
  urldate = {2025-02-17},
  langid = {english}
}

@inproceedings{hernandez-lobato2014predictive,
  title = {Predictive {{Entropy Search}} for {{Efficient Global Optimization}} of {{Black-box Functions}}},
  booktitle = {Advances in {{Neural Information Processing Systems}}},
  author = {{Hern{\'a}ndez-Lobato}, Jos{\'e} Miguel and Hoffman, Matthew W. and Ghahramani, Zoubin},
  year = 2014,
  volume = {27},
  publisher = {Curran Associates, Inc.},
  urldate = {2025-03-31}
}

@inproceedings{ho2016generative,
  title = {Generative {{Adversarial Imitation Learning}}},
  shorttitle = {{{GAIL}}},
  booktitle = {Advances in {{Neural Information Processing Systems}}},
  author = {Ho, Jonathan and Ermon, Stefano},
  year = 2016,
  volume = {29},
  publisher = {Curran Associates, Inc.},
  urldate = {2023-09-15}
}

@article{hoffman2014nouturn,
  title = {The {{No-U-Turn Sampler}}: {{Adaptively Setting Path Lengths}} in {{Hamiltonian Monte Carlo}}},
  shorttitle = {{{NUTS}}},
  author = {Hoffman, Matthew D. and Gelman, Andrew},
  year = 2014,
  journal = {Journal of Machine Learning Research},
  volume = {15},
  number = {47},
  pages = {1593--1623},
  issn = {1533-7928},
  urldate = {2025-11-27}
}

@inproceedings{hoque2022thriftydagger,
  title = {{{ThriftyDAgger}}: {{Budget-Aware Novelty}} and {{Risk Gating}} for {{Interactive Imitation Learning}}},
  shorttitle = {{{ThriftyDAgger}}},
  booktitle = {Proceedings of the 5th {{Conference}} on {{Robot Learning}}},
  author = {Hoque, Ryan and Balakrishna, Ashwin and Novoseller, Ellen and Wilcox, Albert and Brown, Daniel S. and Goldberg, Ken},
  year = 2022,
  month = jan,
  pages = {598--608},
  publisher = {PMLR},
  issn = {2640-3498},
  urldate = {2025-05-18},
  langid = {english}
}

@misc{houlsby2011bayesian,
  title = {Bayesian {{Active Learning}} for {{Classification}} and {{Preference Learning}}},
  shorttitle = {{{BALD}}},
  author = {Houlsby, Neil and Husz{\'a}r, Ferenc and Ghahramani, Zoubin and Lengyel, M{\'a}t{\'e}},
  year = 2011,
  month = dec,
  number = {arXiv:1112.5745},
  eprint = {1112.5745},
  primaryclass = {stat},
  publisher = {arXiv},
  doi = {10.48550/arXiv.1112.5745},
  urldate = {2025-01-16},
  archiveprefix = {arXiv}
}

@inproceedings{hron2018variational,
  title = {Variational {{Bayesian}} Dropout: Pitfalls and Fixes},
  shorttitle = {Variational {{Bayesian}} Dropout},
  booktitle = {Proceedings of the 35th {{International Conference}} on {{Machine Learning}}},
  author = {Hron, Jiri and Matthews, Alex and Ghahramani, Zoubin},
  year = 2018,
  month = jul,
  pages = {2019--2028},
  publisher = {PMLR},
  issn = {2640-3498},
  urldate = {2025-09-25},
  langid = {english}
}

@misc{hu2021lora,
  title = {{{LoRA}}: {{Low-Rank Adaptation}} of {{Large Language Models}}},
  shorttitle = {{{LoRA}}},
  author = {Hu, Edward J. and Shen, Yelong and Wallis, Phillip and {Allen-Zhu}, Zeyuan and Li, Yuanzhi and Wang, Shean and Wang, Lu and Chen, Weizhu},
  year = 2021,
  month = oct,
  number = {arXiv:2106.09685},
  eprint = {2106.09685},
  primaryclass = {cs},
  publisher = {arXiv},
  doi = {10.48550/arXiv.2106.09685},
  urldate = {2023-09-07},
  archiveprefix = {arXiv}
}

@inproceedings{izmailov2020subspace,
  title = {Subspace {{Inference}} for {{Bayesian Deep Learning}}},
  booktitle = {Proceedings of {{The}} 35th {{Uncertainty}} in {{Artificial Intelligence Conference}}},
  author = {Izmailov, Pavel and Maddox, Wesley J. and Kirichenko, Polina and Garipov, Timur and Vetrov, Dmitry and Wilson, Andrew Gordon},
  year = 2020,
  month = aug,
  pages = {1169--1179},
  publisher = {PMLR},
  issn = {2640-3498},
  urldate = {2025-05-07},
  langid = {english}
}

@inproceedings{izmailov2021what,
  title = {What {{Are Bayesian Neural Network Posteriors Really Like}}?},
  booktitle = {Proceedings of the 38th {{International Conference}} on {{Machine Learning}}},
  author = {Izmailov, Pavel and Vikram, Sharad and Hoffman, Matthew D. and Wilson, Andrew Gordon Gordon},
  year = 2021,
  month = jul,
  pages = {4629--4640},
  publisher = {PMLR},
  issn = {2640-3498},
  urldate = {2025-05-15},
  langid = {english}
}

@misc{jackson2025clean,
  title = {A {{Clean Slate}} for {{Offline Reinforcement Learning}}},
  shorttitle = {Unifloral},
  author = {Jackson, Matthew Thomas and Berdica, Uljad and Liesen, Jarek and Whiteson, Shimon and Foerster, Jakob Nicolaus},
  year = 2025,
  month = apr,
  number = {arXiv:2504.11453},
  eprint = {2504.11453},
  primaryclass = {cs},
  publisher = {arXiv},
  doi = {10.48550/arXiv.2504.11453},
  urldate = {2025-04-23},
  archiveprefix = {arXiv}
}

@misc{jaques2019way,
  title = {Way {{Off-Policy Batch Deep Reinforcement Learning}} of {{Implicit Human Preferences}} in {{Dialog}}},
  shorttitle = {Discrete {{BCQ}}},
  author = {Jaques, Natasha and Ghandeharioun, Asma and Shen, Judy Hanwen and Ferguson, Craig and Lapedriza, Agata and Jones, Noah and Gu, Shixiang and Picard, Rosalind},
  year = 2019,
  month = jul,
  number = {arXiv:1907.00456},
  eprint = {1907.00456},
  primaryclass = {cs},
  publisher = {arXiv},
  doi = {10.48550/arXiv.1907.00456},
  urldate = {2025-04-01},
  archiveprefix = {arXiv}
}

@misc{jax2018github,
  title = {{{JAX}}: Composable Transformations of {{Python}}+{{NumPy}} Programs},
  author = {Bradbury, James and Frostig, Roy and Hawkins, Peter and Johnson, Matthew James and Leary, Chris and Maclaurin, Dougal and Necula, George and Paszke, Adam and VanderPlas, Jake and {Wanderman-Milne}, Skye and Zhang, Qiao},
  year = 2018
}

@article{kingma2014adam,
  title = {Adam: {{A Method}} for {{Stochastic Optimization}}},
  shorttitle = {Adam},
  author = {Kingma, Diederik P. and Ba, Jimmy},
  year = 2014,
  month = dec,
  journal = {CoRR},
  urldate = {2024-11-01}
}

@inproceedings{kostrikov2021offline,
  title = {Offline {{Reinforcement Learning}} with {{Implicit Q-Learning}}},
  shorttitle = {{{IQL}}},
  booktitle = {International {{Conference}} on {{Learning Representations}}},
  author = {Kostrikov, Ilya and Nair, Ashvin and Levine, Sergey},
  year = 2021,
  month = oct,
  urldate = {2025-04-24},
  langid = {english}
}

@inproceedings{kumar2021should,
  title = {When {{Should We Prefer Offline Reinforcement Learning Over Behavioral Cloning}}?},
  booktitle = {International {{Conference}} on {{Learning Representations}}},
  author = {Kumar, Aviral and Hong, Joey and Singh, Anikait and Levine, Sergey},
  year = 2021,
  month = oct,
  urldate = {2025-05-16},
  langid = {english}
}

@inproceedings{lakshminarayanan2016simple,
  title = {Simple and {{Scalable Predictive Uncertainty Estimation}} Using {{Deep Ensembles}}},
  shorttitle = {Deep {{Ensemble}}},
  booktitle = {Neural {{Information Processing Systems}}},
  author = {Lakshminarayanan, Balaji and Pritzel, A. and Blundell, C.},
  year = 2016,
  month = dec,
  urldate = {2024-10-25}
}

@misc{larsen2022how,
  title = {How Many Degrees of Freedom Do We Need to Train Deep Networks: A Loss Landscape Perspective},
  shorttitle = {How Many Degrees of Freedom Do We Need to Train Deep Networks},
  author = {Larsen, Brett W. and Fort, Stanislav and Becker, Nic and Ganguli, Surya},
  year = 2022,
  month = feb,
  number = {arXiv:2107.05802},
  eprint = {2107.05802},
  primaryclass = {cs},
  publisher = {arXiv},
  doi = {10.48550/arXiv.2107.05802},
  urldate = {2025-01-18},
  archiveprefix = {arXiv}
}

@article{lee2021bpref,
  title = {B-{{Pref}}: {{Benchmarking Preference-Based Reinforcement Learning}}},
  shorttitle = {B-{{Pref}}},
  author = {Lee, Kimin and Smith, Laura and Dragan, Anca and Abbeel, Pieter},
  year = 2021,
  month = dec,
  journal = {Proceedings of the Neural Information Processing Systems Track on Datasets and Benchmarks},
  volume = {1},
  urldate = {2024-04-26},
  langid = {english}
}

@inproceedings{lee2021pebble,
  title = {{{PEBBLE}}: {{Feedback-Efficient Interactive Reinforcement Learning}} via {{Relabeling Experience}} and {{Unsupervised Pre-training}}},
  shorttitle = {{{PEBBLE}}},
  booktitle = {Proceedings of the 38th {{International Conference}} on {{Machine Learning}}},
  author = {Lee, Kimin and Smith, Laura M. and Abbeel, Pieter},
  year = 2021,
  month = jul,
  pages = {6152--6163},
  publisher = {PMLR},
  issn = {2640-3498},
  urldate = {2024-05-13},
  langid = {english}
}

@misc{levine2020offline,
  title = {Offline {{Reinforcement Learning}}: {{Tutorial}}, {{Review}}, and {{Perspectives}} on {{Open Problems}}},
  shorttitle = {Offline {{RL Tutorial}}},
  author = {Levine, Sergey and Kumar, Aviral and Tucker, George and Fu, Justin},
  year = 2020,
  month = may,
  urldate = {2021-10-23},
  langid = {english}
}

@inproceedings{li2018measuring,
  title = {Measuring the {{Intrinsic Dimension}} of {{Objective Landscapes}}},
  booktitle = {International {{Conference}} on {{Learning Representations}}},
  author = {Li, Chunyuan and Farkhoor, Heerad and Liu, Rosanne and Yosinski, Jason},
  year = 2018,
  month = feb,
  urldate = {2024-11-04},
  langid = {english}
}

@inproceedings{liang2026robometer,
  title = {Robometer: {{Scaling}} General-Purpose Robotic Reward Models via Trajectory Comparisons},
  shorttitle = {Robometer},
  booktitle = {Proceedings of Robotics: {{Science}} and Systems ({{RSS}})},
  author = {Liang, Anthony and Korkmaz, Yigit and Zhang, Jiahui and Hwang, Minyoung and Anwar, Abrar and Kaushik, Sidhant and Shah, Aditya and Huang, Alex S. and Zettlemoyer, Luke and Fox, Dieter and Xiang, Yu and Li, Anqi and Bobu, Andreea and Gupta, Abhishek and Tu, Stephen and B{\i}y{\i}k, Erdem and Zhang, Jesse},
  year = 2026
}

@article{linderman2025dynamax,
  title = {Dynamax: {{A Python}} Package for Probabilistic State Space Modeling with {{JAX}}},
  shorttitle = {Dynamax},
  author = {Linderman, Scott W. and Chang, Peter and {Harper-Donnelly}, Giles and Kara, Aleyna and Li, Xinglong and {Duran-Martin}, Gerardo and Murphy, Kevin},
  year = 2025,
  month = apr,
  journal = {Journal of Open Source Software},
  volume = {10},
  number = {108},
  pages = {7069},
  issn = {2475-9066},
  doi = {10.21105/joss.07069},
  urldate = {2025-04-03},
  langid = {english}
}

@article{lindley1956measure,
  title = {On a {{Measure}} of the {{Information Provided}} by an {{Experiment}}},
  author = {Lindley, D. V.},
  year = 1956,
  month = dec,
  journal = {The Annals of Mathematical Statistics},
  volume = {27},
  number = {4},
  pages = {986--1005},
  publisher = {Institute of Mathematical Statistics},
  issn = {0003-4851, 2168-8990},
  doi = {10.1214/aoms/1177728069},
  urldate = {2025-05-20}
}

@article{lu2023challenges,
  title = {Challenges and {{Opportunities}} in {{Offline Reinforcement Learning}} from {{Visual Observations}}},
  shorttitle = {Visual {{D4RL}}},
  author = {Lu, Cong and Ball, Philip J. and Rudner, Tim G. J. and {Parker-Holder}, Jack and Osborne, Michael A. and Teh, Yee Whye},
  year = 2023,
  month = apr,
  journal = {Transactions on Machine Learning Research},
  issn = {2835-8856},
  urldate = {2025-05-02},
  langid = {english}
}

@article{mackay1992informationbased,
  title = {Information-{{Based Objective Functions}} for {{Active Data Selection}}},
  author = {MacKay, David J. C.},
  year = 1992,
  month = jul,
  journal = {Neural Computation},
  volume = {4},
  number = {4},
  pages = {590--604},
  issn = {0899-7667},
  doi = {10.1162/neco.1992.4.4.590},
  urldate = {2025-03-31}
}

@misc{mahan2024generative,
  title = {Generative {{Reward Models}}},
  author = {Mahan, Dakota and Phung, Duy Van and Rafailov, Rafael and Blagden, Chase and Lile, Nathan and Castricato, Louis and Fr{\"a}nken, Jan-Philipp and Finn, Chelsea and Albalak, Alon},
  year = 2024,
  month = oct,
  number = {arXiv:2410.12832},
  eprint = {2410.12832},
  publisher = {arXiv},
  doi = {10.48550/arXiv.2410.12832},
  urldate = {2024-11-13},
  archiveprefix = {arXiv}
}

@book{murphy2023probabilistic,
  title = {Probabilistic {{Machine Learning}}: {{Advanced Topics}}},
  shorttitle = {{{PML}} 2},
  author = {Murphy, Kevin P.},
  year = 2023,
  publisher = {The MIT Press},
  address = {Cambridge, Massachusetts},
  isbn = {978-0-262-04843-9},
  langid = {english}
}

@book{murphy2023probabilistica,
  title = {Probabilistic {{Machine Learning}}: {{Advanced Topics Supplments}}},
  shorttitle = {{{PML}} 2 {{Supplement}}},
  author = {Murphy, Kevin P.},
  year = 2023,
  publisher = {The MIT Press},
  address = {Cambridge, Massachusetts},
  isbn = {978-0-262-04843-9},
  langid = {english}
}

@inproceedings{myers2021learning,
  title = {Learning {{Multimodal Rewards}} from {{Rankings}}},
  booktitle = {Conference on {{Robot Learning}}},
  author = {Myers, Vivek and Biyik, Erdem and Anari, Nima and Sadigh, Dorsa},
  year = 2021,
  month = sep,
  urldate = {2024-03-24}
}

@article{naeini2015obtaining,
  title = {Obtaining {{Well Calibrated Probabilities Using Bayesian Binning}}},
  author = {Naeini, Mahdi Pakdaman and Cooper, Gregory and Hauskrecht, Milos},
  year = 2015,
  month = feb,
  journal = {Proceedings of the AAAI Conference on Artificial Intelligence},
  volume = {29},
  number = {1},
  issn = {2374-3468},
  doi = {10.1609/aaai.v29i1.9602},
  urldate = {2025-03-30},
  copyright = {Copyright (c)},
  langid = {english}
}

@incollection{neal2011mcmc,
  title = {{{MCMC Using Hamiltonian Dynamics}}},
  booktitle = {Handbook of {{Markov Chain Monte Carlo}}},
  author = {Neal, Radford M.},
  year = 2011,
  publisher = {{Chapman and Hall/CRC}}
}

@inproceedings{osband2016deep,
  title = {Deep {{Exploration}} via {{Bootstrapped DQN}}},
  shorttitle = {Bootstrapped {{DQn}}},
  booktitle = {Advances in {{Neural Information Processing Systems}}},
  author = {Osband, Ian and Blundell, Charles and Pritzel, Alexander and Van Roy, Benjamin},
  year = 2016,
  volume = {29},
  publisher = {Curran Associates, Inc.},
  urldate = {2025-06-03}
}

@inproceedings{osband2018randomized,
  title = {Randomized {{Prior Functions}} for {{Deep Reinforcement Learning}}},
  booktitle = {Advances in {{Neural Information Processing Systems}}},
  author = {Osband, Ian and Aslanides, John and Cassirer, Albin},
  year = 2018,
  volume = {31},
  publisher = {Curran Associates, Inc.},
  urldate = {2025-06-03}
}

@article{osband2022neural,
  title = {The {{Neural Testbed}}: {{Evaluating Joint Predictions}}},
  shorttitle = {The {{Neural Testbed}}},
  author = {Osband, Ian and Wen, Zheng and Asghari, Seyed Mohammad and Dwaracherla, Vikranth and Lu, Xiuyuan and Ibrahimi, Morteza and Lawson, Dieterich and Hao, Botao and O'Donoghue, Brendan and Van Roy, Benjamin},
  year = 2022,
  month = dec,
  journal = {Advances in Neural Information Processing Systems},
  volume = {35},
  pages = {12554--12565},
  urldate = {2026-01-28},
  langid = {english}
}

@inproceedings{osband2023epistemic,
  title = {Epistemic {{Neural Networks}}},
  shorttitle = {{{EpiNet}}},
  booktitle = {Thirty-Seventh {{Conference}} on {{Neural Information Processing Systems}}},
  author = {Osband, Ian and Wen, Zheng and Asghari, Seyed Mohammad and Dwaracherla, Vikranth and Ibrahimi, Morteza and Lu, Xiuyuan and Roy, Benjamin Van},
  year = 2023,
  month = nov,
  urldate = {2025-03-03},
  langid = {english}
}

@misc{osband2023finetuning,
  title = {Fine-{{Tuning Language Models}} via {{Epistemic Neural Networks}}},
  author = {Osband, Ian and Asghari, Seyed Mohammad and Roy, Benjamin Van and McAleese, Nat and Aslanides, John and Irving, Geoffrey},
  year = 2023,
  month = may,
  number = {arXiv:2211.01568},
  eprint = {2211.01568},
  primaryclass = {cs},
  publisher = {arXiv},
  doi = {10.48550/arXiv.2211.01568},
  urldate = {2025-07-24},
  archiveprefix = {arXiv}
}

@article{ouyang2022training,
  title = {Training Language Models to Follow Instructions with Human Feedback},
  shorttitle = {{{InstructGPT}}},
  author = {Ouyang, Long and Wu, Jeffrey and Jiang, Xu and Almeida, Diogo and Wainwright, Carroll and Mishkin, Pamela and Zhang, Chong and Agarwal, Sandhini and Slama, Katarina and Ray, Alex and Schulman, John and Hilton, Jacob and Kelton, Fraser and Miller, Luke and Simens, Maddie and Askell, Amanda and Welinder, Peter and Christiano, Paul F. and Leike, Jan and Lowe, Ryan},
  year = 2022,
  month = dec,
  journal = {Advances in Neural Information Processing Systems},
  volume = {35},
  pages = {27730--27744},
  urldate = {2024-04-24},
  langid = {english}
}

@inproceedings{ovadia2019can,
  title = {Can {{You Trust Your Model}}'s {{Uncertainty}}? {{Evaluating Predictive Uncertainty Under Dataset Shift}}},
  shorttitle = {Can {{You Trust Your Model}}'s {{Uncertainty}}?},
  booktitle = {Proceedings of the 33rd {{International Conference}} on {{Neural Information Processing Systems}}},
  author = {Ovadia, Yaniv and Fertig, Emily and Ren, Jie and Nado, Zachary and Sculley, D. and Nowozin, Sebastian and Dillon, Joshua V. and Lakshminarayanan, Balaji and Snoek, Jasper},
  year = 2019,
  month = dec,
  pages = {14003--14014},
  publisher = {Curran Associates Inc.},
  address = {Red Hook, NY, USA},
  urldate = {2025-03-30}
}

@inproceedings{pan2021effects,
  title = {The {{Effects}} of {{Reward Misspecification}}: {{Mapping}} and {{Mitigating Misaligned Models}}},
  shorttitle = {The {{Effects}} of {{Reward Misspecification}}},
  booktitle = {International {{Conference}} on {{Learning Representations}}},
  author = {Pan, Alexander and Bhatia, Kush and Steinhardt, Jacob},
  year = 2021,
  month = oct,
  urldate = {2026-03-29},
  langid = {english}
}

@misc{papamarkou2024position,
  title = {Position: {{Bayesian Deep Learning}} Is {{Needed}} in the {{Age}} of {{Large-Scale AI}}},
  shorttitle = {Position},
  author = {Papamarkou, Theodore and Skoularidou, Maria and Palla, Konstantina and Aitchison, Laurence and Arbel, Julyan and Dunson, David and Filippone, Maurizio and Fortuin, Vincent and Hennig, Philipp and {Hern{\'a}ndez-Lobato}, Jos{\'e} Miguel and Hubin, Aliaksandr and Immer, Alexander and Karaletsos, Theofanis and Khan, Mohammad Emtiyaz and Kristiadi, Agustinus and Li, Yingzhen and Mandt, Stephan and Nemeth, Christopher and Osborne, Michael A. and Rudner, Tim G. J. and R{\"u}gamer, David and Teh, Yee Whye and Welling, Max and Wilson, Andrew Gordon and Zhang, Ruqi},
  year = 2024,
  month = jun,
  number = {arXiv:2402.00809},
  eprint = {2402.00809},
  primaryclass = {cs, stat},
  publisher = {arXiv},
  urldate = {2024-07-27},
  archiveprefix = {arXiv},
  langid = {english}
}

@misc{pavlovic2025understanding,
  title = {Understanding {{Model Calibration}} -- {{A}} Gentle Introduction and Visual Exploration of Calibration and the Expected Calibration Error ({{ECE}})},
  author = {Pavlovic, Maja},
  year = 2025,
  month = mar,
  number = {arXiv:2501.19047},
  eprint = {2501.19047},
  primaryclass = {stat},
  publisher = {arXiv},
  doi = {10.48550/arXiv.2501.19047},
  urldate = {2025-03-30},
  archiveprefix = {arXiv}
}

@article{peng2019advantageweighted,
  title = {Advantage-{{Weighted Regression}}: {{Simple}} and {{Scalable Off-Policy Reinforcement Learning}}},
  shorttitle = {{{AWR}}},
  author = {Peng, X. B. and Kumar, Aviral and Zhang, Grace and Levine, S.},
  year = 2019,
  month = oct,
  journal = {ArXiv},
  urldate = {2024-04-26}
}

@misc{poddar2024personalizing,
  title = {Personalizing {{Reinforcement Learning}} from {{Human Feedback}} with {{Variational Preference Learning}}},
  shorttitle = {{{VPL}}},
  author = {Poddar, Sriyash and Wan, Yanming and Ivison, Hamish and Gupta, Abhishek and Jaques, Natasha},
  year = 2024,
  month = aug,
  number = {arXiv:2408.10075},
  eprint = {2408.10075},
  publisher = {arXiv},
  doi = {10.48550/arXiv.2408.10075},
  urldate = {2024-10-22},
  archiveprefix = {arXiv}
}

@inproceedings{rajeswaran2018learning,
  title = {Learning {{Complex Dexterous Manipulation}} with {{Deep Reinforcement Learning}} and {{Demonstrations}}},
  shorttitle = {{{DAPG}}},
  booktitle = {Robotics: {{Science}} and {{Systems XIV}}},
  author = {Rajeswaran, Aravind and Kumar, Vikash and Gupta, Abhishek and Vezzani, Giulia and Schulman, John and Todorov, Emanuel and Levine, Sergey},
  year = 2018,
  month = jun,
  volume = {14},
  urldate = {2025-07-20},
  isbn = {978-0-9923747-4-7}
}

@book{rasmussen2005gaussian,
  title = {Gaussian {{Processes}} for {{Machine Learning}}},
  shorttitle = {{{GPML}}},
  author = {Rasmussen, Carl Edward and Williams, Christopher K. I.},
  year = 2005,
  month = nov,
  publisher = {The MIT Press},
  address = {Cambridge, Mass},
  isbn = {978-0-262-18253-9},
  langid = {english}
}

@misc{razin2025what,
  title = {What {{Makes}} a {{Reward Model}} a {{Good Teacher}}? {{An Optimization Perspective}}},
  shorttitle = {What {{Makes}} a {{Reward Model}} a {{Good Teacher}}?},
  author = {Razin, Noam and Wang, Zixuan and Strauss, Hubert and Wei, Stanley and Lee, Jason D. and Arora, Sanjeev},
  year = 2025,
  month = mar,
  number = {arXiv:2503.15477},
  eprint = {2503.15477},
  primaryclass = {cs},
  publisher = {arXiv},
  doi = {10.48550/arXiv.2503.15477},
  urldate = {2025-04-22},
  archiveprefix = {arXiv}
}

@inproceedings{sadigh2017active,
  title = {Active {{Preference-Based Learning}} of {{Reward Functions}}},
  booktitle = {Robotics: {{Science}} and {{Systems XIII}}},
  author = {Sadigh, Dorsa and Dragan, Anca and Sastry, Shankar and Seshia, Sanjit},
  year = 2017,
  month = jul,
  publisher = {{Robotics: Science and Systems Foundation}},
  doi = {10.15607/RSS.2017.XIII.053},
  urldate = {2023-10-07},
  isbn = {978-0-9923747-3-0}
}

@book{sarkka2023bayesian,
  title = {Bayesian {{Filtering}} and {{Smoothing}}},
  author = {S{\"a}rkk{\"a}, Simo and Svensson, Lennart},
  year = 2023,
  month = jun,
  publisher = {Cambridge University Press},
  googlebooks = {WLe9EAAAQBAJ},
  isbn = {978-1-108-92664-5},
  langid = {english}
}

@techreport{settles2009active,
  type = {Technical {{Report}}},
  title = {Active {{Learning Literature Survey}}},
  author = {Settles, Burr},
  year = 2009,
  institution = {University of Wisconsin-Madison Department of Computer Sciences},
  urldate = {2025-05-13},
  langid = {english}
}

@inproceedings{shen2024variational,
  title = {Variational {{Learning}} Is {{Effective}} for {{Large Deep Networks}}},
  shorttitle = {{{IVON}}},
  booktitle = {Proceedings of the 41st {{International Conference}} on {{Machine Learning}}},
  author = {Shen, Yuesong and Daheim, Nico and Cong, Bai and Nickl, Peter and Marconi, Gian Maria and Raoul, Bazan Clement Emile Marcel and Yokota, Rio and Gurevych, Iryna and Cremers, Daniel and Khan, Mohammad Emtiyaz and M{\"o}llenhoff, Thomas},
  year = 2024,
  month = jul,
  pages = {44665--44686},
  publisher = {PMLR},
  issn = {2640-3498},
  urldate = {2025-04-04},
  langid = {english}
}

@article{shin2022benchmarks,
  title = {Benchmarks and {{Algorithms}} for {{Offline Preference-Based Reward Learning}}},
  shorttitle = {{{OPRL}}},
  author = {Shin, Daniel and Dragan, Anca and Brown, Daniel S.},
  year = 2022,
  month = sep,
  journal = {Transactions on Machine Learning Research},
  issn = {2835-8856},
  urldate = {2025-03-07},
  langid = {english}
}

@inproceedings{singhal1988training,
  title = {Training {{Multilayer Perceptrons}} with the {{Extended Kalman Algorithm}}},
  booktitle = {Advances in {{Neural Information Processing Systems}}},
  author = {Singhal, Sharad and Wu, Lance},
  year = 1988,
  volume = {1},
  publisher = {Morgan-Kaufmann},
  urldate = {2025-01-27}
}

@inproceedings{siththaranjan2023distributional,
  title = {Distributional {{Preference Learning}}: {{Understanding}} and {{Accounting}} for {{Hidden Context}} in {{RLHF}}},
  shorttitle = {{{DPL}}},
  booktitle = {The {{Twelfth International Conference}} on {{Learning Representations}}},
  author = {Siththaranjan, Anand and Laidlaw, Cassidy and {Hadfield-Menell}, Dylan},
  year = 2023,
  month = oct,
  urldate = {2024-11-13},
  langid = {english}
}

@inproceedings{snoek2015scalable,
  title = {Scalable {{Bayesian Optimization Using Deep Neural Networks}}},
  shorttitle = {{{DNGO}}},
  booktitle = {Proceedings of the 32nd {{International Conference}} on {{Machine Learning}}},
  author = {Snoek, Jasper and Rippel, Oren and Swersky, Kevin and Kiros, Ryan and Satish, Nadathur and Sundaram, Narayanan and Patwary, Mostofa and Prabhat, Mr and Adams, Ryan},
  year = 2015,
  month = jun,
  pages = {2171--2180},
  publisher = {PMLR},
  issn = {1938-7228},
  urldate = {2025-01-24},
  langid = {english}
}

@inproceedings{sorensen2024position,
  title = {Position: {{A Roadmap}} to {{Pluralistic Alignment}}},
  shorttitle = {Position},
  booktitle = {Proceedings of the 41st {{International Conference}} on {{Machine Learning}}},
  author = {Sorensen, Taylor and Moore, Jared and Fisher, Jillian and Gordon, Mitchell and Mireshghallah, Niloofar and Rytting, Christopher Michael and Ye, Andre and Jiang, Liwei and Lu, Ximing and Dziri, Nouha and Althoff, Tim and Choi, Yejin},
  year = 2024,
  month = jul,
  series = {{{ICML}}'24},
  volume = {235},
  pages = {46280--46302},
  publisher = {JMLR.org},
  address = {Vienna, Austria},
  urldate = {2025-03-03}
}

@article{srivastava2014dropout,
  title = {Dropout: {{A Simple Way}} to {{Prevent Neural Networks}} from {{Overfitting}}},
  shorttitle = {Dropout},
  author = {Srivastava, Nitish and Hinton, Geoffrey and Krizhevsky, Alex and Sutskever, Ilya and Salakhutdinov, Ruslan},
  year = 2014,
  journal = {Journal of Machine Learning Research},
  volume = {15},
  number = {56},
  pages = {1929--1958},
  issn = {1533-7928},
  urldate = {2025-01-31}
}

@misc{swamy2025all,
  title = {All {{Roads Lead}} to {{Likelihood}}: {{The Value}} of {{Reinforcement Learning}} in {{Fine-Tuning}}},
  shorttitle = {All {{Roads Lead}} to {{Likelihood}}},
  author = {Swamy, Gokul and Choudhury, Sanjiban and Sun, Wen and Wu, Zhiwei Steven and Bagnell, J. Andrew},
  year = 2025,
  month = mar,
  number = {arXiv:2503.01067},
  eprint = {2503.01067},
  primaryclass = {cs},
  publisher = {arXiv},
  doi = {10.48550/arXiv.2503.01067},
  urldate = {2025-05-14},
  archiveprefix = {arXiv}
}

@book{thrun2005probabilistic,
  title = {Probabilistic {{Robotics}}},
  author = {Thrun, Sebastian and Burgard, Wolfram and Fox, Dieter},
  year = 2005,
  month = aug,
  publisher = {MIT Press},
  googlebooks = {2Zn6AQAAQBAJ},
  isbn = {978-0-262-20162-9},
  langid = {english}
}

@inproceedings{tien2022causala,
  title = {Causal {{Confusion}} and {{Reward Misidentification}} in {{Preference-Based Reward Learning}}},
  shorttitle = {Causal {{PbRL}}},
  booktitle = {The {{Eleventh International Conference}} on {{Learning Representations}}},
  author = {Tien, Jeremy and He, Jerry Zhi-Yang and Erickson, Zackory and Dragan, Anca and Brown, Daniel S.},
  year = 2022,
  month = sep,
  urldate = {2026-03-29},
  langid = {english}
}

@inproceedings{todorov2012mujoco,
  title = {{{MuJoCo}}: {{A}} Physics Engine for Model-Based Control},
  shorttitle = {{{MuJoCo}}},
  booktitle = {2012 {{IEEE}}/{{RSJ International Conference}} on {{Intelligent Robots}} and {{Systems}}},
  author = {Todorov, Emanuel and Erez, Tom and Tassa, Yuval},
  year = 2012,
  month = oct,
  pages = {5026--5033},
  issn = {2153-0866},
  doi = {10.1109/IROS.2012.6386109},
  urldate = {2026-07-21}
}

@article{tran2020methods,
  title = {Methods for Comparing Uncertainty Quantifications for Material Property Predictions},
  author = {Tran, Kevin and Neiswanger, Willie and Yoon, Junwoong and Zhang, Qingyang and Xing, Eric and Ulissi, Zachary W},
  year = 2020,
  month = may,
  journal = {Machine Learning: Science and Technology},
  volume = {1},
  number = {2},
  pages = {025006},
  publisher = {IOP Publishing},
  issn = {2632-2153},
  doi = {10.1088/2632-2153/ab7e1a},
  urldate = {2025-03-29},
  langid = {english}
}

@article{virtanen2020scipy,
  title = {{{SciPy}} 1.0: Fundamental Algorithms for Scientific Computing in {{Python}}},
  shorttitle = {{{SciPy}} 1.0},
  author = {Virtanen, Pauli and Gommers, Ralf and Oliphant, Travis E. and Haberland, Matt and Reddy, Tyler and Cournapeau, David and Burovski, Evgeni and Peterson, Pearu and Weckesser, Warren and Bright, Jonathan and {van der Walt}, St{\'e}fan J. and Brett, Matthew and Wilson, Joshua and Millman, K. Jarrod and Mayorov, Nikolay and Nelson, Andrew R. J. and Jones, Eric and Kern, Robert and Larson, Eric and Carey, C. J. and Polat, {\.I}lhan and Feng, Yu and Moore, Eric W. and VanderPlas, Jake and Laxalde, Denis and Perktold, Josef and Cimrman, Robert and Henriksen, Ian and Quintero, E. A. and Harris, Charles R. and Archibald, Anne M. and Ribeiro, Ant{\^o}nio H. and Pedregosa, Fabian and {van Mulbregt}, Paul},
  year = 2020,
  month = mar,
  journal = {Nature Methods},
  volume = {17},
  number = {3},
  pages = {261--272},
  publisher = {Nature Publishing Group},
  issn = {1548-7105},
  doi = {10.1038/s41592-019-0686-2},
  urldate = {2025-11-26},
  copyright = {2020 The Author(s)},
  langid = {english}
}

@misc{weber2025laplax,
  title = {Laplax -- {{Laplace Approximations}} with {{JAX}}},
  shorttitle = {Laplax},
  author = {Weber, Tobias and Mucs{\'a}nyi, B{\'a}lint and Rommel, Lenard and Christie, Thomas and Kas{\"u}schke, Lars and Pf{\"o}rtner, Marvin and Hennig, Philipp},
  year = 2025,
  month = jul,
  number = {arXiv:2507.17013},
  eprint = {2507.17013},
  primaryclass = {cs},
  publisher = {arXiv},
  doi = {10.48550/arXiv.2507.17013},
  urldate = {2025-08-17},
  archiveprefix = {arXiv}
}

@article{wirth2017survey,
  title = {A {{Survey}} of {{Preference-Based Reinforcement Learning Methods}}},
  shorttitle = {{{PbRL Survey}}},
  author = {Wirth, Christian and Akrour, Riad and Neumann, Gerhard and F{\"u}rnkranz, Johannes},
  year = 2017,
  month = jan,
  journal = {The Journal of Machine Learning Research},
  volume = {18},
  number = {1},
  pages = {4945--4990},
  issn = {1532-4435}
}

@misc{yang2024bayesian,
  title = {Bayesian {{Reward Models}} for {{LLM Alignment}}},
  author = {Yang, Adam X. and Robeyns, Maxime and Coste, Thomas and Shi, Zhengyan and Wang, Jun and {Bou-Ammar}, Haitham and Aitchison, Laurence},
  year = 2024,
  month = jul,
  number = {arXiv:2402.13210},
  eprint = {2402.13210},
  primaryclass = {cs},
  publisher = {arXiv},
  doi = {10.48550/arXiv.2402.13210},
  urldate = {2025-01-31},
  archiveprefix = {arXiv}
}

@inproceedings{yang2024rank2reward,
  title = {{{Rank2Reward}}: {{Learning Shaped Reward Functions}} from {{Passive Video}}},
  shorttitle = {{{Rank2Reward}}},
  booktitle = {2024 {{IEEE International Conference}} on {{Robotics}} and {{Automation}} ({{ICRA}})},
  author = {Yang, Daniel and Tjia, Davin and Berg, Jacob and Damen, Dima and Agrawal, Pulkit and Gupta, Abhishek},
  year = 2024,
  month = may,
  pages = {2806--2813},
  doi = {10.1109/ICRA57147.2024.10610873},
  urldate = {2024-10-21}
}

@inproceedings{yuan2023unirlhf,
  title = {Uni-{{RLHF}}: {{Universal Platform}} and {{Benchmark Suite}} for {{Reinforcement Learning}} with {{Diverse Human Feedback}}},
  shorttitle = {Uni-{{RLHF}}},
  booktitle = {The {{Twelfth International Conference}} on {{Learning Representations}}},
  author = {Yuan, Yifu and Hao, Jianye and Ma, Yi and Dong, Zibin and Liang, Hebin and Liu, Jinyi and Feng, Zhixin and Zhao, Kai and Zheng, Yan},
  year = 2023,
  month = oct,
  urldate = {2026-02-27},
  langid = {english}
}

@inproceedings{zhang2024generative,
  title = {Generative {{Verifiers}}: {{Reward Modeling}} as {{Next-Token Prediction}}},
  shorttitle = {Generative {{Verifiers}}},
  booktitle = {The 4th {{Workshop}} on {{Mathematical Reasoning}} and {{AI}} at {{NeurIPS}}'24},
  author = {Zhang, Lunjun and Hosseini, Arian and Bansal, Hritik and Kazemi, Mehran and Kumar, Aviral and Agarwal, Rishabh},
  year = 2024,
  month = oct,
  urldate = {2025-05-08},
  langid = {english}
}

@misc{zhou2024autonomous,
  title = {Autonomous {{Improvement}} of {{Instruction Following Skills}} via {{Foundation Models}}},
  shorttitle = {{{SOAR}}},
  author = {Zhou, Zhiyuan and Atreya, Pranav and Lee, Abraham and Walke, Homer and Mees, Oier and Levine, Sergey},
  year = 2024,
  month = oct,
  number = {arXiv:2407.20635},
  eprint = {2407.20635},
  primaryclass = {cs},
  publisher = {arXiv},
  doi = {10.48550/arXiv.2407.20635},
  urldate = {2026-04-23},
  archiveprefix = {arXiv}
}

\clearpage

\appendix
\setcounter{figure}{0}
\renewcommand{\thefigure}{A.\arabic{figure}}
\renewcommand{\textfraction}{0.15}
\renewcommand{\floatpagefraction}{0.85}
\renewcommand{\topfraction}{0.85}

\section{Technical Appendices and Supplementary Material}\label{appendix:main}
Our code is available in the JAX \citep{jax2018github} framework at \url{https://github.com/yutaizhou/bnn_pref}.
For implementation of the reward learning algorithms, we use Dynamax \citep{linderman2025dynamax} for extended Kalman filtering (EKF), Laplax \citep{weber2025laplax} for Laplace approximation, and Blackjax \citep{cabezas2024blackjax} for MCMC.
For offline RL, we use Unifloral \citep{jackson2025clean} for implementation of implicit Q-learning (IQL). 
All statistical tests are done using SciPy \citep{virtanen2020scipy}. 
Unless stated otherwise, all experiments are done on a single node with 8 NVIDIA RTX A6000 GPUs via SLURM sharding. 

\begin{figure}[tbp]
    \centering
    \includegraphics[width=0.6\linewidth]{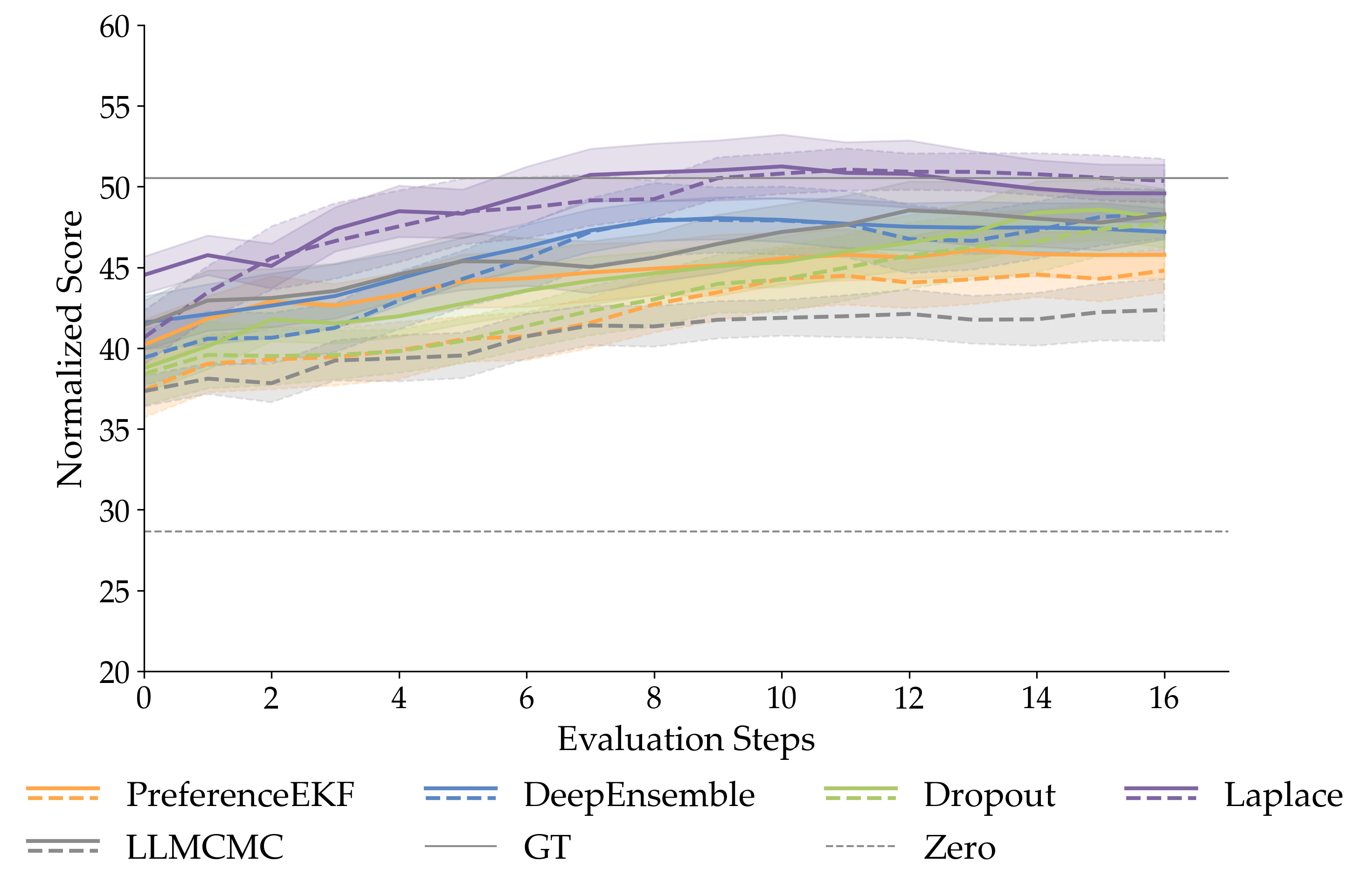}
    \caption{
    Task-aggregate offline RL rollout performance across all tasks for each reward learning algorithm variant, all using the InfoGain acquisition function: comparison of the random (dashed line) and active (solid line) variants of the learned reward models, with rollout performance aggregated over 12 D4RL tasks (mean±s.e. over 5 seeds).
    }
    \label{fig:rlScore_agg}
\end{figure}

For figures that aggregate across tasks and per-task seeds (e.g., \autoref{fig:logpdf_agg}, \autoref{fig:train_duration}, \autoref{fig:calibration}, \autoref{fig:rlScore_agg}), we aggregate as follows: given a dependent variable per step, we pool at each step across 12 tasks $\times ~n$ seeds per task, and plot the per-step mean performance over $12n$ runs along with either standard error or 95\% bootstrap interval for the confidence bounds. 
In our main preference learning results in \autoref{fig:logpdf_agg}, the dependent variable is test-likelihood for preference learning after every step of acquired query label.
We take similar approaches for our calibration results in \autoref{fig:calibration}, where the dependent variable per step is expected calibration error or Brier score. For policy learning results in \autoref{fig:rlScore_agg}, the steps are evaluation rollouts every 40K gradient updates, and the dependent variable is environment rollout return.

\subsection{EKF with Bradley-Terry Likelihood } \label{appendix:ekf_formula}
Here we provide the exact form of the EKF belief update procedure for posterior inference upon receiving a new query, where we use the BT model for the measurement function. For more details, please see chapter 8.3 of \citet{murphy2023probabilistic}. 

For convenience, we first reproduce the general form of the EKF update procedure from \autoref{sec:method}. Using the formulation of sequential Bayesian inference, we perform posterior inference of neural network parameters from streaming data $\mathcal{D}_{1: i-1}=\{(Q_1, y_1), \ldots,(Q_{i-1}, y_{i-1})\}$, where $Q_i=\{\tau_a,\tau_b\}$ is the pairwise preference query and $y_i$ is the binary preference label. Starting from some prior belief $\bel[0]=p(\paramRM)$\footnote{We use $\paramRM$ to broadly refer to model parameters, but for subspace inference specifically, our belief is over $\paramSubspace$.} on the parameters, our posterior after observing $i$ samples can be expressed using Bayes' rule as follows:
\begin{equation}
\begin{aligned}
p(\paramRM_i \mid \dataset_{1:i})
&\propto \underbrace{p(\dataset_{i} \mid \paramRM_{i})}_{\text{Measurement}} \ p(\paramRM_{i} \mid \dataset_{1:i-1}) \\
p(\paramRM_{i} \mid \dataset_{1:i-1})
&= \int \underbrace{p(\paramRM_{i} \mid \paramRM_{i-1})}_{\text{Dynamics}} \ \underbrace{p(\paramRM_{i-1} \mid \dataset_{1:i-1})}_{\text{Previous posterior}} d\paramRM_{i-1}
\label{eq:sequential_bayesian_appendix}
\end{aligned}
\end{equation}
where $p(\paramRM_{i-1} \mid \dataset_{1:i-1})$ is the posterior belief over parameters after observing $i-1$ samples, which is combined with a parameter dynamics model and measurement model to form the posterior after observing the $i^{\textrm{th}}$ example $\dataset_i$. 
We assume additive Gaussian noise for both the dynamics model $p(\paramRM_i \mid \paramRM_{i-1}) = \mathcal{N}(\paramRM_i \mid g(\paramRM_{i-1}), \dynamicsNoise)$ and the measurement model $p(\dataset_i \mid \paramRM_{i}) = \mathcal{N}(y_i \mid h(\paramRM_{i}, Q_i), \measurementNoise)$, where $\dynamicsNoise \in \mathbb{R}^{|\paramRM| \times |\paramRM|}$ and $\measurementNoise \in \mathbb{R}^{|y| \times |y|}$ are prespecified Gaussian noise covariance matrices, and $g: \mathbb{R}^{|\paramRM|} \rightarrow \mathbb{R}^{|\paramRM|}$ and $h: \mathbb{R}^{|\paramRM|} \times \mathbb{R}^{|Q|} \rightarrow \mathbb{R}^{|y|}$ are deterministic dynamics and measurement functions (how neural network model parameters change over time, and the likelihood of observed preference data given current model parameters), respectively.

To apply the above formalism to preference learning of neural network reward models, we model the dynamics using an identity function $g(x)=x$, and the measurements using the BT model $h(\paramRM_{i}, \dataset_i)=p_\theta(y\mid \trajWin, \trajLose)=p_\theta(\tau_a \succ \tau_b)$ computed using the learned RM $\rewardLearned$ (\autoref{eq:bradleyTerry_appendix}):
\begin{align}
p_\theta(y\mid \trajWin, \trajLose)
&=p_\theta(\trajWin \succ \trajLose) =\frac{\exp(\beta \cdot \mathcal{R}_{\paramRM}(\trajWin))}{\exp(\beta \cdot\mathcal{R}_{\paramRM}(\trajWin)) + \exp(\beta \cdot\mathcal{R}_{\paramRM}(\trajLose))}\:. \label{eq:bradleyTerry_appendix}
\end{align}

Assumptions on additive Gaussian noise and nonlinear dynamics and measurement functions make the neural network inference objective in \autoref{eq:sequential_bayesian} solvable in closed-form with the EKF algorithm, where the posterior takes a Gaussian form $\bel[i]=p(\paramRM_i \mid \dataset_{1:i})=\mathcal{N}(\mean_i,\cov_i)$ with mean $\mean_i \in \mathbb{R}^{|\paramRM|}$ and covariance $\cov_i \in \mathbb{R}^{|\paramRM| \times |\paramRM|}$. For belief initialization, we set $\mean_0$ to be the zero vector and $\cov_0$ to be a diagonal matrix.

The EKF algorithm alternates between a belief prediction step and a belief update step to update $\bel[i]=p(\paramRM_i \mid \dataset_{1:i})=\mathcal{N}(\mean_i,\cov_i)$ in light of new data $D_i=\{Q_i, y_i\}$. The predict step is as follows, using the identity function for model parameter dynamics function $g(x)=x$: 
\begin{align}
    \begin{aligned}
    \boldsymbol{\mu}_{i\mid i-1} & =g\left(\boldsymbol{\mu}_{i-1}\right) = \boldsymbol{\mu}_{i-1}\\
    \boldsymbol{\Sigma}_{i \mid i-1} & =\mathbf{G}_i \boldsymbol{\Sigma}_{i-1} \mathbf{G}_i^{\top}+\mathbf{U},
    \end{aligned}
\end{align}
where $\mathbf{G}_i \in \mathbb{R}^{|\paramRM| \times |\paramRM|}$ is the Jacobian matrix of the model dynamics function. In the case of an identity function, $\mathbf{G}_i$ is just an identity matrix. 

The update step is as follows, using the BT likelihood for measurement function $h(\paramRM_{i}, \dataset_i)=p_\theta(y\mid \trajWin, \trajLose)=p_\theta(\tau_a \succ \tau_b)$: 
\begin{align}
    \begin{aligned}
    \hat{\boldsymbol{y}}_i & =h\left(\boldsymbol{\mu}_{i \mid i-1}, D_i\right) \\
    &= p_\theta(y \mid \trajWin, \trajLose) \\
    \mathbf{S}_i & =\mathbf{H}_i \boldsymbol{\Sigma}_{i \mid i-1} \mathbf{H}_i^{\top}+\mathbf{V}_i \\
    \mathbf{K}_i & =\boldsymbol{\Sigma}_{i \mid i-1} \mathbf{H}_i^{\top} \mathbf{S}_i^{-1} \\
    \boldsymbol{\mu}_{i} & =\boldsymbol{\mu}_{i \mid i-1}+\mathbf{K}_i\left(\boldsymbol{y}_i-\hat{\boldsymbol{y}}_i\right) \\
    \boldsymbol{\Sigma}_{i}   &=\boldsymbol{\Sigma}_{i \mid i-1}-\mathbf{K}_i \mathbf{S}_i \mathbf{K}_i^{\top},
    \end{aligned}
\end{align}
where $\mathbf{H}_i \in \mathbb{R}^{|y| \times |\paramRM|}$ is the Jacobian matrix of the measurement function. In the case of the BT likelihood, $|y|=2$ as it is a Bernoulli probability distribution given the return of two trajectories. Each row of $\mathbf{H}_i$ is just the gradient of the probability of preferring the corresponding trajectory over the other with respect to the reward model parameters (or subspace dimension thereof). We obtain both Jacobian matrices via Jax's automatic differentiation capability using the Dynamax library \cite{jax2018github, linderman2025dynamax}.

\subsubsection{On linearization of the Bradley-Terry Likelihood:} 
First recall that we denote $h(\theta_{i},Q_{i}=\{\tau_{a},\tau_{b}\})$ as the EKF measurement function that predicts the probability of a preference label $y_{i}$ for the pairwise query $Q_{i}$ given current reward model (subspace) parameters $\theta_{i}$. We further note that the BT likelihood of preference $\tau_{a} \succ \tau_{b}$ can be defined using the sigmoid function $h(\boldsymbol{\theta})=P_{\boldsymbol{\theta}}\left(\tau_a \succ \tau_b\right)=\sigma\left(r_{\boldsymbol{\theta}}\left(\tau_a\right)-r_{\boldsymbol{\theta}}\left(\tau_b\right)\right)$. To apply EKF, we linearize $h(\theta_{i},Q_{i})$ around the mean of the current model parameter estimate (which we assume is also Gaussian) $\boldsymbol{\mu}_{i \mid i-1}$, which we obtain from EKF’s prediction step. We apply first-order Taylor expansion:
$$
h(\boldsymbol{\theta}) \approx h\left(\boldsymbol{\mu}_{i \mid i-1}\right)+\mathbf{H}_i\left(\boldsymbol{\theta}-\boldsymbol{\mu}_{i \mid i-1}\right)
$$
, where $\mathbf{H}_i \in \mathbb{R}^{|y| \times |\theta|}$ is the Jacobian matrix of the measurement function, which captures the sensitivity of the linearized BT measurement function with respect to the parameters $\boldsymbol{\theta}$. We derive the explicit form of $\mathbf{H_{i}}$ using the chain rule. Recall the derivative of the sigmoid function $\sigma'(x)=\sigma(x)(1-\sigma(x))$, and letting $z=r_{\theta}(\tau_a)-r_{\theta}(\tau_{b})$: 
$$
\mathbf{H}_i=\frac{\partial \sigma(z)}{\partial \boldsymbol{\theta}}=\sigma'(z)\nabla_{\theta}z=\sigma(z)(1-\sigma(z))\left(\nabla_{\boldsymbol{\theta}} r_{\boldsymbol{\theta}}\left(\tau_a\right)-\nabla_{\boldsymbol{\theta}} r_{\boldsymbol{\theta}}\left(\tau_b\right)\right)
$$
We can interpret the term $\sigma'(z)$ as a weighting coefficient for the difference in reward model gradient $\left(\nabla_{\boldsymbol{\theta}} r_{\boldsymbol{\theta}}\left(\tau_a\right)-\nabla_{\boldsymbol{\theta}} r_{\boldsymbol{\theta}}\left(\tau_b\right)\right)$. We note that $\sigma'(z)$ is maximized at ($\max_{z} \sigma'(z)=0.25$) when $z=r_{\theta}(\tau_a)-r_{\theta}(\tau_{b})=0$, i.e., when both pairwise comparison items have the same reward and thus high uncertainty under the BT likelihood as to which item is preferred. Conversely, $\lim_{|z| \to \infty }\sigma'(z)=0$, i.e., when one item has much higher reward than the other and thus strong confidence / low uncertainty under the BT likelihood, the difference in reward model gradient vanishes. In summary, under the linearized measurement model, high reward model uncertainty over the preference label leads to higher value for $\mathbf{H_{i}}$ and thus stronger updates to model parameters, as captured by Kalman gain $\mathbf{K}_i=\mathbf{\Sigma}_{i \mid i-1} \mathbf{H}_i^{\top}\left(\mathbf{H}_i \mathbf{\Sigma}_{i \mid i-1} \mathbf{H}_i^{\top}+\mathbf{V}\right)^{-1}$. With low uncertainty and low $\mathbf{H_{i}}$, Kalman gain $\mathbf{K_{i}}$ tends towards zero, making small or no updates to model parameters. 

\subsubsection{On the locally Gaussian assumption of the Bradley-Terry Likelihood:} 
BT distribution is inherently a Bernoulli distribution, which has variance of $p(1-p)$ that is maximized at $0.25$ when $p=0.5$, i.e. when the model is maximally uncertain about preference label. Under the zero-mean Gaussian noise assumption, we change the uncertainty representation from Bernoulli variance to Gaussian variance, which we specify using constant covariance matrix of $\mathbf{V}=0.07 \cdot \mathbf{I}$.  This roughly perturbs the predicted BT likelihood of preferring $\tau_{a} \succ \tau_{b}$ with probability of 0.07 to account for label error, thus preventing the model from making large updates towards overly confident predictions.

\subsubsection{On EKF hyperparameters:}\label{appendix:ekf_hyperparameter} 
The main hyperparameters of Kalman filters are the dynamics noise covariance $\dynamicsNoise \in \mathbb{R}^{|\paramRM| \times |\paramRM|}$, the measurement noise covariance $\measurementNoise \in \mathbb{R}^{|y| \times |y|}$, and the belief initialization covariance $\priorNoise \in \mathbb{R}^{|\paramRM| \times |\paramRM|}$ in the initial belief $\bel[0]=p(\paramRM_0)=\mathcal{N}(0,\priorNoise)$. 
As the goal of applying Bayesian filters to train neural networks is to enable sequential learning from potentially non-stationary data without overfitting to data it has seen so far, we apply weak parameter regularization by specifying small dynamics noise of $\dynamicsNoise= 0.0001 \cdot \mathbf{I}$; this serves to continuously apply weak perturbation to model parameters so as to prevent overfitting. On the other hand, to enable model learning via posterior updates, we set prior noise to a moderate level of $\priorNoise=0.07\cdot \mathbf{I}$. 

We apply measurement noise to deal with potentially noisy data, which in the domain of learning from pairwise preferences amounts to dealing with flipped preference labels, e.g., among two trajectories, $\trajWin$ is supposed to be the better trajectory, but an annotator mistakenly indicated $\trajLose$ as the preferred item. In our synthetic label experiments, we set about $5\%-10\%$ of our generated labels as flipped for each task. 
We set our measurement noise covariance $\measurementNoise= 0.07 \cdot \mathbf{I}$ which roughly perturbs the predicted BT likelihood of preferring $\trajWin \succ \trajLose$ with probability of 0.07 to account for label error. Since the BT model is inherently a Bernoulli distribution where a correct preference label prediction only requires a predicted probability greater than 50\%, we believe our chosen measurement noise is of appropriate scale.

All noise hyperparameters were swept roughly on a $\log_3$ scale. We found \subEKF's performance to be sensitive to all noise hyperparameters. For example, large $\priorNoise$ and small $\measurementNoise$ would lead to very strong posterior updates, leading to overfitting behavior where test-likelihood would increase for a few queries before steady decline. On the other extreme, small $\priorNoise$ and large $\measurementNoise$ would lead to weak posterior updates, causing underfitting behavior where test-likelihood barely sees any improvement. For $\dynamicsNoise$, we found that values much higher than $\dynamicsNoise= 0.0001 \cdot \mathbf{I}$ prevented model from learning altogether, while tiny values lead to numerical instability.

\subsection{Preference-based Reward Learning} \label{appendix:pref_learning}

\subsubsection{Statistical testing} \label{appendix:stats}
To provide statistical significance to the main claims from \autoref{section:experiments_activelearning}, we conduct hypothesis testing of 1) whether the active variant of each algorithm outperforms its random variant and 2) whether active \subEKF outperforms active variants of other Bayesian deep learning baselines. For the summary statistic of each active reward learning experiment run, we compute the normalized area under curve (AUC) of the log-likelihood plot in \autoref{fig:logpdf_agg}. This measures the rate of improvement for log-likelihood.

Since all runs from \autoref{fig:logpdf_agg} are performed using the same set of 12 random seeds and the same train/test dataset split, we conduct our hypothesis testing using one-sided bootstrap hypothesis test to compare the normalized AUC between two sets of runs. We additionally compute the $95\%$ confidence interval as well as Cohen's $d$ for effect size. 
In the first 5 rows of \autoref{tab:ttest}, we show the performance of active versus random variant of each algorithm. We see that active \ensemble, \laplaceapprox and \llmcmc outperform their random counterparts in normalized AUC with high statistical significance, and \dropout completely fails to do so. While \subEKF outperforms its random counterparts on average according to \autoref{fig:logpdf_agg}, it does so with low statistical significance. We thus conclude that active \subEKF performs on par with its random variant, but is unable to outperform it.

In the last 4 rows of \autoref{tab:ttest}, we show the performance of active \subEKF versus active variant of other baselines. We see that active \subEKF outperforms active variants of all baselines in normalized AUC with high statistical significance, with exception of \llmcmc, where their performances were on par with each other. Compared to \llmcmc, our method requires much less time to perform posterior inference (see \autoref{fig:train_duration}) and does not require storage of all queries seen so far, which we see as major advantages despite similar downstream preference learning performance.

\begin{table*}[tbp]
\centering
\begin{tabular}{lcccc}
\toprule
Test & mean diff & $p$-value & Cohen's $d$ & 95\% CI \\
\midrule
EKF (A vs.\ R)      &  0.01 & 0.077 &  0.59 (medium) & $(0.00,\infty)$ \\
DeepEnsemble (A vs.\ R) &  0.12 & $<0.001$ &  4.93 (large) & $(0.10,\infty)$ \\
Dropout (A vs.\ R)  & -0.02 & 0.825 & -0.36 (small) & $(-0.05,\infty)$ \\
Laplace (A vs.\ R)  &  0.04 & 0.011 &  0.90 (large) & $(0.01,\infty)$ \\
LLMCMC (A vs.\ R)   &  0.03 & $<0.001$ &  1.92 (large) & $(0.02,\infty)$ \\
\hline\hline
EKF vs.\ DeepEnsemble   &  0.05 & $<0.001$ &  2.26 (large) & $(0.04,\infty)$ \\
EKF vs.\ Dropout    &  0.25 & $<0.001$ &  5.23 (large) & $(0.21,\infty)$ \\
EKF vs.\ Laplace    &  0.21 & $<0.001$ &  5.17 (large) & $(0.18,\infty)$ \\
EKF vs.\ LLMCMC     &  0.01 & 0.064 &  0.57 (medium) & $(0.00,\infty)$ \\
\bottomrule
\end{tabular}
\caption{One-sided bootstrap tests comparing active vs.\ random variants of each algorithm, and active EKF vs.\ active variant of other baseline algorithms.}
\label{tab:ttest}
\end{table*}

\subsubsection{Runtime experiments} \label{appendix:runtime_table}

Due to readability of the runtime scale, we provide the table version of \autoref{fig:train_duration} in \autoref{tab:appendix_runtime}.
Similarly, we provide the table version of \autoref{fig:scaling_M_logpdf} in \autoref{tab:appendix_scaling_M_logpdf}.

\begin{table}[tbp]
\centering
\caption{Runtime in minutes. Table version of \autoref{fig:train_duration}.}
\label{tab:appendix_runtime}
\begin{tabular}{lrrrrr}
\toprule
Runtime & PreferenceEKF & DeepEnsemble & Dropout & Laplace & LLMCMC \\
\midrule
Active  & $12.1 \pm 0.1$   & $97.7 \pm 0.1$   & $62.1 \pm 0.0$  & $354.6 \pm 2.0$  & $780.2 \pm 40.2$ \\
Random  & $9.9 \pm 0.1$    & $97.5 \pm 0.7$   & $60.8 \pm 0.2$  & $354.1 \pm 0.7$  & $571.5 \pm 11.6$ \\
\bottomrule
\end{tabular}
\end{table}

\begin{table}[tbp]
\centering
\caption{Final likelihood vs.\ ensemble size $M$. Table version of \autoref{fig:scaling_M_logpdf}.}
\label{tab:appendix_scaling_M_logpdf}
\begin{tabular}{lrrr}
\toprule
Final Likelihood & PreferenceEKF & DeepEnsemble & Dropout \\
\midrule
$M{=}5$   & $-0.220 \pm 0.038$ & $-0.518 \pm 0.055$ & $-0.774 \pm 0.022$ \\
$M{=}15$  & $-0.292 \pm 0.022$ & $-0.538 \pm 0.114$ & $-0.612 \pm 0.138$ \\
$M{=}30$  & $-0.267 \pm 0.020$ & $-0.462 \pm 0.095$ & $-0.704 \pm 0.130$ \\
$M{=}50$  & $-0.249 \pm 0.050$ & $-0.508 \pm 0.082$ & $-0.823 \pm 0.035$ \\
$M{=}75$  & $-0.279 \pm 0.024$ & $-0.477 \pm 0.100$ & $-0.680 \pm 0.220$ \\
$M{=}100$ & $-0.255 \pm 0.018$ & $-0.308 \pm 0.047$ & $-0.721 \pm 0.168$ \\
$M{=}150$ & $-0.223 \pm 0.000$ & $-0.368 \pm 0.068$ & $-0.629 \pm 0.037$ \\
\bottomrule
\end{tabular}
\end{table}

\subsubsection{Implementation details} \label{appendix:ekf_implementation}
Unless otherwise stated, all reward learning experiments are done using subspace dimensionality $|\paramSubspace|=200$,  query budget $B=60$, and partial trajectory of length 50. All neural network reward models are represented using multi-layer perceptrons (MLP) with two hidden layers of 64 units. We apply normalization to all input features. \subEKF and \dropout use $M=100$ model parameter samples to compute the acquisition function and posterior predictive distribution, while \ensemble trains $M=5$ independent networks, each with different weight initialization and randomness for minibatch shuffling.

All tasks use a pool of 150K pairwise partial trajectory queries drawn from the trajectory dataset to perform random or active querying over, and 3000 test queries for log-likelihood evaluation. For generation of noisy-optimal synthetic labels, we apply trajectory return normalization before passing trajectory pairs through the BT model \autorefp{eq:bradleyTerry} to compute the likelihood $p_\theta(\trajWin \succ \trajLose)$. We use temperature parameter of $\beta=7$, resulting in roughly 5-15\% mistaken preference labels per task.

Before the sequential learning phase starting on Line \autoref{algl:for_loop}, all algorithms receive a small dataset consisting of $\tau=8$ query-response pairs for belief initialization, i.e., all algorithms observe a total of $\tau+B=8 + 60=68$ samples. All algorithms run variants of gradient descent (GD) on the warm-up dataset for $420$ optimizer steps. While \subEKF uses SGD with learning rate of 1e-4, momentum of 0.9, and batch size of 1, \ensemble and \dropout use Adam \citep{kingma2014adam} with learning rate of 1e-4 along with default hyperparameters from Optax \citep{deepmind2020jax}, and batch size of 8.

\subEKF constructs the subspace by running SVD on the GD iterates obtained from running SGD on the warmup dataset. We throw away the first $20$ out of the $420$ GD iterates and keep only every other remaining iterate, for a total of $(420-20)/2=200$ iterates. Thus, SVD takes in a model parameter array of shape $(200 \times |\paramRM|)$, and returns a projection matrix $\projMatrix$ of shape $(200 \times |\paramSubspace|)$ by keeping only the top $|\paramSubspace|=200$ principal components. The final GD iterate is used as the full space parameter offset $\paramRMOffset$, which, along with projection matrix $\projMatrix$, is used to transform from the subspace back up to the full space for, e.g. computing predictive distributions as described in \autoref{sec:method}. Finally, \subEKF performs belief initialization \autorefp{algl:belief_init} in the subspace using a zero-mean isotropic Gaussian of dimension $|\paramSubspace|=200$.

On the belief update step (Line \autoref{algl:belief_update}), \subEKF learns from only the most recent query-label pair, while \ensemble and \dropout learn from all data seen so far. 
Note that the specific filtering algorithm we use is the iterated EKF \citep{bell1993iterated}, which repeatedly re-linearizes the measurement model around the estimated posterior. Empirically, we observed better log-likelihood evaluation performance in exchange for marginally extra runtime. We refer to the number of such re-linearization steps on every new sample as $n_{\operatorname{linearize}}$. For further details on iterated EKF, refer to Section 8.3.2.2 of \citet{murphy2023probabilistica}. 
We use $n_{\operatorname{linearize}}=5$ for our experiments, but found that the performance of \subEKF to be relatively robust for this hyperparameter. We detail our choice of prior, dynamics, and observation noise levels in \autoref{appendix:ekf_hyperparameter}.

\textbf{On methods for subspace construction:}
The SVD-based approach and the random projection approach are the two primary methods for neural network subspace construction studied across literature \citep{izmailov2020subspace, larsen2022how}. Our default implementation of \subEKF uses the SVD-based subspace construction method, where we first run SGD on an initial preference labeled dataset, then apply SVD on the SGD iterates to obtain a subspace projection matrix.
We also experimented with using the Adam optimizer instead of SGD to produce the iterates, but found this to lead to poor empirical performance. This is consistent with previous works which found that SGD with a high constant learning rate is crucial to producing parameter iterates with enough variance to construct a subspace effective for optimization and inference \citep{fort2020deep}. We hypothesize that Adam's per-parameter learning rate adaptation scheme results in more performant loss minimization but less varied parameter iterates across the optimization trajectory, thus producing a subspace that does not span the full parameter space enough for effective inference.

As an alternative to the SVD-based subspace construction approach, the projection matrix can be obtained via random projections by computing $\mathbf{A} \in \mathbb{R}^{|\paramRM| \times |\paramSubspace|}$ as a random Gaussian matrix with columns normalized to 1 \citep{li2018measuring}. See \autoref{sec:experiment_subdim} for a study comparing the two approaches. See also \autoref{appendix:no_warmup} for a usage of the random projection method for cases where we don't have access to an initial dataset, thus removing \subEKF's usage of SGD-based initialization altogether.

We additionally note that \subEKF's early performance upon belief initialization, prior to the active learning / random sampling phase, is often much higher compared to all baseline methods. We hypothesize that this is due to \subEKF using SGD only as a means to construct the subspace projection matrix, but the actual belief is initialized as a zero-mean Gaussian in this learned subspace. Compared to methods that rely heavily on SGD such as \ensemble, \dropout, and \laplaceapprox, the subspace approach may simply be less overfitted to the initial dataset.



\subsubsection{Baseline algorithms} \label{appendix:baselines}
The primary tradeoff that Bayesian deep learning (BDL) algorithms are concerned with is the computational tractability and approximation quality of the posterior distribution over model parameters given data $p(\paramRM \mid \dataset)$. We selected \ensemble and \dropout as baselines due to 1) their popularity for representing uncertainty in neural networks and 2) their simplicity in that they only rely on standard neural network training techniques such as SGD and dropout, without any classic Bayesian inference algorithms. We selected \laplaceapprox and \llmcmc as they represent state-of-the-art works in scaling classic inference algorithms to the high-dimensional parameter space of neural network training.

For high-dimensional models such as neural networks, the posterior can be highly multi-modal, which can be difficult to approximate for algorithms that use unimodal distributions (typically Gaussian) such as Laplace approximation and extended Kalman filters. On the other hand, while Markov chain Monte Carlo (MCMC) has been the gold standard for posterior approximation \citep{izmailov2021what}, they are very difficult to scale to large models with many parameters.
As such, many BDL algorithms try to ``be Bayesian'' over only a subset or subspace of model parameters, or rely on ensembling to hopefully reach multiple posterior modes. 
Here we provide a high-level description of the five classes of BDL algorithms we use for our experiments, how they perform belief initialization (Line \autoref{algl:belief_init}) and belief update (Line \autoref{algl:belief_update}), the corresponding implementation details, as well as where they have been used in the reward learning literature.

\textbf{\ensemble} and \textbf{\dropout} are among the most widely-used BDL algorithms for reward modeling and more generally, uncertainty quantification in neural networks \citep{christiano2017deep,gleave2022uncertainty,chen2020randomized,hoque2022thriftydagger, jaques2019way}. They approximate the posterior by relying on randomness (e.g., weight initialization, mini-batch sampling order) to train multiple models and average over their predictions. While \ensemble has the computational burden of actually training multiple neural networks, \dropout masks out a subset of model parameters during training and computes the posterior predictive distribution by averaging predictions from multiple model copies with different weight masks during inference time, thus requiring training of only one model. The idea for both approaches is for the multiple resulting models to act as samples from the posterior distribution. All $M$ models trained under the \ensemble method receive a different stream of mini-batches for training. \dropout uses weight dropout probability of $0.3$ for all experiments, during both training and inference. For both methods, belief initialization is done by running SGD on an initial dataset, and belief update is performed by running SGD on all data seen so far.

\textbf{\laplaceapprox}: While Laplace approximation (LA) has traditionally been used for smaller models in logistic regression and Gaussian process-based regression models \citep{biyik2020active, rasmussen2005gaussian}, recent advancements such as those in \cite{dangel2025position, daxberger2024laplace} have made the technique highly scalable to neural network architectures. Combined with parameter-efficient fine-tuning techniques such as LoRA \citep{hu2021lora}, LA has even been applied to transformer-scaled reward models \citep{yang2024bayesian}. By approximating likelihood curvature around a model solution trained via maximum likelihood methods such as gradient descent, LA constructs a local Gaussian approximation to the model posterior. We use the full curvature approximation-based approach of \cite{weber2025laplax} to perform LA over the entire reward model, with prior precision value of 1000. Once the curvature information has been constructed for the Gaussian posterior approximation, we can sample an arbitrary number of model parameters from the posterior. Both belief initialization and belief update are done by first running SGD on all data seen so far, then performing LA on the final SGD iterate.

\textbf{\llmcmc}: Despite the high quality posterior approximation of MCMC methods for smaller models such as linear models \citep{biyik2020asking,hadfield-menell2017inverse}, they are not widely used for neural network posterior inference due to their poor scalability to parameter count. Most applications of MCMC to BDL train the entire NN model using more efficient maximum likelihood methods like gradient descent, then perform MCMC only over the parameters of the final layer. We chose this ``last-layer Bayesian'' approach as it has been shown to strike a good balance between computational tractability and approximation quality \citep{brown2020safe, snoek2015scalable}. The specific MCMC sampler we use is NUTS \citep{hoffman2014nouturn}. On each active learning step, we construct a new log-density function using the aggregated dataset using all samples seen so far. For belief initialization, we use 500 warm-up MCMC iterations followed by 500 additional iterations. For belief update steps, since the log-density function should not differ too much with one additional aggregated sample, we set warm-up iterations to be 20, followed by 500 additional iterations. We then subsample $M$ models from the resulting MCMC iterates to form our sampling-based posterior.

\textbf{\subEKF}: While the preceding described methods perform optimization and inference over either the full model parameter set or a subset thereof, \subEKF finds a low-dimensional subspace (as opposed to just a subset of the parameters) within the full parameter space, and performs inference within the subspace. The main insight of subspace inference approaches \citep{daxberger2021bayesian} is that due to the overparameterized nature of neural networks, capturing posterior information only over a constrained subspace would be a sufficient alternative to posterior inference over the whole network. Once a Gaussian approximation is obtained via subspace Kalman filtering, we can sample an arbitrary number of model parameters from the posterior.

\subsubsection{Acquisition functions} \label{appendix:acq}
The InfoGain acquisition function introduced in \autoref{eq:infogain} was developed by \citet{biyik2020asking} for active reward learning using linear reward models. To motivate its origin, we first express the InfoGain objective in three equivalent forms below due to symmetry of mutual information. 
\begin{subequations}
\begin{align}
Q_i^* 
& =\underset{Q_i}{\arg \max }~ I\left(\paramRM ; y_i \mid Q_i, \bel[i-1]\right) \label{eq_appendix:infogain}\\
& =\underset{Q_i}{\arg \max }~ H\left(\paramRM \mid Q_i, \bel[i-1]\right)- \E_{y_i} \left[H(\paramRM \mid y_i, Q_i, \bel[i-1])\right] \label{eq_appendix:infogain_es}\\
& =\underset{Q_i}{\arg \max }~ H\left(y_i \mid Q_i, \bel[i-1]\right)-\E_{\paramRM} \left[H(y_i \mid \paramRM, Q_i)\right], \label{eq_appendix:infogain_pes}
\end{align}
\end{subequations}

where $\bel[i-1]=p(\paramRM \mid \datasetQuery_{1:i-1})$ is the posterior distribution over RM parameters after learning from $(i-1)$ queries. The idea of mutual information-based acquisition functions is rooted in the concept of expected information gain studied in Bayesian optimal experiment design and active data selection \citep{mackay1992informationbased, lindley1956measure}. 
It was later extended to Bayesian optimization using Gaussian process models under the methods Bayesian active learning by disagreement (BALD) \citep{houlsby2011bayesian}, entropy search (ES) \citep{hennig2012entropy}, and predictive entropy search (PES) \citep{hernandez-lobato2014predictive}. 
In particular, the mutual information objective function in \autoref{eq_appendix:infogain} is expressed in its ES form in \autoref{eq_appendix:infogain_es}, and expressed in its equivalent but computationally efficient PES form in \autoref{eq_appendix:infogain_pes}. 

Our \subEKF method focuses on efficient sampling of high-dimensional neural network model parameters to approximate the predictive distribution for optimizing \autoref{eq_appendix:infogain_pes}, which we derive as follows. We refer to Section 5 of \citet{biyik2020asking} for further interpretations of the objective, and Section 9.1 of their work for derivation of the sampling-based approximation shown in \autoref{eq:infogain_sample}.
\begin{align*}
Q_i^* & = \underset{Q_i}{\arg \max } \ I\left(\boldsymbol{\theta} ; y_i \mid Q_i, \boldsymbol{b}^{i-1}\right) \\
& = \underset{Q_i}{\arg \max } \  H\left(y_i \mid Q_i, \boldsymbol{b}^{i-1}\right) - H\left(y_i \mid \boldsymbol{\theta}, Q_i, \boldsymbol{b}^{i-1}\right)  \\
& = \underset{Q_i}{\arg \max } \ H\left(y_i \mid Q_i, \boldsymbol{b}^{i-1}\right) - \mathbb{E}_{\boldsymbol{\theta} \sim p(\boldsymbol{\theta} \mid \boldsymbol{b}^{i-1})}\left[H\left(y_i \mid \boldsymbol{\theta}, Q_i\right)\right]  \\
& =\underset{Q_i}{\arg \max }~ H\left(y_i \mid Q_i, \bel[i-1]\right)-\E_{\paramRM} \left[H(y_i \mid \paramRM, Q_i)\right]
\end{align*}

\begin{figure*}[tbp]
    \centering
    \includegraphics[width=0.85\textwidth]{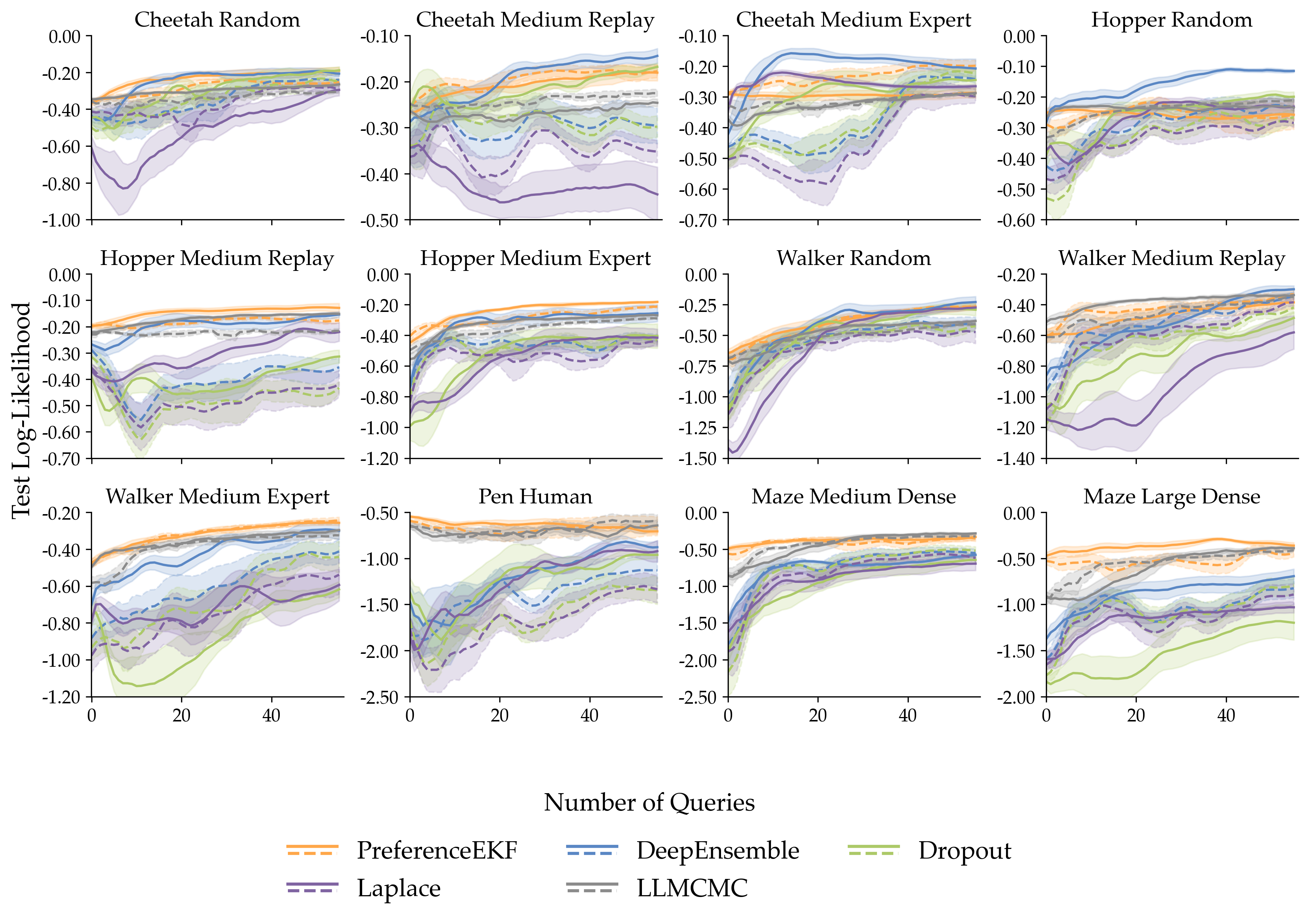}
    \caption{Per-task reward learning performance using InfoGain acquisition function: comparison of the random (dashed line) and active (solid line) variants of the algorithms across 12 D4RL tasks (mean$\pm$s.e. over 12 seeds). In all tasks, active \subEKF either performs on par with or outperforms other algorithms in terms of sample-efficiency and final log-likelihood.}
    \label{fig:logpdf_all}
\end{figure*}

Although our main experiments all use the InfoGain acquisition function to showcase the advantage of being able to sample from high-dimensional neural network parameter distributions, the \subEKF method is agnostic to the acquisition function used for active learning. 
While \autoref{fig:logpdf_agg} and \autoref{fig:logpdf_all} showcase the aggregate and per-task log-likelihood results for active preference-based reward learning experiments using InfoGain, here we show additional results using two more commonly-used acquisition functions, disagreement and entropy. Disagreement selects the query $Q_i$ for which the predicted preference label $\mathbbm{1} (\trajWin^i \succ\trajLose^i)$ has the highest variance across the ensemble or sampled models, and entropy selects the queries for which the Bradley-Terry posterior predictive distribution exhibits the highest entropy. 

\begin{figure*}[tbp]
    \centering
    \begin{subfigure}{0.49\textwidth}
        \centering
        \includegraphics[width=\textwidth]{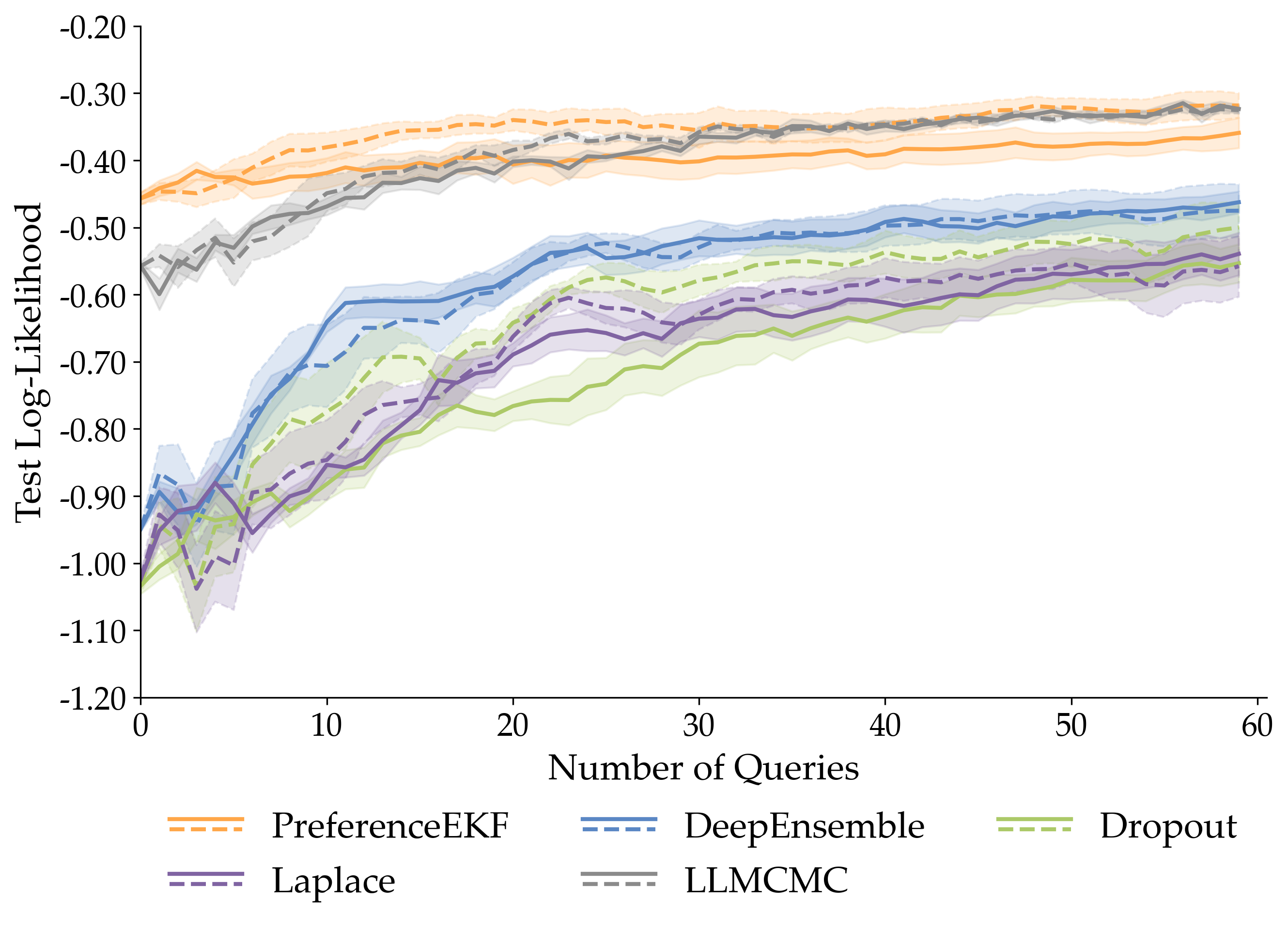}
        \caption{}
        \label{fig:disagreement_logpdf_agg}
    \end{subfigure}
    \hfill
    \begin{subfigure}{0.49\textwidth}
        \centering
        \includegraphics[width=\textwidth]{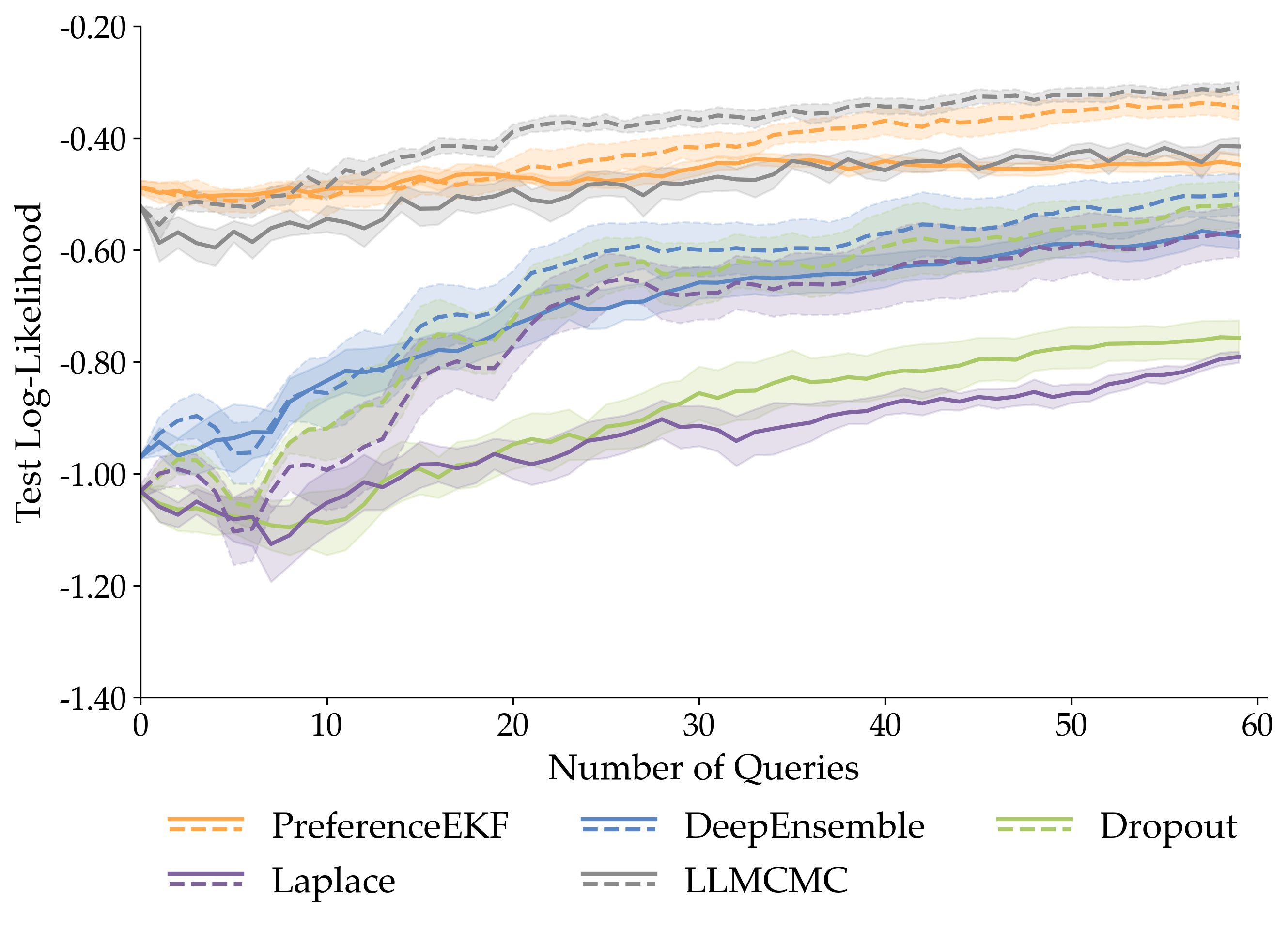}
        \caption{}
        \label{fig:entropy_logpdf_agg}
    \end{subfigure}
    \caption{%
    \autoref{fig:disagreement_logpdf_agg} shows log-likelihood comparison of the random (dashed line) and active (solid line) variants of each algorithm using the disagreement acquisition function (higher means better fitting of annotator preference distribution).
    \autoref{fig:entropy_logpdf_agg} shows the results for entropy-based acquisition function.
    }
    \label{fig:disagreemtn_and_entropy_logpdf_agg}
\end{figure*}

\begin{figure*}[tbp]
    \centering
    \includegraphics[width=0.85\linewidth]{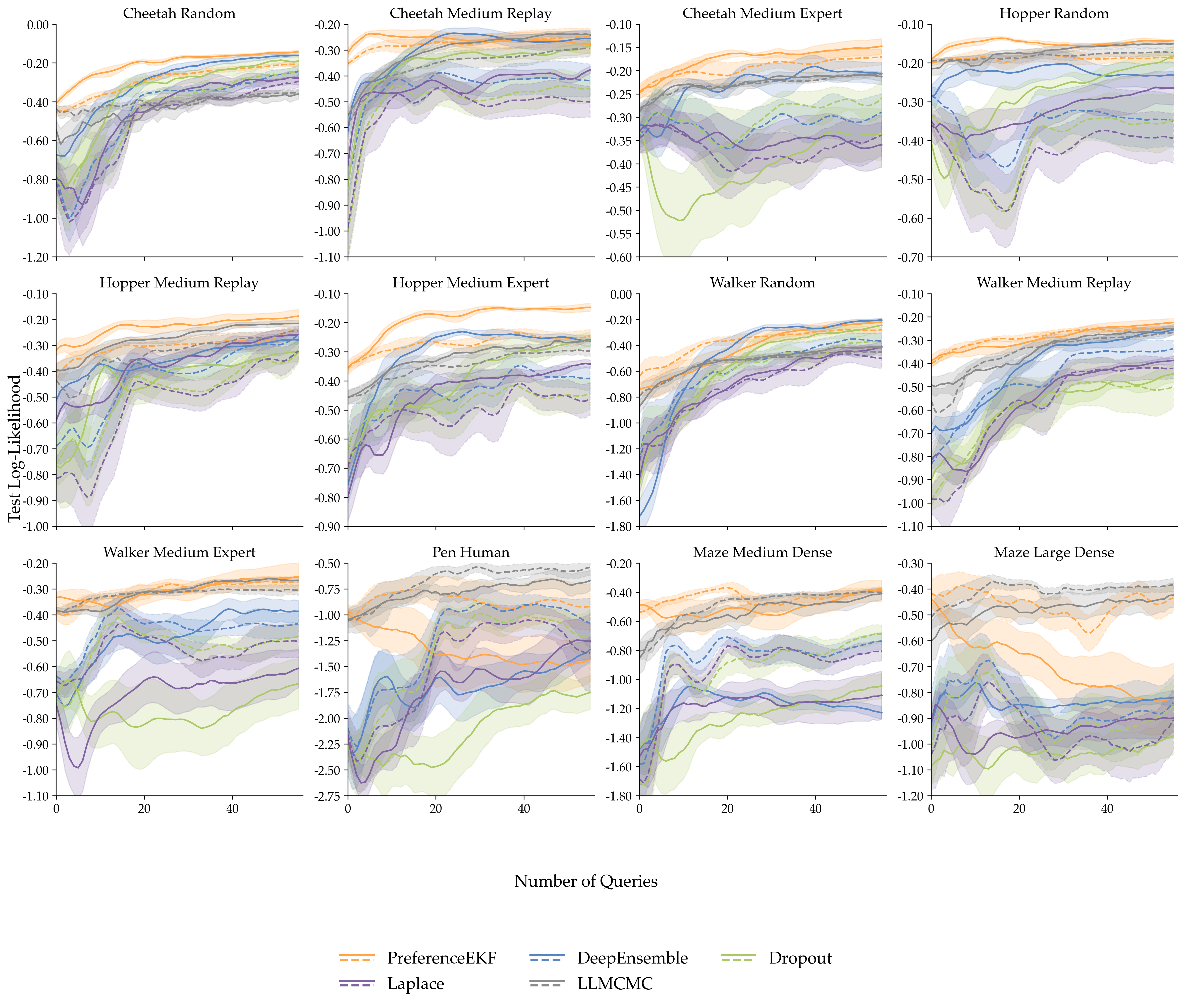}
    \caption{
    Per-task reward learning performance using the disagreement acquisition function: comparison of the random (dashed line) and active (solid line) variants of the algorithms across 12 D4RL tasks for preference-based reward modeling (mean±s.e. over 12 seeds). In most tasks, active \subEKF either performs on par with or outperforms other algorithms in terms of sample-efficiency and final log-likelihood. Pen Human and Maze Large Dense are particular outlier cases where active \subEKF severely underperforms, which explains why the aggregate results in \autoref{fig:disagreement_logpdf_agg} look unfavorably for active \subEKF relative to its random variant.
    }
    \label{fig:disagreement_logpdf}
\end{figure*}

\begin{figure*}[tbp]
    \centering
    \includegraphics[width=0.85\linewidth]{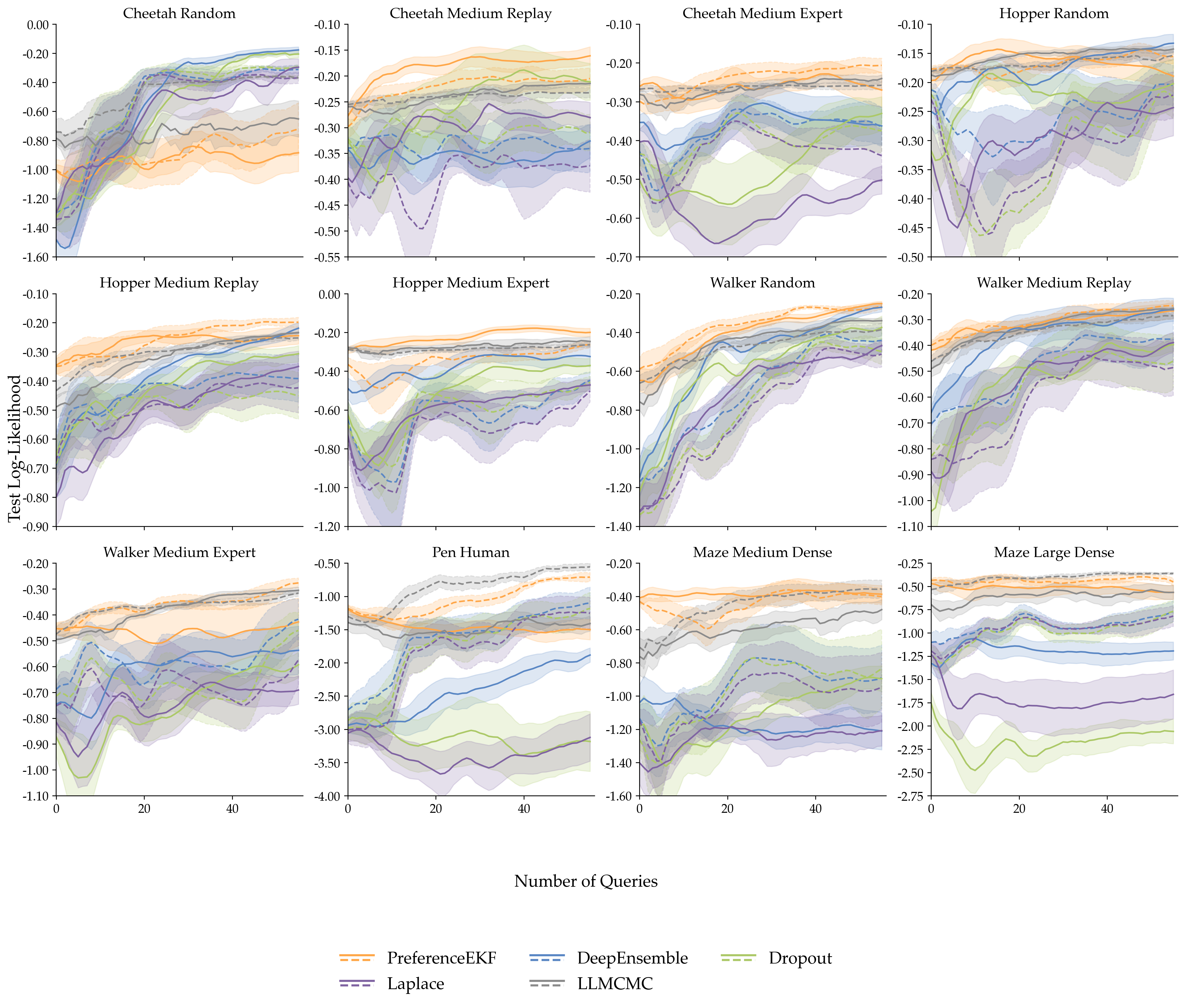}
    \caption{
    Per-task reward learning performance using the entropy acquisition function: comparison of the random (dashed line) and active (solid line) variants of the algorithms across 12 D4RL tasks for preference-based reward modeling (mean±s.e. over 12 seeds). 
    }
    \label{fig:entropy_logpdf}
\end{figure*}

We show in \autoref{fig:disagreemtn_and_entropy_logpdf_agg} that both disagreement and entropy acquisition functions resulted in similar trends, where although \subEKF and \llmcmc perform the best overall, neither algorithm's active learning variant outperformed their random counterpart. This is in contrast to the InfoGain acquisition function result in \autoref{fig:logpdf_agg}, where all algorithms' active variants outperformed their random variants. This demonstrates that while \subEKF and \llmcmc prove to be the most effective at learning a posterior for fitting the annotator's preference distribution, the choice of acquisition function still matters greatly for sample-efficient active learning, with InfoGain being the best performing acquisition function overall, followed by disagreement and then entropy. We further show per-task preference-learning results for InfoGain, disagreement, and entropy in \autoref{fig:logpdf_all}, \autoref{fig:disagreement_logpdf}, and \autoref{fig:entropy_logpdf}, respectively.

\subsubsection{Reward learning using same number of model samples} \label{appendix:same_nmodels}
In our main experiment results, while \ensemble trains $M=5$ separate reward models and uses them to approximate the posterior, all other algorithms explicitly learn a posterior distribution over model parameters, and can thus sample an arbitrary number of model parameters for computing acquisition functions and making predictions; our experiments used $M=100$. 

\begin{figure}[tbp]
    \centering
    \includegraphics[width=0.6\linewidth]{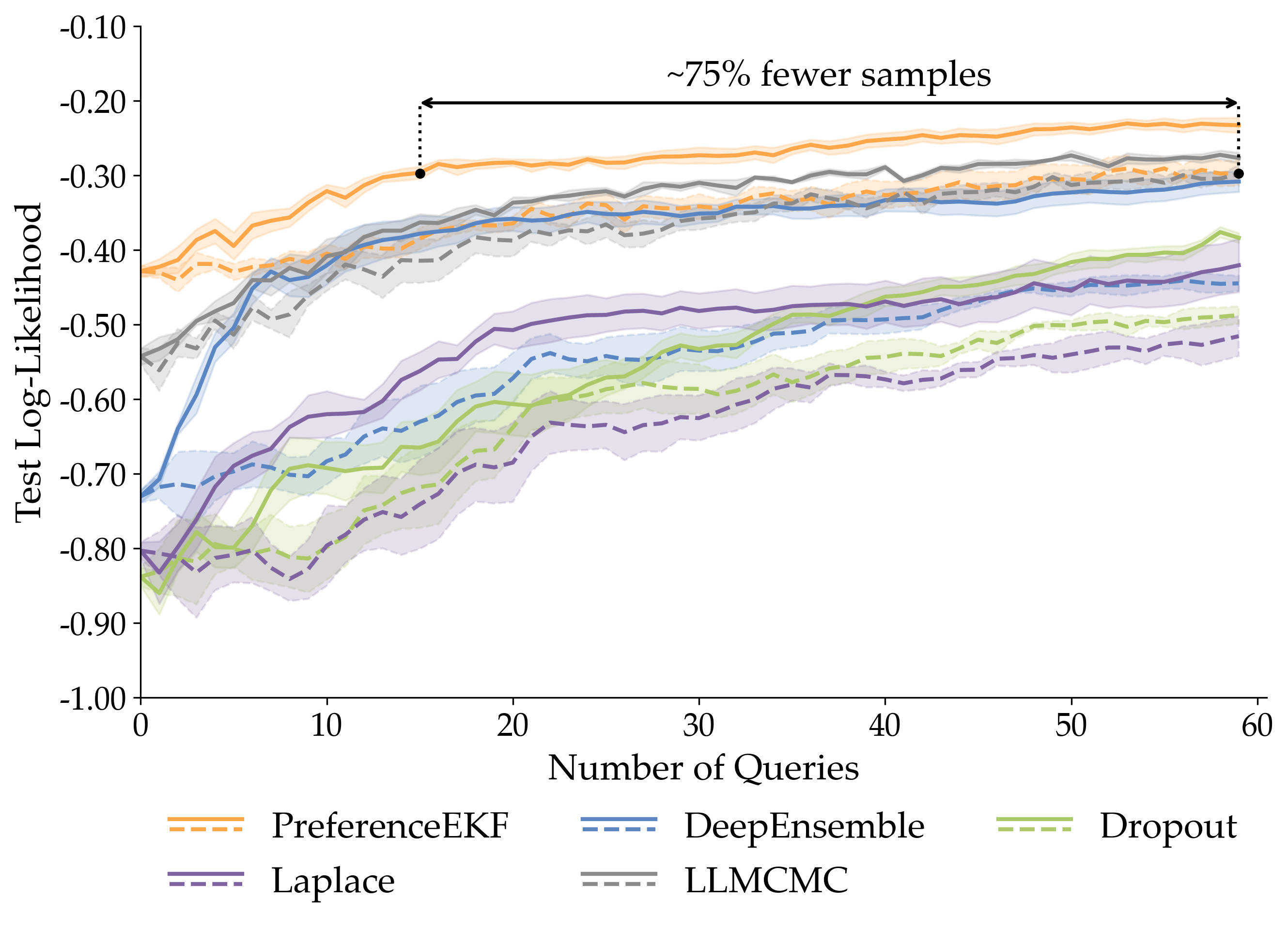}
    \caption{
    Task-aggregate reward learning performance with all algorithms using the same number of model samples ($M=5$). Runs are averaged over all 12 tasks of 5 seeds each.
    }
    \label{fig:sameM}
\end{figure}

This raises the question of whether \subEKF's superior preference learning sample efficiency is solely due to the larger number of posterior samples, or whether the learned posterior indeed captures the annotator's preference. To investigate this, we set $M=5$ for all algorithms and see in \autoref{fig:sameM} that \subEKF still outperforms all methods in test log-likelihood. This signals that higher model sample count is not the only factor that can explain \subEKF's superior sample efficiency, and that the subspace approach for uncertainty representation indeed results in a learned posterior that captures the annotator's preferences well.

\subsubsection{Training runtime under the same compute budget} \label{appendix:runtime}
\autoref{sec:experiment_scale} showed that \subEKF had much faster training runtime compared to the other methods. The setting was biased to be favorable towards \subEKF, as it only needs to perform one belief update step on the most recent query (thanks to its sequential learning nature), whereas all the other methods need to perform multiple gradient descent updates on all data seen so far so as to ensure that SGD converges to a maximum likelihood solution. 

\begin{figure*}[tbp]
    \centering
    \begin{subfigure}{0.49\textwidth}
        \centering
        \includegraphics[width=\textwidth]{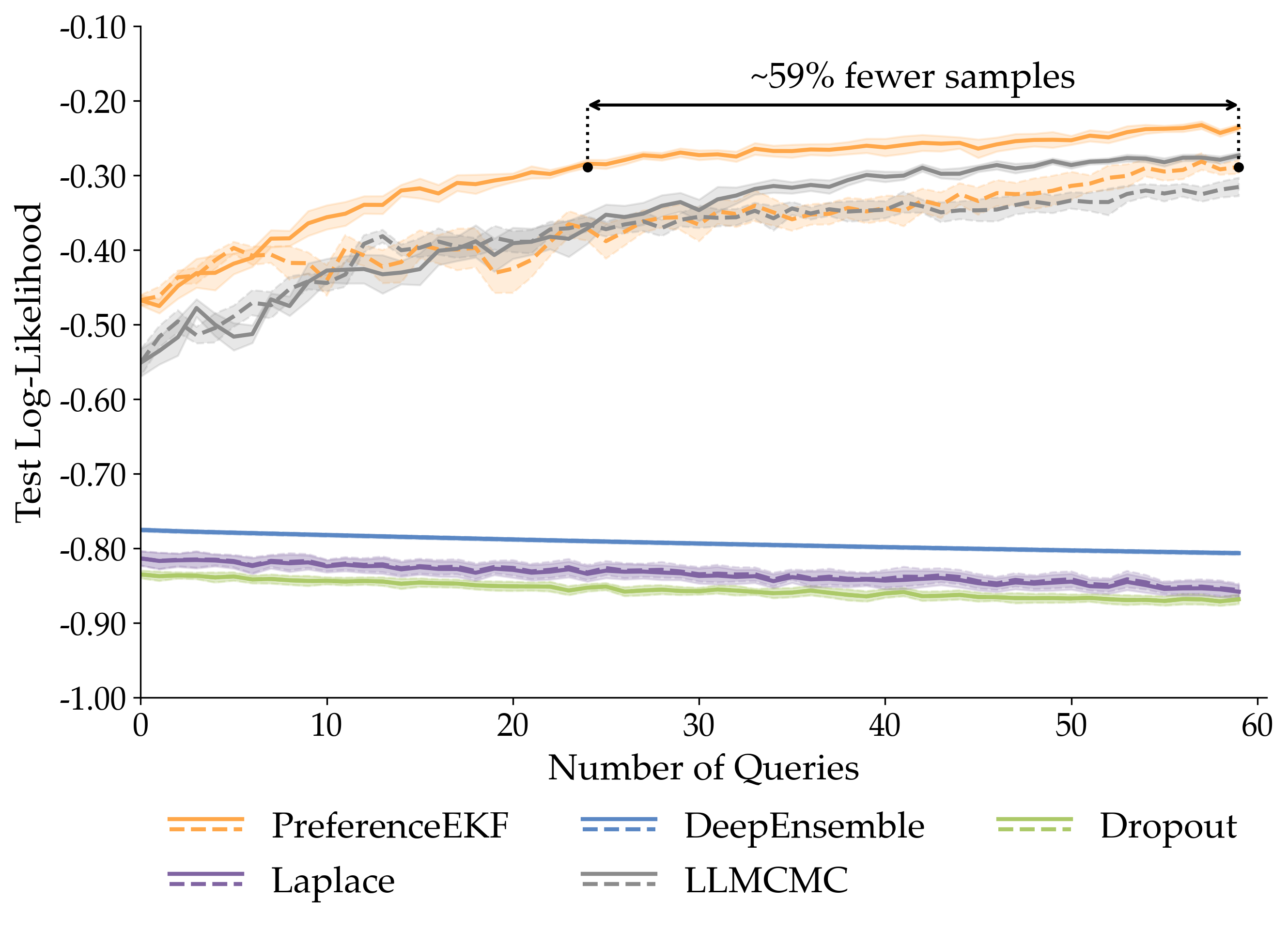}
        \caption{}
        \label{fig:sameUpdateLogpdf}
    \end{subfigure}
    \begin{subfigure}{0.49\textwidth}
        \centering
        \includegraphics[width=\textwidth]{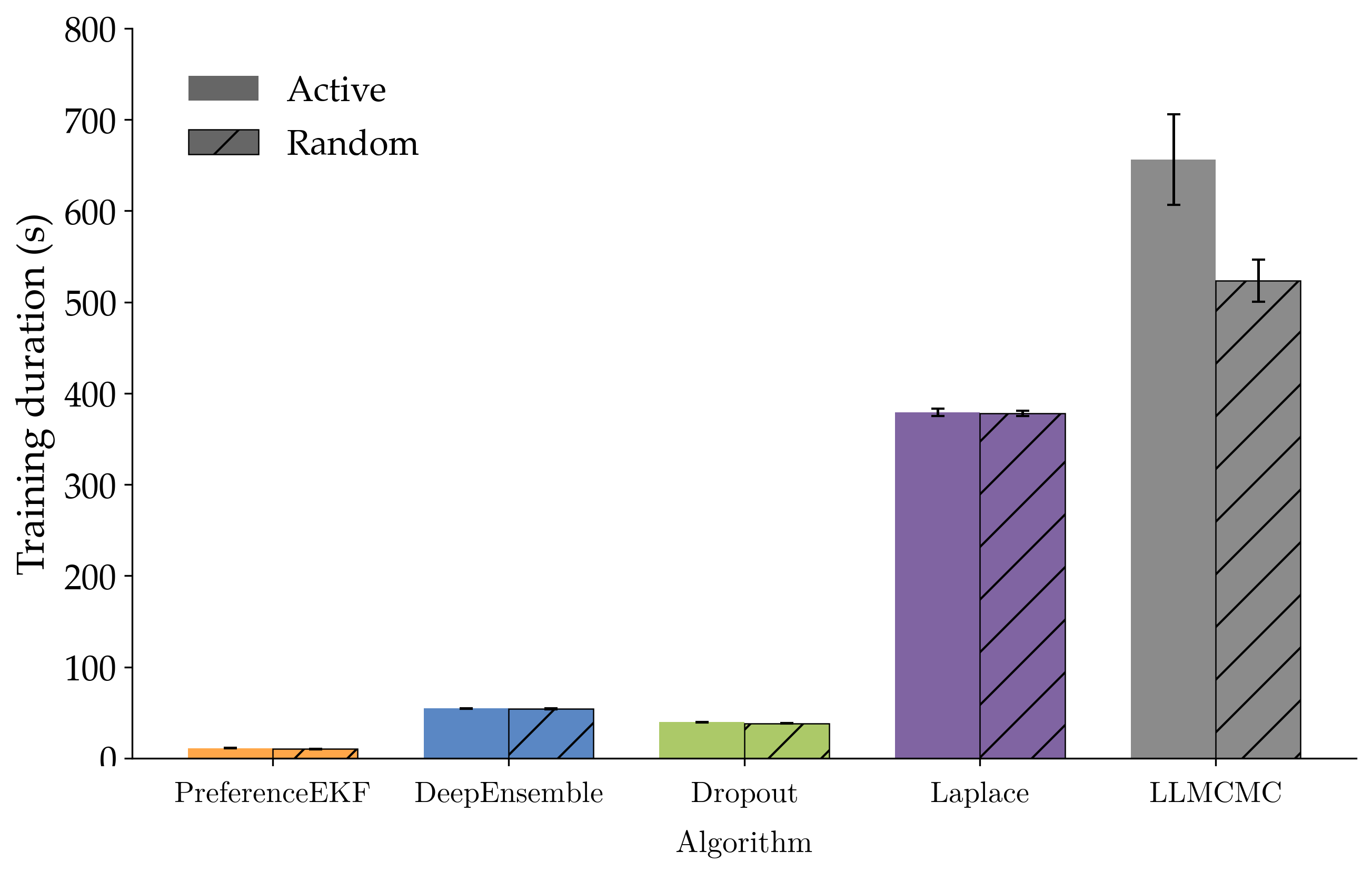}
        \caption{}
        \label{fig:sameUpdateDuration}
    \end{subfigure}
    \caption{
    \autoref{fig:sameUpdateLogpdf} shows log-likelihood comparison of the random (dashed line) and active (solid line) variants of each algorithm using the InfoGain acquisition function, with reduced belief update compute budget. We see that all methods that rely heavily on SGD fail to learn, while \subEKF and \llmcmc retain good performance.
    \autoref{fig:sameUpdateDuration} shows training runtime duration of both active and random variants of each algorithm.
    Each line plot and bar plot is aggregated over 12 D4RL tasks (mean$\pm$s.e.\ over 12 seeds).%
    }
    \label{fig:sameUpdate}
\end{figure*}

To ensure a more fair runtime comparison, we set the number of SGD iterations for belief update of all baselines to match the compute budget of EKF's belief update step, and additionally set the number of model samples for all methods to be $M=5$ to match that of \ensemble. We show in \autoref{fig:sameUpdateLogpdf} that under the setting with reduced number of SGD steps, the methods that rely on SGD (\ensemble, \dropout, \laplaceapprox) failed to converge, and were thus unable to fit the preference distribution as indicated by low log-likelihood. \subEKF and \llmcmc were still able to fit the preference distribution, with the former still retaining a clear lead in having the fastest training runtime as shown in \autoref{fig:sameUpdateDuration}.

\subsubsection{Reward learning without an initial dataset} \label{appendix:no_warmup}
In \autoref{sec:method}, it was shown that subspace construction for \subEKF can be done via either SVD on SGD iterates trained on an initial dataset, or random projections. \autoref{alg:subspaceEKF} further indicates that regardless of the subspace construction method, \subEKF still relies on access to an initial dataset of already labeled queries.

\begin{figure}[tbp]
    \centering
    \includegraphics[width=0.6\linewidth]{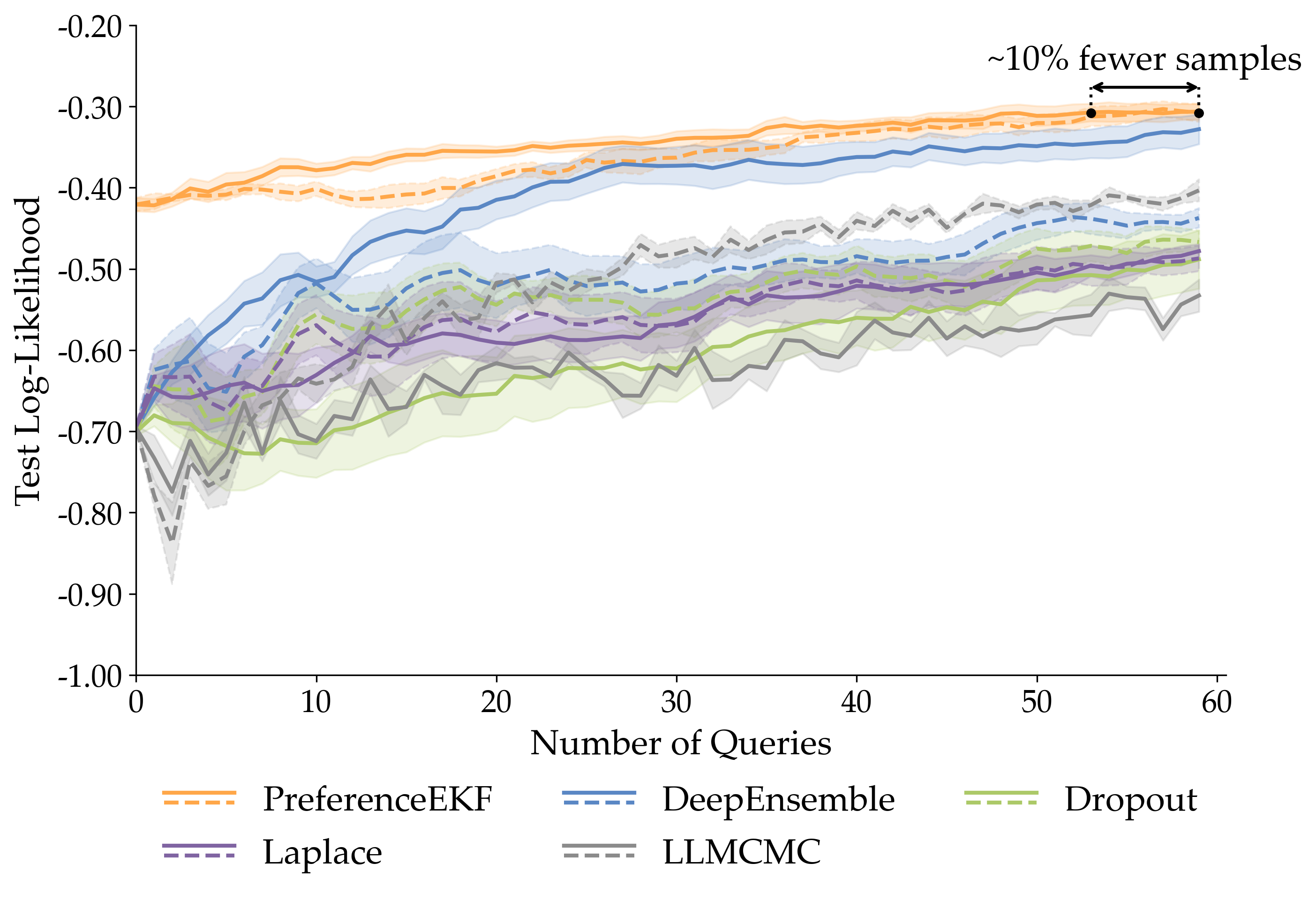}
    \caption{
    Task-aggregate reward learning performance with all algorithms using no initial dataset, thus no warmup SGD phase. Runs are averaged over all 12 tasks of 5 seeds each (mean$\pm$s.e.).
    }
    \label{fig:niter=0}
\end{figure}

It is desirable for any reward learning algorithm to still work in domains where such an initial dataset is unavailable. To investigate the reliance of all methods on access to initial data, we show in \autoref{fig:niter=0} reward learning results where we remove access to any initial data, thus making posterior updates possible only with an annotator in the loop. In this case, \subEKF uses the random projection method for subspace construction. We see that \subEKF still outperforms all baselines, showcasing its flexibility in learning reward models even without access to existing labeled queries.

\subsubsection{Reward learning from multiple human annotators} \label{appendix:human}
\begin{figure}[tbp]
    \centering
    \includegraphics[width=0.6\linewidth]{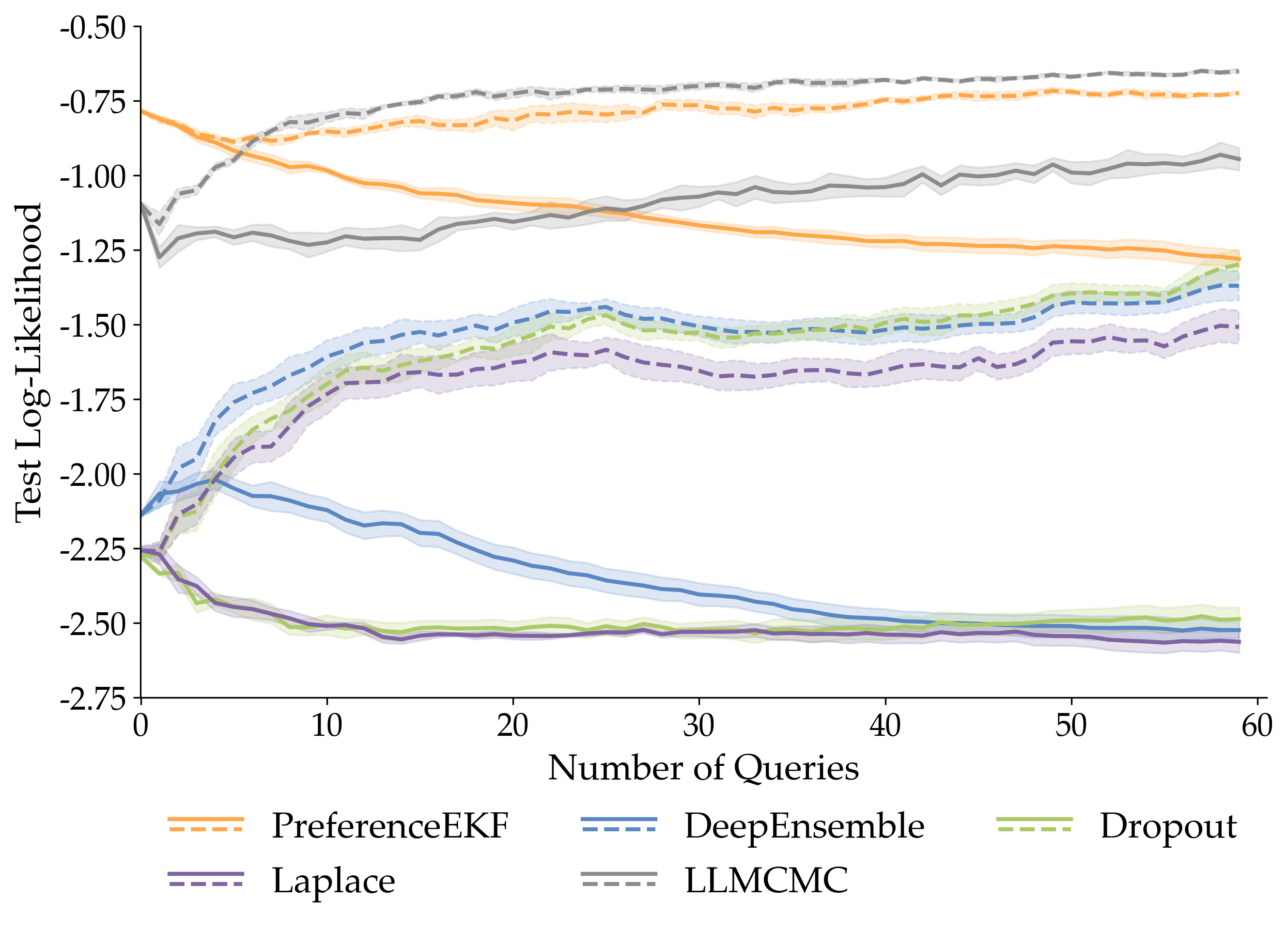}
    \caption{
    Task-aggregate reward learning performance with all algorithms using crowd-sourced real human preference labels. Runs are averaged over all 12 tasks of 5 seeds each (mean$\pm$s.e.).
    }
    \label{fig:human}
\end{figure}

Our main experiments are conducted exclusively with synthetic oracle preference labels, where of the two trajectories being compared, the trajectory with higher summed reward is designated as the preferred trajectory. To test the methods' ability to learn rewards from real human preferences, we use the crowd-sourced preference labels from \cite{yuan2023unirlhf} to perform reward learning. We show in \autoref{fig:human} that no methods reached great log-likelihood evaluation, and none of the methods' active variant was able to outperform their random variant. This is likely due to the crowd-sourced nature of the labels, which may induce multi-modal preference distribution underlying the labels that may be difficult for our benchmark algorithms to capture. We emphasize that our work's main contribution is a sample-efficient active reward learning algorithm for the single annotator setting, and we leave adaptation of our work to multi-annotator settings to future work. 

\subsubsection{Reward learning from real robotics data} \label{appendix:real_robotics}
We extend the application of our method to real-world robotics datasets, where we leverage sparse binary task success or failure signal as preference feedback. This setting is common in recent robot reward model works such as \cite{yang2024rank2reward} and \cite{liang2026robometer}. 
We use the rollout datasets from SOAR \citep{zhou2024autonomous}, where the trajectories are collected by a fleet of 5 WidowX robot arms over a variety of manipulation tasks, such as putting a blue block in a wooden bowl or transferring a mushroom from a bowl to a table. 
The trajectory observation data are 7-dimensional proprioceptive states encoding end effector translation (XYZ) and rotation (roll, pitch, yaw), along with gripper open/close state (scalar). For a given task, we generate preference labels by sampling one successful trajectory and one failed trajectory, and label the successful trajectory as the preferred one. 

\begin{figure}[tbp]
    \centering
    \includegraphics[width=0.6\linewidth]{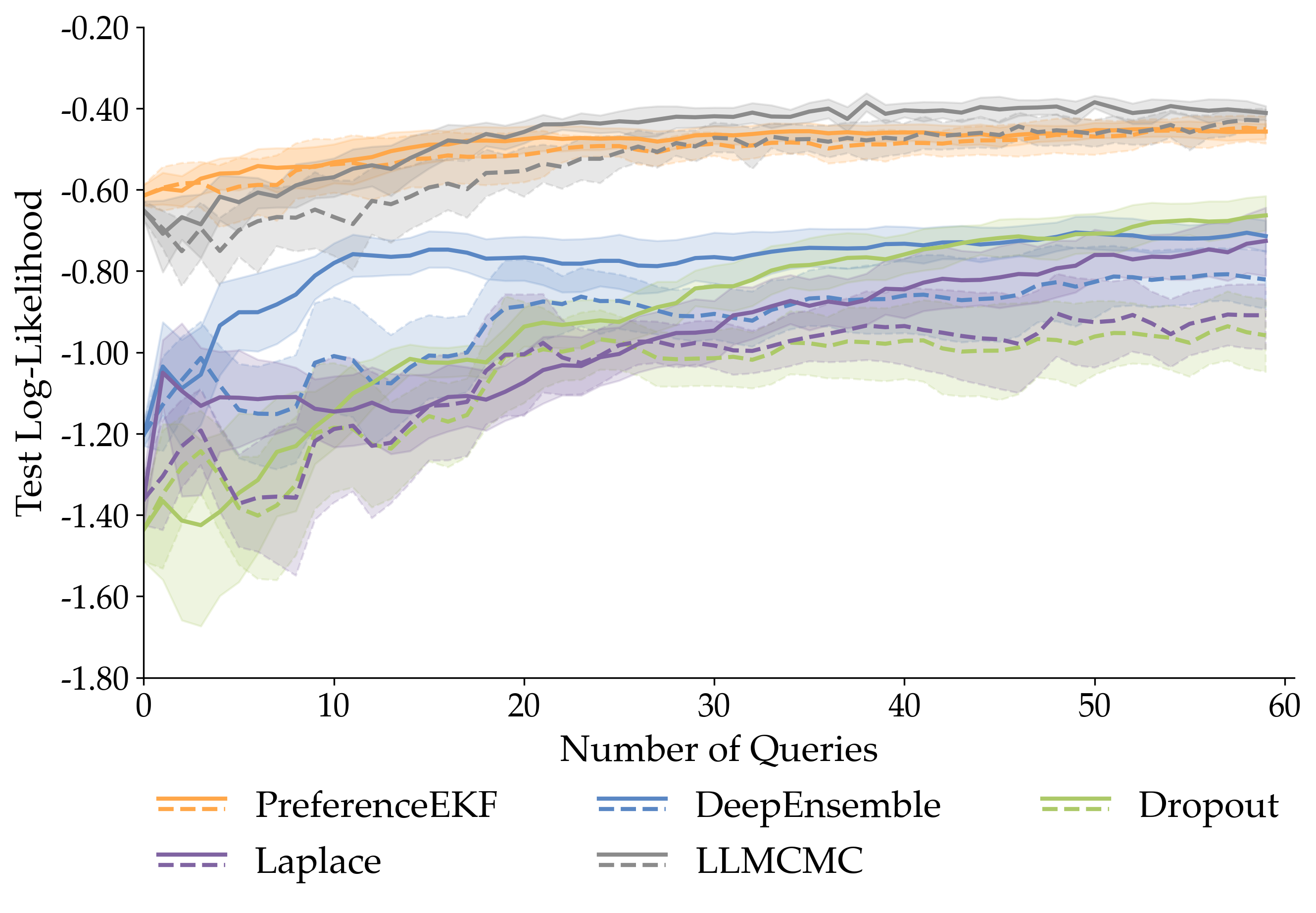}
    \caption{
    Task-aggregate reward learning performance of all methods on the SOAR dataset, with random (dashed line) and active (solid line) variants of each method. Runs are aggregated over 3 tasks (mean $\pm$ 95\% bootstrap confidence interval over 12 seeds per task).
    }
    \label{fig:realrobot_logpdf_agg}
\end{figure}

\begin{figure}[tbp]
    \centering
    \includegraphics[width=0.9\linewidth]{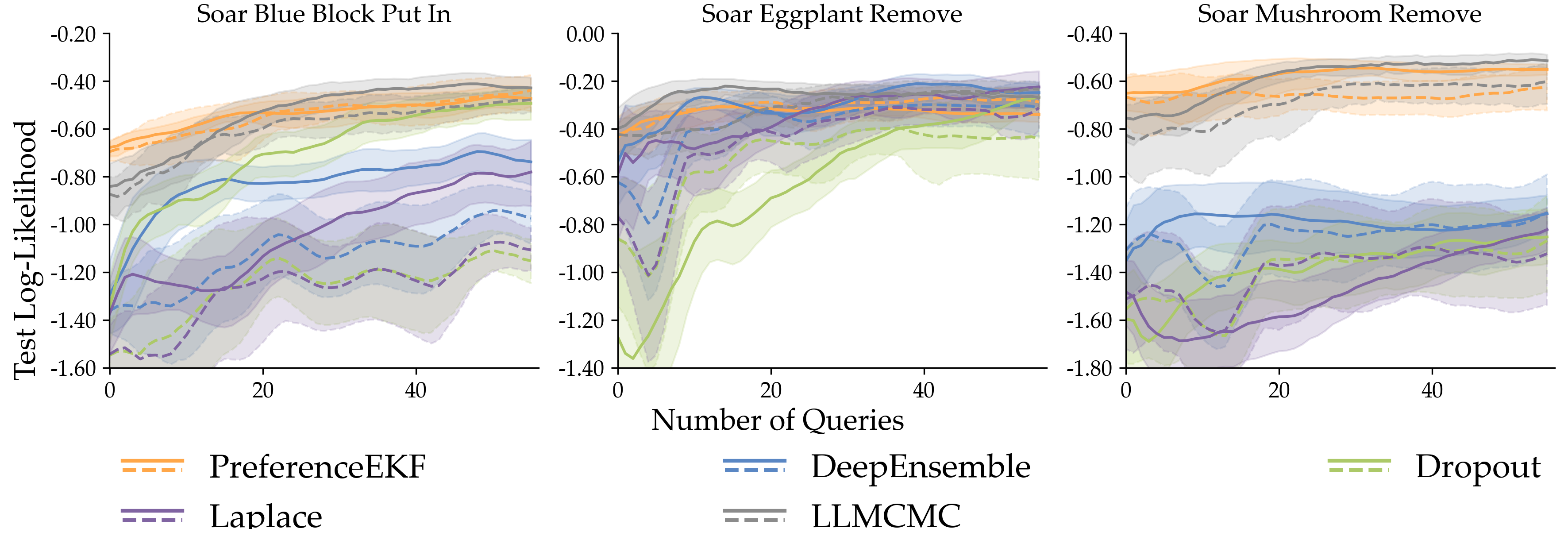}
    \caption{
    Per-task reward learning performance of all methods on the SOAR dataset, with random (dashed line) and active (solid line) variants (mean $\pm$ 95\% bootstrap confidence interval over 12 seeds).
    }
    \label{fig:realrobot_logpdf_all}
\end{figure}

Since the SOAR dataset was primarily collected for the purpose of learning from suboptimal data free of human supervision in the real world, and our pairwise sampling procedure requires (success, failed) trajectory pairings, we found that many task datasets from SOAR contained either 1) too few trajectories in total or 2) way more failed trajectories than successful ones. We narrowed down our evaluation task suite down to 3 tasks that contained sufficient number of (success, failed) pairings, and evaluated all methods on all tasks. All trajectories are of fixed length of 100 steps, with no partial segment sampling as we do in our main results in \autoref{section:experiments_activelearning}.
In \autoref{fig:realrobot_logpdf_agg}, we show that \subEKF and \llmcmc both achieve the best performance over all methods considered, showcasing the applicability of our method to real world robotics data. We further show per-task performance in \autoref{fig:realrobot_logpdf_all}. 

\subsubsection{Reward learning from pixel data} \label{appendix:pixel}
While our main results in \autoref{sec:experiments} are performed on state-based control tasks, here we showcase the applicability of \subEKF to pixel-based tasks. We focus on the Visual D4RL (V-D4RL) benchmark \citep{lu2023challenges}, which contains rendered pixel-image observations corresponding to datasets from the state-based D4RL benchmark. 

Our pixel-based reward model architecture consists of an ImageNet-pretrained ResNet18 image encoder with embedding dimension of $512$ \citep{deng2009imagenet,he2016deep} as the backbone and a two-layer MLP with $256$ hidden units per layer as the reward prediction head. We finetune the entire reward model via SGD as part of the belief initialization step of Line \autoref{algl:belief_init}, and perform EKF inference within the subspace of only the reward head parameters while keeping the finetuned backbone frozen. We take a similar approach for the baseline methods, \ensemble and \dropout; due to computational constraints, we did not include \laplaceapprox and \llmcmc.
Due to the increased task and model complexity, we construct a subspace with dimensionality of $500$ (compared to $200$ in the state-based tasks with smaller reward models), and use random projection to do so since a larger subspace benefits equally from random projection versus SVD-based construction techniques as shown in \autoref{fig:subdim_ablation}. 

Since EKF's belief update procedure scales cubically with dimensionality of the observation space, we use a measurement likelihood function (\autoref{eq:sequential_bayesian}) over trajectory embeddings rather than raw trajectory pixels. We compute embeddings from the final layer of the ResNet18 backbone before the reward prediction head, and mean-pool the embeddings across all timesteps of a trajectory segment to obtain embeddings that aggregate trajectory-level information.
Empirically, raw pixel observations over trajectory segment lengths of $10$ steps with images of height, width, channel $(84,84,3)$ would result in observation dimension of $10 \times 84 \times 84 \times 3=211,680$ per trajectory, while mean-pooled embedding-based observation results in dimension of $512$ per trajectory.

\begin{figure}[tbp]
    \centering
    \includegraphics[width=0.6\linewidth]{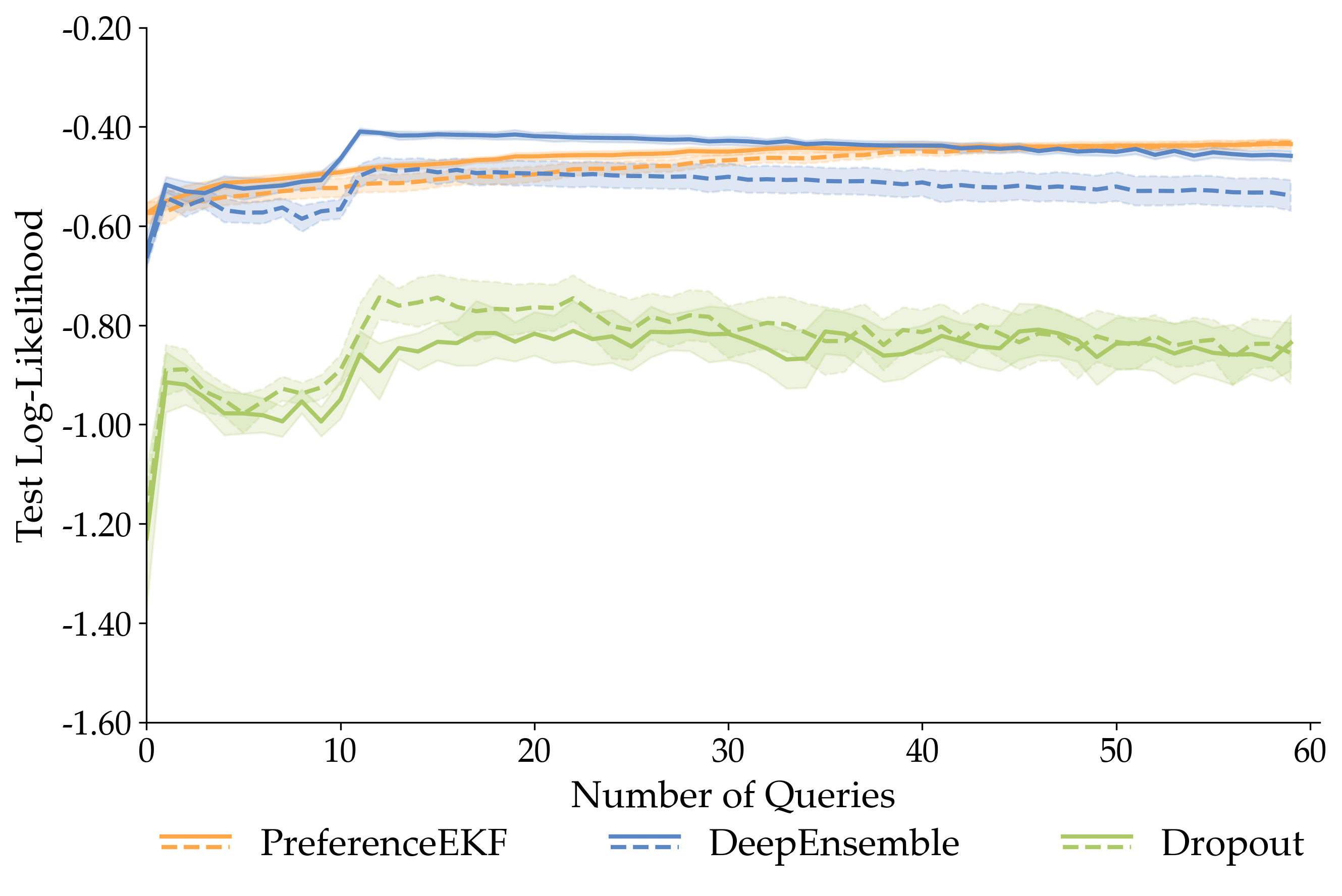}
    \caption{
    Task-aggregate reward learning performance of \subEKF on the pixel-based V-D4RL benchmark, with random (dashed line) and active (solid line) variants. Runs are aggregated over 3 pixel-based VD4RL tasks (mean±s.e. over 5 seeds).
    }
    \label{fig:pixel_logpdf_agg}
\end{figure}

\begin{figure}[tbp]
    \centering
    \includegraphics[width=0.9\linewidth]{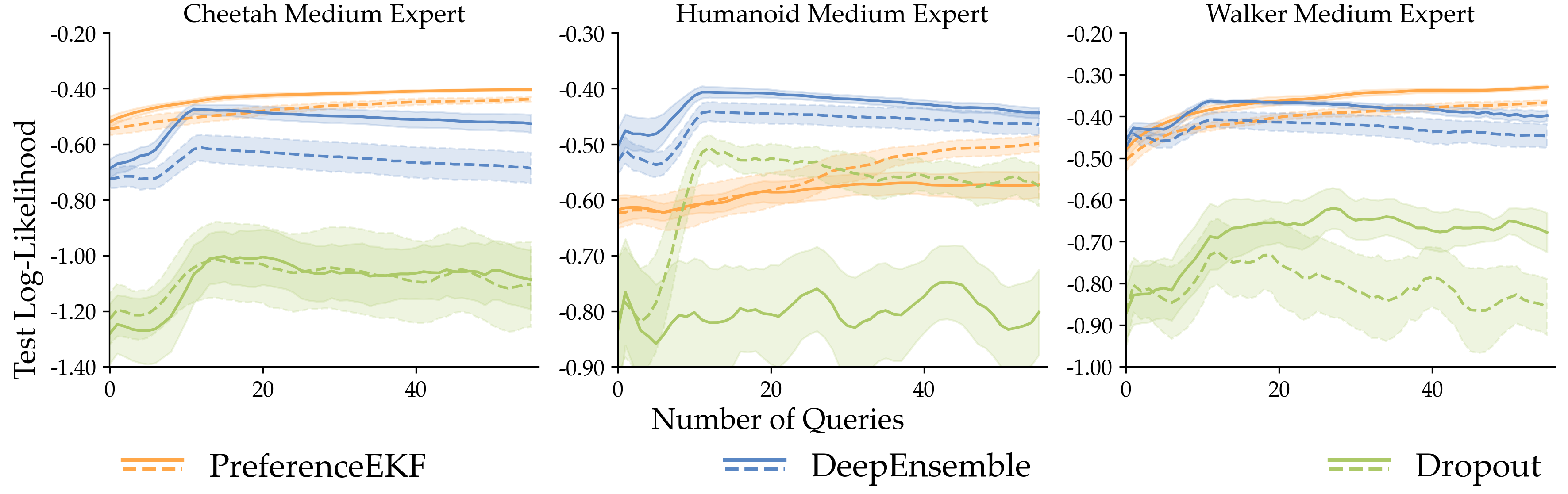}
    \caption{
    Per-task reward learning performance of \subEKF on the pixel-based V-D4RL benchmark, with random (dashed line) and active (solid line) variants. (mean±s.e. over 5 seeds).
    }
    \label{fig:pixel_logpdf_all}
\end{figure}

To finetune the pixel-based reward model which includes the entire ResNet18 backbone, we start with a much bigger initial query dataset of $150$ (compared to just 8 in state-based experiments), and use a reduced learning rate of $0.0001$ over $3000$ mini-batches with batch size $16$.
In \autoref{fig:pixel_logpdf_agg} and \autoref{fig:pixel_logpdf_all}, we show that \subEKF is indeed a viable method for active preference-based reward learning, and performs on par with \ensemble while \dropout's performance suffers. While the performance of active versus random sampling varies across the three chosen pixel-based tasks, the active variant of \subEKF as a whole shows promising improvement over the random variant. We leave research on EKF variants that efficiently scale with observation dimension, and more parameter-efficient subspace inference methods such as those based on LoRA \citep{hu2021lora} to future work.

\subsubsection{Model calibration experiments} \label{appendix:calibration}

\begin{figure*}[tbp]
    \centering
    \includegraphics[width=0.85\linewidth]{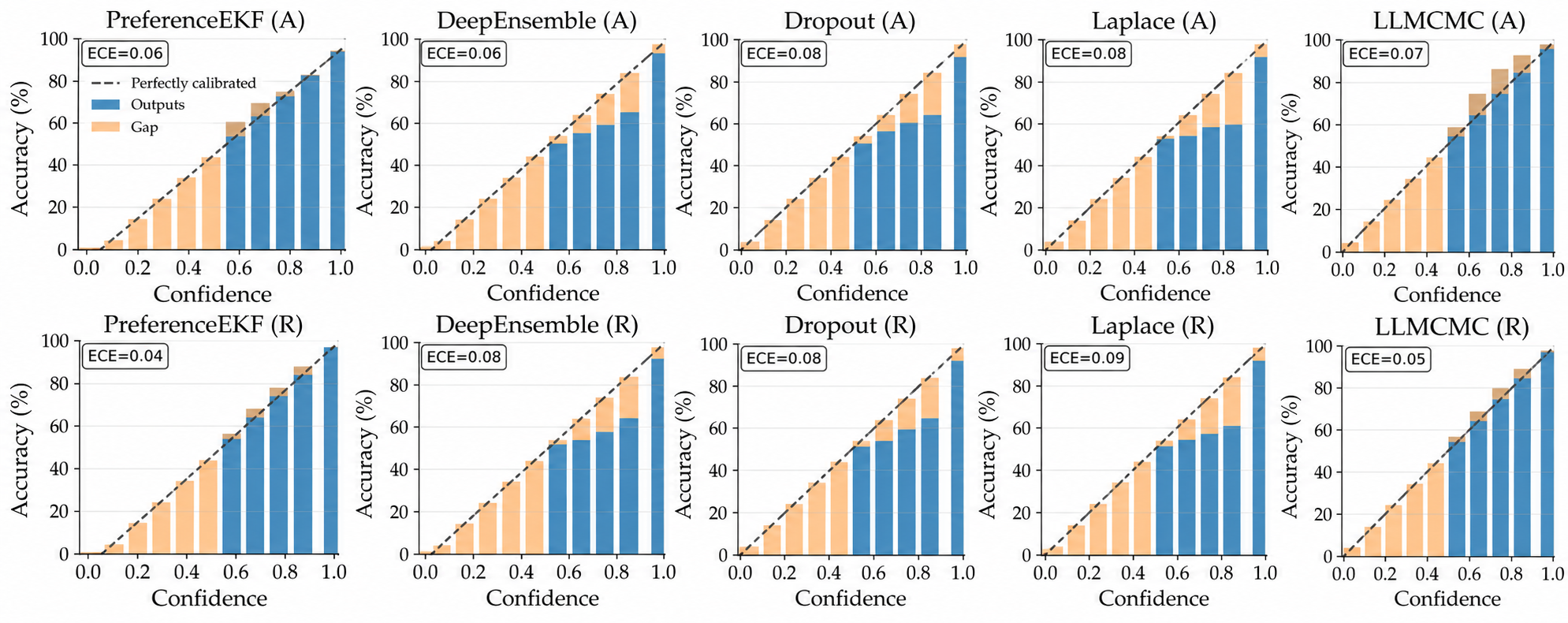}
    \caption{Task-aggregate reliability diagram for all five methods' random and active variants. Diagrams are aggregated across all tasks and seeds.}
    \label{fig:reliability}
\end{figure*}

In addition to the results from \autoref{sec:experiment_calibration} on expected calibration error and Brier scores, we provide in \autoref{fig:reliability} reliability diagrams computed from model predictions over all tasks and seeds. Due to the per-timestep parameterization of the reward model for computing the Bradley-Terry loss function \autoref{eq:bradleyTerry}, our binary preference query dataset is implemented to always have the second item be preferred over the first item. This corresponds to label of always $1$, hence why the reliability diagrams only show calibration for half of the probability line. For both reliability diagram and expected calibration error, we discretize the $[0,1]$ probability space into 10 bins. Upon inspection, we can see that \subEKF and \llmcmc exhibit the lowest model calibration error.

\subsection{Offline reinforcement learning} 

\subsubsection{Policy performance results} \label{appendix:offlineRL_results}
\autoref{fig:rlScore_agg} shows comparison of policy optimization using the reward models learned from random and active variants of each algorithm, aggregated across 12 D4RL tasks in the offline RL setting (mean$\pm$s.e. over 5 seeds). \autoref{fig:rlScore_all} shows per-task results for offline RL evaluations. All results here are shown with a moving average over the last 5 evaluations.

We observe that despite the marked difference in log-likelihood-based preference learning evaluation between methods (\autoref{fig:logpdf_agg}), when the reward models produced by each method are used in the offline RL setting for policy optimization, they all led to policies of similar rollout performance. This lack of consistent correlation between supervised learning of reward models and reinforcement learning of policies is a known behavior in the RL from preference feedback literature, across both language modeling and control domains \citep{gao2022scaling, tien2022causala, pan2021effects}.  
We emphasize that the primary contribution of our work is an efficient reward learning algorithm for learning from preference feedback, and show that the resulting reward model can produce policies that reach competitive performance with policies that learn from reward models produced by other preference learning algorithms. We do not claim that our method can automatically lead to stronger policy performance. We leave investigation of the relationship between the learned reward posterior and how it affects policy optimization to further work \citep{razin2025what,swamy2025all}. 

\subsubsection{Implementation Details} \label{appendix:offlineRL_implementations}

The extent to which offline RL algorithms leverage reward information for policy optimization, i.e., whether reward-induced policy performance is a good metric for assessing learned reward models, is heavily dependent on the trajectory dataset: when run on datasets consisting solely of expert demonstrations, offline RL algorithms will largely ignore reward information and adopt a behavioral cloning-like learning strategy. On the other hand, it is generally difficult to train a policy from a dataset consisting of purely random behavior \citep{kumar2021should}.

\begin{figure*}[tbp]
\centering
\includegraphics[width=0.85\textwidth]{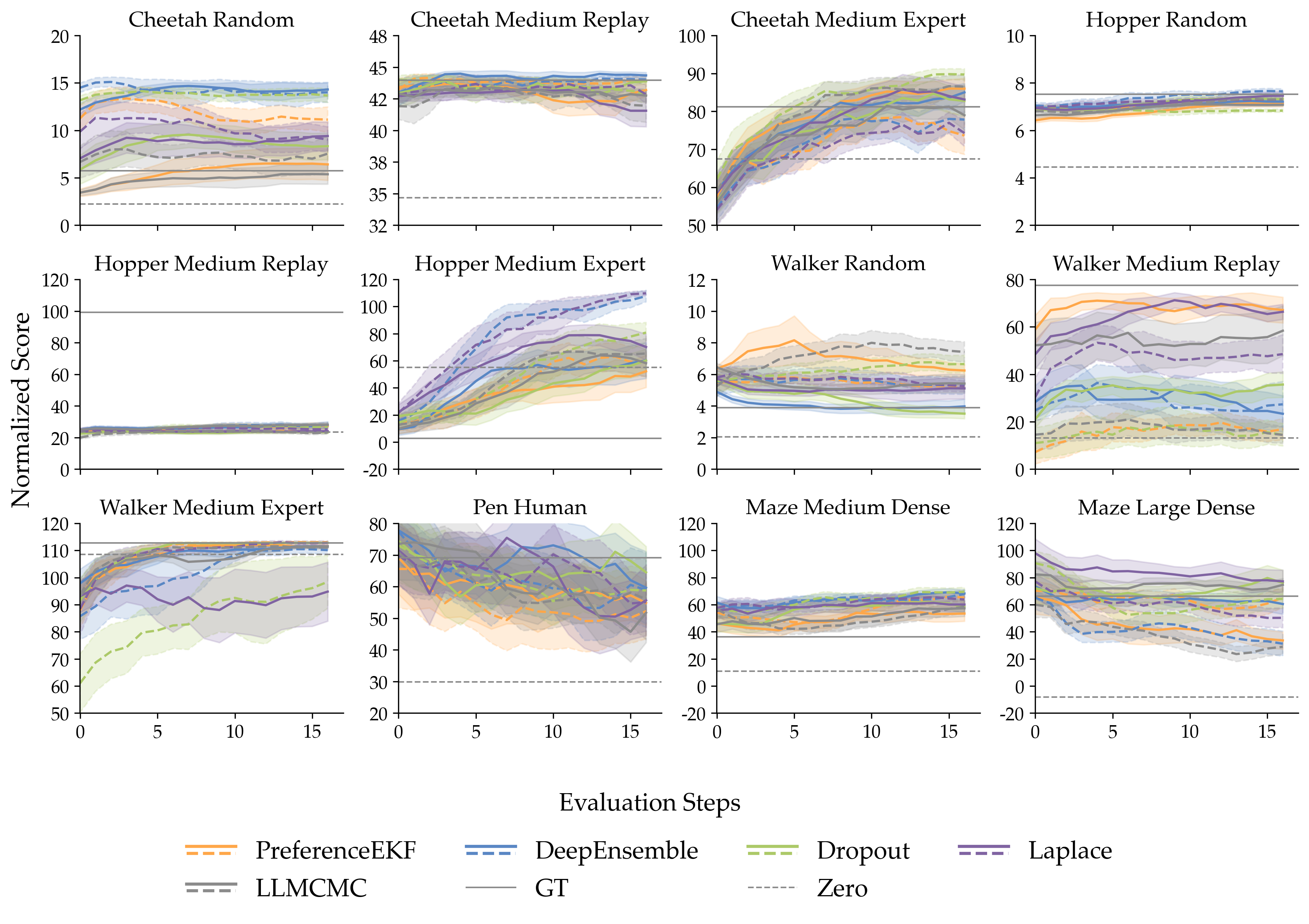}
\caption{
    Per-task offline RL rollout performance across all tasks for each reward learning algorithm variant, all using the InfoGain acquisition function: comparison of the RM learned using random (dashed line) and active (solid line) variants of the algorithms across 12 D4RL tasks in the offline RL setting (mean$\pm$s.e. over 5 seeds). 
    Black solid line indicates the performance of a policy trained on ground truth reward (GT), and black dotted line for a policy trained without reward information (Zero). In most tasks, active \subEKF performs on par with other algorithms in terms of rollout score.
}
\label{fig:rlScore_all}
\end{figure*}

Following the experiment methodology of \citet{shin2022benchmarks} for our offline RL experiments, we add two reference performance scores to every task as shown in \autoref{fig:rlScore_all}: we refer to ``GT'' as the score from an offline RL policy trained on $\datasetTraj$ labeled with ground-truth environment reward information, and ``Zero'' as score from a policy trained on $\datasetTraj$ with reward information zeroed out. This serves to test whether an offline RL algorithm is able to effectively leverage reward information for a given trajectory dataset.
For most tasks, GT and Zero serve as upper and lower performance bounds for learned policies.

All offline RL experiments were done by running implicit Q-learning (IQL \citet{kostrikov2021offline})  on trajectory transition datasets labeled with different types of rewards, e.g., ground truth environment reward, zeroed out reward, or preference-learned reward. An IQL agent consists of four neural networks: main and target Q-network, a Gaussian policy network, and a state-value network. All four networks have two hidden layers of 256 units each and are trained using the same optimizer configuration with cosine decay learning rate schedule. Policy extraction is done with advantage-weighted regression (AWR \citet{peng2019advantageweighted}). 
All training runs are done using 1M update steps with 5 rollouts every 50K steps for evaluation. We apply normalization to both reward and observation features, and further apply clipping for reward values exceeding 10. All hyperparameters are detailed in \autoref{table:iql}. 

\begin{table}[tbp]
\centering
\caption{Shared hyperparameters for IQL across all tasks. Here ``Iterations'' refers to the number of minibatch updates.}
\label{table:iql}
\begin{tabular}{ll}
\toprule
\textbf{Name} & \textbf{Value} \\
\midrule
Optimizer & Adam \\
Learning rate & $0.0003$ \\
Betas & (0.9, 0.999) \\
Iterations & 1M \\
Batch size & 256 \\
Discount factor $\gamma$ & 0.99 \\
Target net update step size & 0.005 \\
Expectile $\tau$ & 0.7 \\
Advantage temperature $\beta$ & 3.0 \\
Exponential advantage clip & 100 \\
\bottomrule
\end{tabular}
\end{table}

\subsection{Scaling Experiments.} \label{appendix:scaling}
JAX offers efficient function vectorization using $\texttt{jax.vmap}$. While we use this to parallelize ensemble model training and prediction in most experiments in \autoref{sec:experiments}, we do not use this for the scalability experiments in \autoref{sec:experiment_scale}. Parallelized training and prediction of up to $M=150$ models with up to 2M parameters (in the case of the three layer neural networks with 1024 units each) can lead to out-of-memory errors. We instead use Python's native for loop to perform ensemble model training and prediction sequentially. All scalability experiments were done on CPU instead of GPU to avoid out-of-memory errors.



\subsection{LLM Usage} \label{appendix:llm}
We used LLMs primarily for writing Python visualization scripts, figures/tables typesetting in LaTeX, finding related work on subspace construction methods, and debugging JAX compilation / model loading errors. We did not use LLMs for paper writing, research ideation, or implementing the core algorithm parts.

\end{document}